 \documentclass[final,5p,times,twocolumn,numeric]{elsarticle}

\usepackage{makecell}
\usepackage{amssymb}
\usepackage{adjustbox}
\usepackage{lipsum}
\usepackage{enumitem}
\usepackage{tikz}
\usepackage{fontawesome}
\usepackage{scalerel}
\usepackage[edges]{forest}
\definecolor{hidden-draw}{RGB}{0,0,0}
\definecolor{hidden-pink}{RGB}{168,191,143}

\usepackage[colorlinks]{hyperref}
\usepackage{amsmath} 
\usepackage{multirow}
\usepackage{bbding}
\usepackage{graphicx}
\usepackage{subfigure}
\usepackage{bm}
\usepackage{colortbl}  
\usepackage{array} 
\usepackage{tabularx}
\usepackage{framed}
\usepackage{ulem}
\usepackage{booktabs}
\usepackage{stfloats}
\usepackage{xcolor}

\newcommand{\add}[1]{{\color{black}{#1}}} 
\newcommand{\mo}[1]{{\color{black}{#1}}} 
\definecolor{mycolor}{RGB}{134,150,167}
\definecolor{myorange}{RGB}{253,229,206}
\definecolor{mygreen}{RGB}{213,232,212}
\definecolor{myblue}{RGB}{180,199,231}
\definecolor{slightblue}{RGB}{189,215,238}

\definecolor{reference}{RGB}{55,126,168}
\AtBeginDocument{
  \hypersetup{
            linkcolor=reference,
            anchorcolor=reference,
            citecolor=reference,
            filecolor=reference,
            urlcolor=reference
}
}

\newcommand{\red}[1]{\textcolor{red}{#1}}

\newcommand{\fire}{\includegraphics[width=0.3cm]{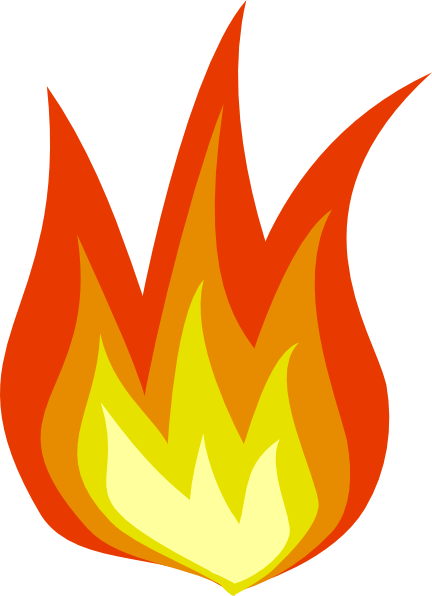}}

\journal{}

\begin{document}

\begin{frontmatter}



\title{Deep Learning for Cross-Domain Data Fusion in Urban Computing: \\Taxonomy, Advances, and Outlook}


\author[ustgz]{Xingchen Zou\fnref{label}}
\author[ustgz]{Yibo Yan\fnref{label}}
\author[ustgz]{Xixuan Hao}
\author[ustgz]{Yuehong Hu}
\author[ustgz]{Haomin Wen}
\author[ustgz]{Erdong Liu}
\author[jdt]{\\Junbo Zhang}
\author[thu]{Yong Li}
\author[swjtu]{Tianrui Li}
\author[jdt]{Yu Zheng}
\author[ustgz]{Yuxuan Liang\corref{cor1}}
\affiliation[ustgz]{organization={The Hong Kong University of Science and Technology (Guangzhou)},
            city={Guangzhou},
            country={China}}
\affiliation[jdt]{organization={JD Technology \& JD Intelligent Cities Research},
            city={Beijing},
            country={China}}
\affiliation[thu]{organization={Tsinghua University},
            city={Beijing},
            country={China}}
\affiliation[swjtu]{organization={Southwest Jiaotong University},
            city={Chengdu},
            country={China}}
 \cortext[cor1]{Y. Liang is the corresponding author. Email: yuxliang@outlook.com}
 \fntext[label]{These authors contributed equally to this work.}


\begin{abstract}

As cities continue to burgeon, \textit{Urban Computing} emerges as a pivotal discipline for sustainable development by harnessing the power of cross-domain data fusion from diverse sources (e.g., geographical, traffic, social media, and environmental data) and modalities (e.g., spatio-temporal, visual, and textual modalities). Recently, we are witnessing a rising trend that utilizes various deep-learning methods to facilitate cross-domain data fusion in smart cities. To this end, we propose the first survey that systematically reviews the latest advancements in deep learning-based data fusion methods tailored for urban computing. Specifically, we first delve into data perspective to comprehend the role of each modality and data source. Secondly, we classify the methodology into four primary categories: \textit{feature-based}, \textit{alignment-based}, \textit{contrast-based}, and \textit{generation-based} fusion methods. Thirdly, we further categorize multi-modal urban applications into seven types: \textit{urban planning}, \textit{transportation}, \textit{economy}, \textit{public safety}, \textit{society}, \textit{environment}, and \textit{energy}. Compared with previous surveys,  we focus more on the synergy of deep learning methods with urban computing applications. Furthermore, we shed light on the interplay between \textit{Large Language Models (LLMs)} and urban computing, postulating future research directions that could revolutionize the field. We firmly believe that the taxonomy, progress, and prospects delineated in our survey stand poised to significantly enrich the research community. The summary of the comprehensive and up-to-date paper list can be found at \url{https://github.com/yoshall/Awesome-Multimodal-Urban-Computing}.

\end{abstract}




\end{frontmatter}




\section{Introduction}
\label{introduction}

Cities,  indispensable components of modern civilization, have undergone transformative trajectories propelled by human advancements in cultural, financial, political, and technological domains \cite{chen2023impact,xu2023exploring,zheng2015methodologies,wen2014dynamic}. Despite their pivotal role in societal progress, the unprecedented surge in global urbanization since the 19th century has precipitated formidable sustainability challenges including energy consumption \cite{wang2020does,wu2019does}, environmental pollution \cite{liang2019effect,khan2022energy,wang2024boosting}, socio-economic disparities \cite{yao2021urbanization,li2019urbanization}, and urban traffic issues  \cite{kruszyna2021dependencies,li2019urbanization,ouallane2022fusion,liang2024time}. In the 21st century, the profound strides achieved in machine learning and spatio-tempral data mining have manifested in myriad successful applications across diverse domains, such as finance \cite{huang2020deep,ozbayoglu2020deep,bin2023rhpmf}, biology \cite{mahmud2021deep,tang2019recent}, and healthcare \cite{stiglic2020interpretability,shahid2019applications}. This surge in technological prowess has incited a notable shift in research focus, with scholars now directing their attention toward harnessing these advancements to optimize the intricate facets of urban planning, operations, management, etc. A pivotal contribution to this evolving discourse is the pioneering work \cite{zheng2014urban}, which encapsulated and elucidated these endeavors by introducing the concept of \textbf{Urban Computing}. This paradigm leverages sensing technologies and expansive computing infrastructure to scrutinize voluminous data emanating from urban spaces.   The fundamental objective is to gain profound insights into the dynamics of cities, thereby addressing challenges such as traffic congestion  \cite{ji2023spatio,zhang2021traffic,pan2020spatio}, energy consumption \cite{johari2020urban,hashem2023urban,piccialli2024graphite}, and air quality pollution \cite{liang2023airformer,yi2018deep,yi2020predicting}. 

\begin{figure*}[!ht]
    \centering
    \includegraphics[width=0.9\textwidth]{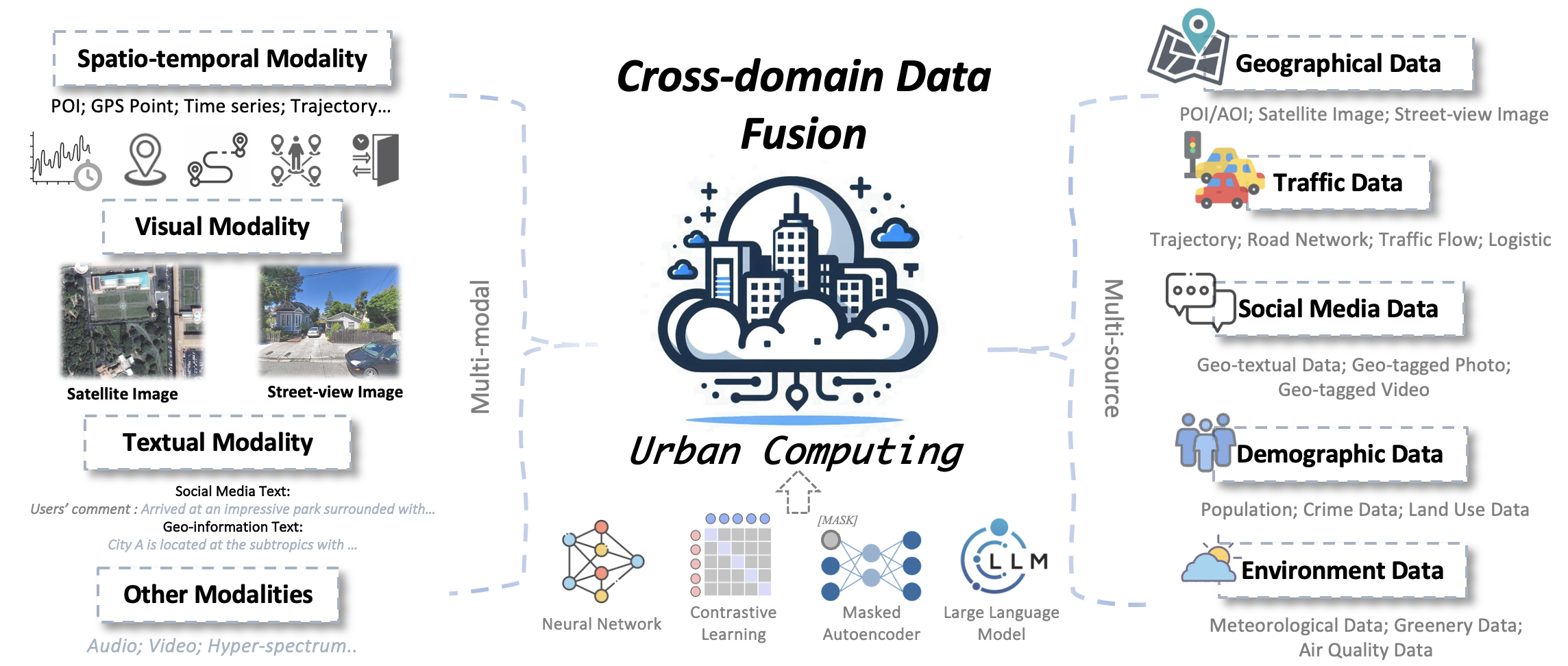}
    \vspace{-1em}
    \caption{A sketch of cross-domain urban computing. \textbf{Left}: It involves the integration of urban data from diverse modalities, including spatio-temporal, visual, textual, and other modalities, through the process of data fusion. \textbf{Right}: Generally, these urban data derive from multiple sources, such as geographical data, transportation, social media, demography, and environment.}
    \label{fig:intro}
\end{figure*}

Urban computing necessitates the integration of extensive and diverse datasets sourced from various sources and modalities~\cite{zheng2014urban,zheng2015methodologies, fadhel2024comprehensive}, also referred to as \textbf{Cross-Domain Data Fusion}, which arises from the recognition that relying solely on a singular data source or modality may prove inadequate for the holistic implementation of urban tasks. \add{For example, in the realm of traffic prediction, it becomes imperative to assimilate meteorological forecast data with geographical information. This involves taking into account the congestion induced by rainfall and the influence of school and business hours on traffic flow during peak periods. In the context of urban planning, one must combine population density and economic activity data. This includes evaluating factors such as population density and income levels when devising plans for new commercial districts to ascertain their viability. Furthermore, in the field of public safety management, integrating crime data with socio-economic data is crucial. This entails considering unemployment rates and education levels in high-crime areas when strategizing police deployment plans.} In recent years, a growing number of studies in urban computing are expanding cross-domain data fusion to encompass diverse sources like sensors \cite{ji2016urban}, satellites \cite{liu2023knowledge}, social media \cite{zhao2020pgeotopic}, and citizen-generated data \cite{zhao2016towards}. Additionally, there is a trend towards introducing new data modalities, including text (e.g., social media posts \cite{zhao2016towards} and geographic information \cite{huang2022ernie,chen2021location}) and images (e.g., satellite \cite{liu2023knowledge,xiFirstLawGeography2022b,li2022predicting} and street-view images \cite{liu2023knowledge,li2022predicting,huang2023comprehensive}). Figure \ref{fig:intro} depicts the cross-domain data fusion in urban computing from the views of data modality and source.

Prior research, exemplified by \citet{zheng2014urban}, emphasized the critical role of cross-domain data fusion in amalgamating information from multiple sources. With the emergence of data fusion studies in urban computing, \citet{zheng2015methodologies} classified the related fusion methodologies into three types: stage-based, feature level-based, and semantic-based data fusion. Furthermore, within the purview of semantic-based methods, a more intricate taxonomy emerges, delineating four subtypes: multi-view learning-based, similarity-based, probabilistic dependency-based, and transfer learning-based methods.

\begin{figure}[!h]
    \centering
    \vspace{-1em}
    \includegraphics[width=0.45\textwidth]{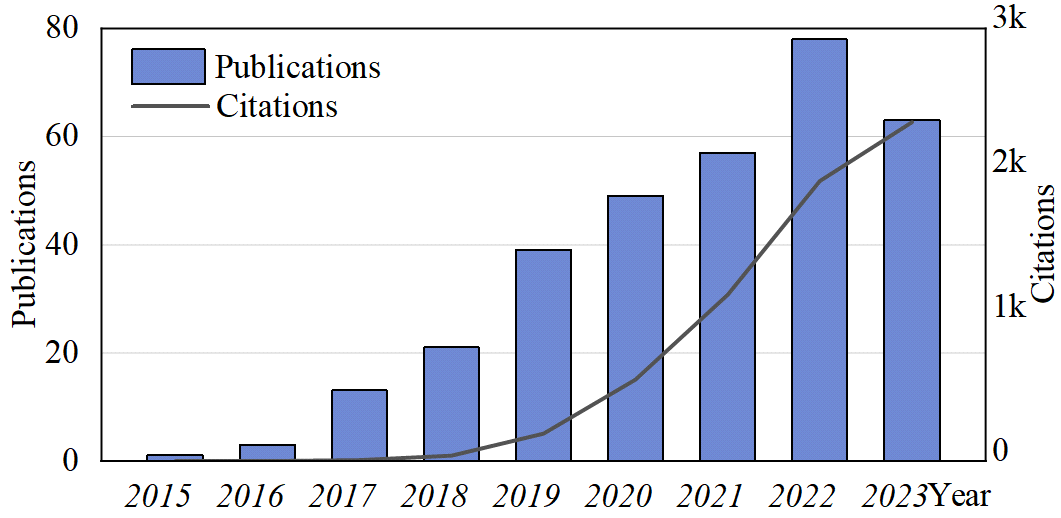}
    \vspace{-1em}
    \caption{\add{Times cited and publications over time for Deep Learning in Urban Computing on prestigious venues(Source: Web of Science)}}
    \vspace{-1em}
    \label{fig:citation}

\end{figure}

Over the past decade, the emergence of Deep Learning (DL) has asserted dominance in processing spatio-temporal data in urban computing~\cite{wang2020deep,jin2023spatio}. In contrast to traditional machine learning methods, these models present distinct superiority, characterized by larger model capacity, automated feature extraction capabilities, and inherent compatibility with cross-domain data fusion. \add{As Figure \ref{fig:citation} illustrates, there has been a significant uptick in both the number of papers published and the citations received for research related to deep learning in urban computing since 2015.} The paradigm shift derived from deep learning renders previous surveys, especially \cite{zheng2015methodologies}, on urban data fusion somewhat obsolete, as traditional taxonomy may not aptly capture the nuances and differences among these advanced methodologies. In light of this issue, \emph{our survey is dedicated to bridging this gap and provides a contemporary perspective by offering a comprehensive and updated taxonomy that aligns with the era of deep learning.} Through a thorough examination of deep learning-based cross-domain data fusion methods, we seek to establish a robust foundation for understanding and navigating the landscape of urban computing.

Specifically, we commence by presenting a novel taxonomy that classifies existing urban data sources into five distinct types, while concurrently categorizing fusion methods into four types. This systematic classification offers valuable insights into the intricate integration of diverse modalities within urban computing. Secondly, we systematically categorize widely used datasets and outline common application scenarios for urban computing fusion models. Thirdly, inspired by the breakthrough of Large Language Models (LLMs) in financial time series \cite{wu2023bloomberggpt,yu2023temporal}, medicine \cite{thirunavukarasu2023large,singhal2023large}, law \cite{cui2023chatlaw}, and climate \cite{schimanski2023climatebert,laud2023climabench}, we further summarize inspiring research that integrates LLMs into urban computing, which complements the basic taxonomy of cross-domain data fusion in urban computing.

\textbf{Related Surveys}. Several surveys have recently explored the application of deep learning-based data fusion across various domains. Notably, \citet{zheng2015methodologies} conducted a comprehensive investigation into the methodologies proposed for cross-domain big data fusion prior to 2015. Their research reveals that machine learning-based data fusion was the de-facto dominant approach in urban computing during that period, albeit with challenges in comprehending cross-modal relations. Subsequently, with the remarkable success of deep learning models such as recurrent neural networks (RNN) and convolutional neural networks (CNN) in representation learning, \citet{wang2020deep} conducted an exhaustive review on the utilization of deep learning for spatio-temporal data mining, with a special focus on the fusion of multi-source spatio-temporal data. \citet{liu2020urban} further provided a summary of deep learning-based data fusion methodologies for urban big data fusion before 2020. Though these surveys shed light on the advancements in deep learning to their respective fields, they primarily concentrated on specific aspects and do not extensively cover cross-modal data fusion or the latest paradigms, including contrast-based and generation-based approaches which have gained significant popularity since 2021. In other words, none of them have provided a comprehensive and up-to-date taxonomy framework for data fusion methodologies in the context of urban computing. 

Furthermore, existing surveys lack a specific focus on the data and application perspectives of these DL models within the field of urban computing. For instance, \citet{gao2022generative} summarized the fusion models for spatio-temporal data based on generative adversarial networks (GAN), while \citet{deldari2022beyond} concentrated on self-supervised representation learning on the fusion of multi-modal data in the general domain. Besides, there are a couple of surveys investigating the deep learning-based data fusion in specific applications (i.e., a subdomain) of urban computing, such as crowd flow prediction \cite{xie2020urban}, intelligent transportation \cite{yuan2021survey}, and social event detection \cite{afyouni2022multi}. These facts highlight the need for a new survey to serve as a guide for future endeavors in data fusion in urban computing concepts.

To this end, this paper aims to provide a \textit{comprehensive} and \textit{up-to-date} review of deep learning-based data fusion methodologies,  explicitly tailored for cross-domain data fusion in urban computing. Our intention is not only documenting the latest advancements but also illuminating available resources and practical applications, and identifying potential research directions. Table \ref{tab:survey-compare} succinctly summarizes the key differences between our survey and other relevant surveys in the field. Through our exploration, we seek to provide a valuable resource for researchers and stakeholders, fostering an enhanced understanding of the intricacies surrounding the integration of diverse urban data modalities via deep learning approaches.

\begin{table*}[tp]
\caption{A thorough comparison between related surveys and ours, focusing on scopes (i.e., Specific versus Urban Computing), relevant modalities (e.g., general spatio-temporal data, image, text, and others), and primary topics of focus (i.e., data sources and modalities utilized for urban computing (Data), data fusion models and techniques (Fusion Model), application domains and downstream tasks in urban computing (Application) and LLMs for urban computing).}
\label{tab:survey-compare}
\resizebox{\textwidth}{!}{%
\begin{tabular}{lcccccccccccc}
\toprule[1.5pt]
\multicolumn{1}{c}{} &
  \multicolumn{1}{c}{} &\multicolumn{1}{c}{}&
  \multicolumn{2}{c}{\textbf{Scope}} &
  \multicolumn{4}{c}{\textbf{Modality}} &
  \multicolumn{4}{c}{\textbf{Focus}} \\
  \cmidrule(r){4-5} \cmidrule(r){6-9}\cmidrule(r){10-13}
\multicolumn{1}{c}{\multirow{-2}{*}{\textbf{Survey}}} &
  \multicolumn{1}{c}{\multirow{-2}{*}{\textbf{Year}}} &
    \multicolumn{1}{c}{\multirow{-2}{*}{\textbf{Venue}}} &
  \textbf{Specific} &
  \textbf{Urban Computing} &
  \textbf{Spatio-temporal} &
  \textbf{Image} &
  \textbf{Text} &
  \textbf{Others} &
  \textbf{Data} &
  \textbf{Fusion Model} &
  \textbf{Application} &
  \textcolor{purple}{ \fire} \textit{\textbf{LLM}} \\ \midrule
\citet{zheng2015methodologies} &
  2015 &IEEE Trans. Big Data&
   &
  \Checkmark &
  {\color[HTML]{2191A8} \Checkmark} &
  {\color[HTML]{2191A8} \Checkmark} &
  {\color[HTML]{2191A8} \Checkmark} &
  {\color[HTML]{E7524C} \XSolidBrush} &
  {\color[HTML]{E7524C} \XSolidBrush} &
  {\color[HTML]{2191A8} \Checkmark} &
  {\color[HTML]{E7524C} \XSolidBrush} &
  {\color[HTML]{E7524C} \XSolidBrush} \\
\citet{wang2020deep} &
  2020 & IEEE TKDE&
  \Checkmark &
   &
  {\color[HTML]{2191A8} \Checkmark} &
  {\color[HTML]{E7524C} \XSolidBrush} &
  {\color[HTML]{E7524C} \XSolidBrush} &
  {\color[HTML]{E7524C} \XSolidBrush} &
  {\color[HTML]{2191A8} \Checkmark} &
  {\color[HTML]{2191A8} \Checkmark} &
  {\color[HTML]{2191A8} \Checkmark} &
  {\color[HTML]{E7524C} \XSolidBrush} \\
\citet{liu2020urban} &
  2020 & Information Fusion &
   &
  \Checkmark &
  {\color[HTML]{2191A8} \Checkmark} &
  {\color[HTML]{2191A8} \Checkmark} &
  {\color[HTML]{2191A8} \Checkmark} &
  {\color[HTML]{2191A8} \Checkmark} &
  {\color[HTML]{E7524C} \XSolidBrush} &
  {\color[HTML]{2191A8} \Checkmark} &
  {\color[HTML]{E7524C} \XSolidBrush} &
  {\color[HTML]{E7524C} \XSolidBrush} \\
\citet{xie2020urban} &
  2020 & Information Fusion&
  \Checkmark &
   &
  {\color[HTML]{2191A8} \Checkmark} &
  {\color[HTML]{E7524C} \XSolidBrush} &
  {\color[HTML]{E7524C} \XSolidBrush} &
  {\color[HTML]{E7524C} \XSolidBrush} &
  {\color[HTML]{2191A8} \Checkmark} &
  {\color[HTML]{2191A8} \Checkmark} &
  {\color[HTML]{E7524C} \XSolidBrush} &
  {\color[HTML]{E7524C} \XSolidBrush} \\
\citet{yuan2021survey} &
  2021 & Springer Data Sci. Eng. &
  \Checkmark &
   &
  {\color[HTML]{2191A8} \Checkmark} &
  {\color[HTML]{E7524C} \XSolidBrush} &
  {\color[HTML]{E7524C} \XSolidBrush} &
  {\color[HTML]{E7524C} \XSolidBrush} &
  {\color[HTML]{2191A8} \Checkmark} &
  {\color[HTML]{2191A8} \Checkmark} &
  {\color[HTML]{2191A8} \Checkmark} &
  {\color[HTML]{E7524C} \XSolidBrush} \\
\citet{afyouni2022multi} & 
  2022 &Information Fusion &
  \Checkmark &
   &
  {\color[HTML]{2191A8} \Checkmark} &
  {\color[HTML]{2191A8} \Checkmark} &
  {\color[HTML]{2191A8} \Checkmark} &
  {\color[HTML]{E7524C} \XSolidBrush} &
  {\color[HTML]{2191A8} \Checkmark} &
  {\color[HTML]{E7524C} \XSolidBrush} &
  {\color[HTML]{2191A8} \Checkmark} &
  {\color[HTML]{E7524C} \XSolidBrush} \\
\citet{gao2022generative} &
  2022 &ACM TIST &
  \Checkmark & 
   &
  {\color[HTML]{2191A8} \Checkmark} &
  {\color[HTML]{E7524C} \XSolidBrush} &
  {\color[HTML]{E7524C} \XSolidBrush} &
  {\color[HTML]{E7524C} \XSolidBrush} &
  {\color[HTML]{E7524C} \XSolidBrush} &
  {\color[HTML]{2191A8} \Checkmark} &
  {\color[HTML]{2191A8} \Checkmark} &
  {\color[HTML]{E7524C} \XSolidBrush} \\
\citet{deldari2022beyond} & 
  2022 &Unpublished&
  \Checkmark &
   &
  {\color[HTML]{2191A8} \Checkmark} &
  {\color[HTML]{2191A8} \Checkmark} &
  {\color[HTML]{2191A8} \Checkmark} &
  {\color[HTML]{2191A8} \Checkmark} &
  {\color[HTML]{E7524C} \XSolidBrush} &
  {\color[HTML]{2191A8} \Checkmark} &
  {\color[HTML]{E7524C} \XSolidBrush} &
  {\color[HTML]{E7524C} \XSolidBrush} \\ \midrule
\textbf{Ours} &
  2024 &-&
   &
  \Checkmark &
  {\color[HTML]{2191A8} \Checkmark} &
  {\color[HTML]{2191A8} \Checkmark} &
  {\color[HTML]{2191A8} \Checkmark} &
  {\color[HTML]{2191A8} \Checkmark} &
  {\color[HTML]{2191A8} \Checkmark} &
  {\color[HTML]{2191A8} \Checkmark} &
  {\color[HTML]{2191A8} \Checkmark} &
  {\color[HTML]{2191A8} \Checkmark} \\ \bottomrule[1.5pt]
\end{tabular}%
}
\end{table*}

\textbf{Our Contributions}. Compared to previous surveys, the contributions of our survey can be summarized as follows:

\begin{itemize}[leftmargin=*]
\vspace{-0.3em}
\item \textbf{Comprehensive and Up-to-Date Survey.} To the best of our knowledge, this is the first comprehensive survey that systematically reviews studies on deep learning techniques for cross-domain data fusion models in urban computing. We firmly believe that the taxonomy, progress, and prospects introduced in this paper can significantly promote the development of this field.

\vspace{-0.3em}
\item \textbf{Novel and Structured Taxonomy.} We present a novel taxonomy, which organizes current research efforts from three perspectives: i) data sources, which mainly include spatio-temporal, visual, and textual modalities; ii) fusion methods, consisting of feature-based, alignment-based, contrast-based, and generation-based fusion methods; iii) applications, spanning diverse domains such as urban planning, transportation, economy, public safety, society, environment, and energy.

\vspace{-0.3em}
\item \textbf{Extensive Dataset Compilation.} Our survey thoroughly compiles and categorizes popular datasets in urban computing, taking into consideration their sources, temporal coverage, and spatial distribution characteristics. Additionally, we outline the prevailing common application scenarios for urban computing fusion models, examining their practical contributions and acknowledging their limitations in a series of downstream applications, respectively.

\vspace{-0.3em}
\item \textbf{Future Research Outlook.} We endeavor to identify and provide detailed explanations of several promising directions for future research, covering various aspects, including data privacy protection, the establishment of open benchmarks, the diversification of applications, and the optimization of efficiency. Moreover, capitalizing on the progress of LLMs (e.g., GPT-4 \cite{openai2023gpt4} and Sora \cite{sora2024}) and their notable benefits in fusing multi-modal and multi-source data, we further investigate their innovative applications and propose potential approaches for urban computing.
\vspace{-0.3em}
\end{itemize}

\textbf{Organization}. 
The structure of the remaining sections of this paper is as follows:
in Section 2, we present the overall taxonomy of deep-learning-based data fusion methods in urban computing, providing a broad overview of the field before delving into the specific perspectives and intricacies.
Section 3 offers a comprehensive and detailed overview of the data utilized in urban computing, covering various modalities and sources.
In Section 4, we elaborate on the fusion methods employed in urban computing, discussing their approaches and techniques.
Section 5 encapsulates the extensive applications we have compiled, highlighting the practical implementations and contributions of data fusion models in urban computing.
Section 7 outlines the challenges and promising avenues for future research in this domain, identifying potential areas of improvement and exploration. 
Finally, we conclude our paper in Section 8.

\section{Taxonomy} 
\label{taxonomy}
\begin{figure*}[htb!]
\label{fig:taxonmy}
    \centering
    \includegraphics[width=0.95\textwidth]{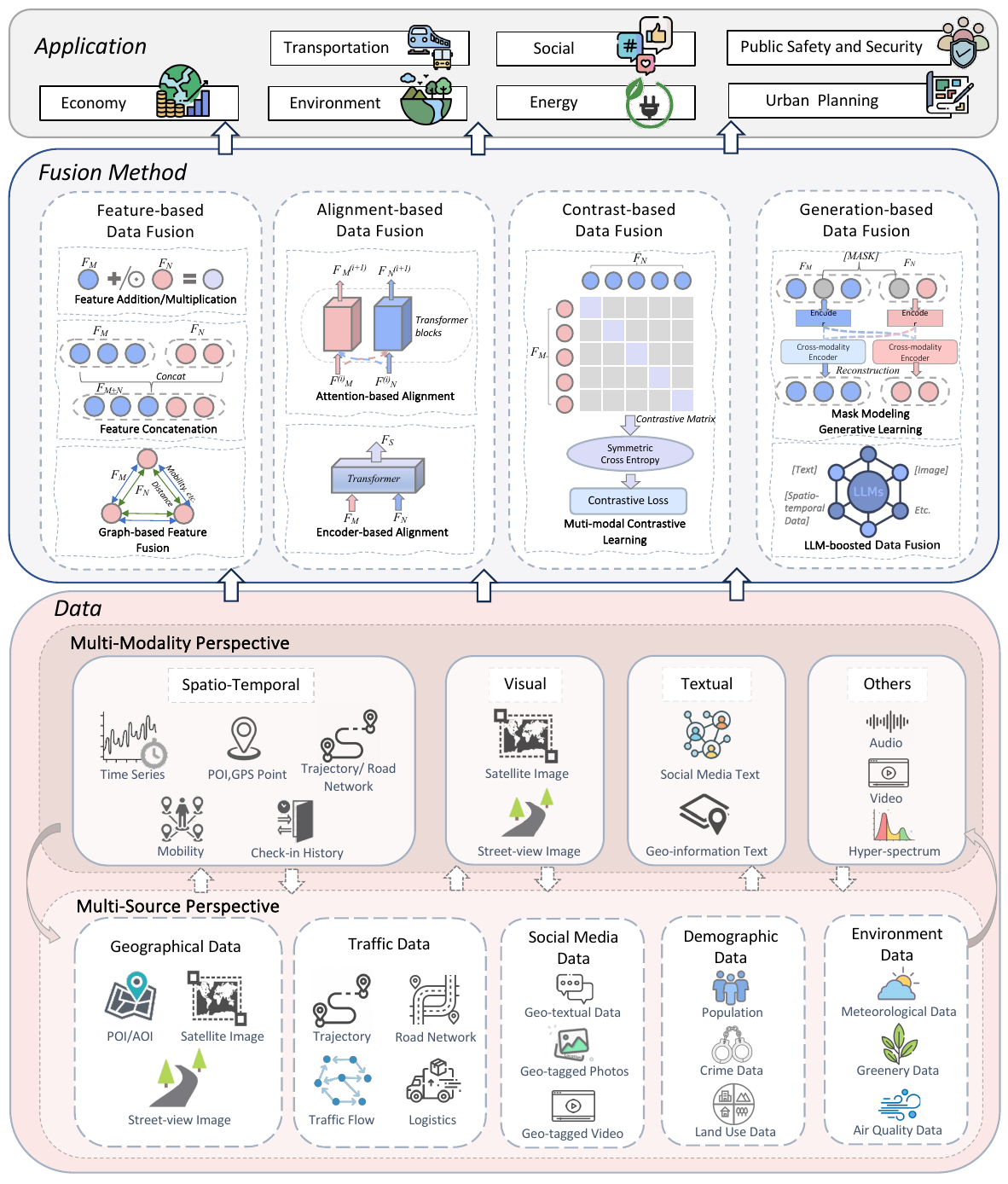}
    \caption{The taxonomy framework for deep learning-based cross-domain data fusion in urban computing in our survey. The framework is structured around three dimensions: data, fusion method, and application. Within each perspective, we categorize existing research into different categories to provide a comprehensive and well-organized review.}
\end{figure*}

This section provides a taxonomy of deep learning for multi-source and multi-modal data fusion in urban computing. As shown in Figure \ref{fig:taxonmy}, our survey is structured along three dimensions: data in cross-domain fusion in urban computing, modality fusion methods, and applications based on data fusion. A detailed synopsis of the related works can be found in Table \ref{tab:survey-compare}.

In Section \ref{SectionData}, from the data perspective, we divided the data normally utilized in urban computing into five categories according to data sources: \textit{geographical data}, \textit{traffic data}, \textit{social media data}, \textit{demographic data} and \textit{environmental data}. Additionally, we have also classified the data from a modality perspective, including \textit{spatio-temporal data}, \textit{visual data}, \textit{textual data}, and other types of data. These two categorizations allow for a systematic understanding and analysis of the different types of data used in urban computing research. We further present a comprehensive overview of public datasets used in cross-domain data fusion in urban computing in Table \ref{tab:dataset-table}.

From the fusion methodology perspective, Section \ref{FusionMethod} covers a comprehensive review of existing data fusion methods in urban computing, categorized into \textit{feature-based}, \textit{alignment-based}, \textit{contrast-based}, and \textit{generation-based fusion}. In each category, we subdivide the existing literature into several types based on the models' properties. A detailed taxonomy from a modality fusion perspective can be found in Figure \ref{fig:taxonmy}. Additionally, in the generation-based fusion section, we also focus on the recent application of LLM for data fusion in urban computing which offers valuable insights for the research community.

In Section \ref{ApplicationSection}, we divide the multi-modal application in urban computing into seven categories: \textit{urban planning}, \textit{transportation}, \textit{economy}, \textit{public safety and security}, \textit{social}, \textit{environment} and \textit{energy}. We explore the superiority of multi-modal data fusion methodologies in each type of downstream task.

\section{Data Perspective}
\label{SectionData}

This section delves into the datasets used in cross-domain data fusion in urban computing. Based on the literature available since 2015, we categorize diverse urban data and perform a statistical analysis of their distributions. Furthermore, we discuss how each type of data is incorporated into different research and real-world scenarios.

\subsection{Overview}
Based on the characteristics of the data and their diverse sources from various domains, this survey categorizes datasets utilized in the field of cross-domain data fusion in urban computing into six segments, including geographical data, traffic data, social network data, demographic data, environment data, and other data (i.e., data that cannot be categorized into the aforementioned types, such as healthcare data). As illustrated in Figure \ref{fig:intro} and \ref{fig:taxonmy}, these categories are defined as follows:
\begin{itemize}[leftmargin=*]
    \item \textbf{Geographical data} refers to geographical information of specific locations on the Earth's surface, such as coordinates (i.e., latitude and longitude). This category extends to spatial attributes, including but not limited to topography, land use, and physical features.

    \item \textbf{Traffic data} encompasses information related to the movement of vehicles and pedestrians, including factors like traffic flow, congestion, speed, and road conditions.

    \item \textbf{Social media data} comprises user-generated content from online platforms, encompassing geo-tagged text, images, and videos, offering insights into user behaviors, sentiment, and emerging trends.

    \item \textbf{Demographic data} involves statistical information about populations, including characteristics such as age, gender, ethnicity, income, and education.

    \item \textbf{Environment data} incorporates information about the natural world, covering aspects like climate, air quality, biodiversity, and pollution levels.
\end{itemize}

\begin{figure}[htb]
    \centering
    \includegraphics[width=0.45\textwidth]{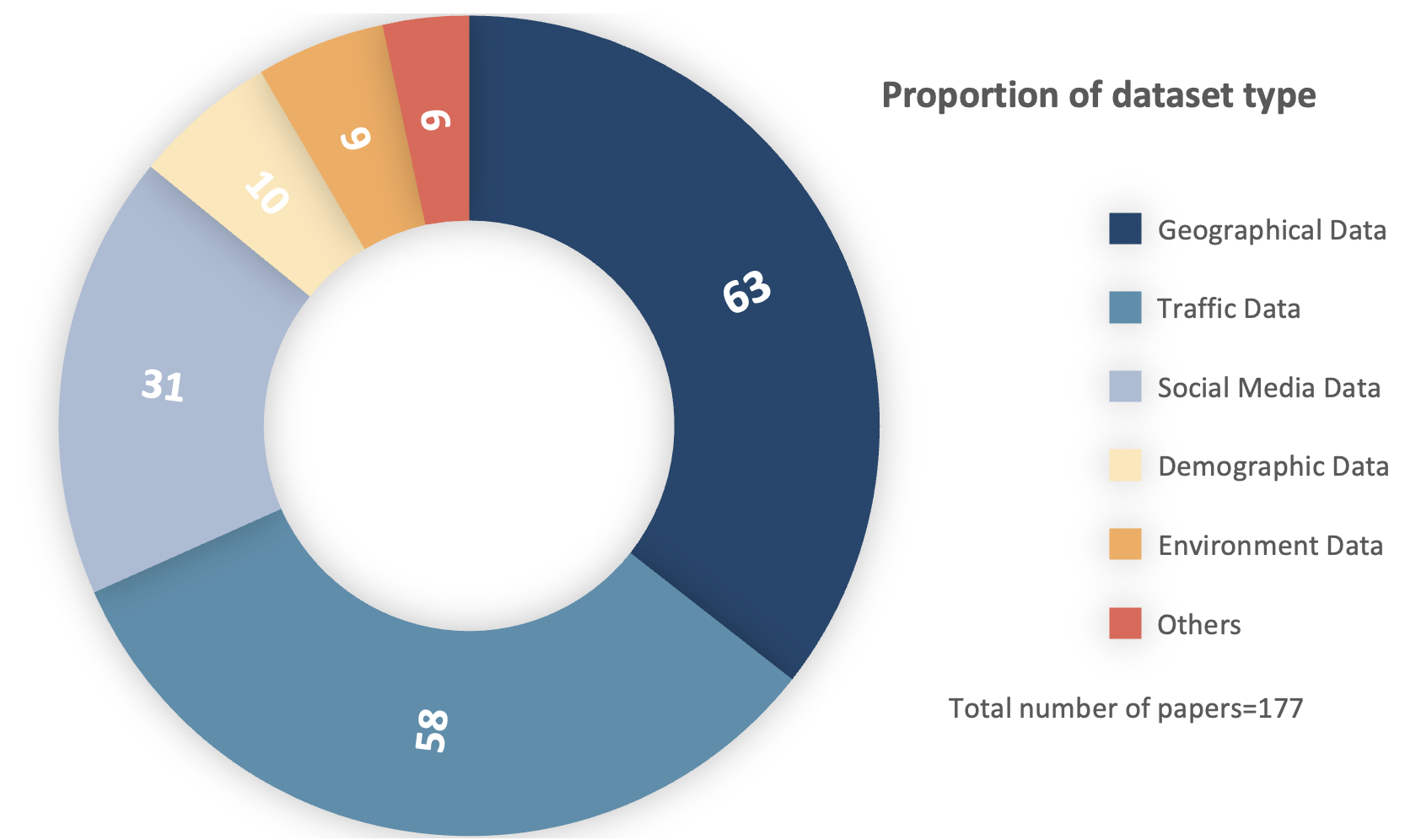}
    \vspace{-1em}
    \caption{Proportion of dataset type among highly related papers within the scope of cross-domain data fusion in urban computing.}
    \label{data1}
\end{figure} 

Based on the aforementioned categorization, Table \ref{tab:dataset-table} summarized open-sourced datasets commonly used for cross-domain data fusion in urban computing. Figure \ref{data1} presents the distribution of dataset modalities across all investigated papers in this survey. From the pie chart, it is evident that geographical and transportation data are the most crucial datasets in urban computing, with a majority of papers (approaching 70\%) opting for them as a primary modality. Following closely is social media data, which plays a significant role in urban computing, often combined with other modal datasets to address the complex and dynamic challenges of urban environments. Demographic data and environment data are also common datasets, however, only around 11\% of studies choose to incorporate these two types of data in their research.

\begin{figure}[htb]
    \centering
    \includegraphics[width=0.45\textwidth]{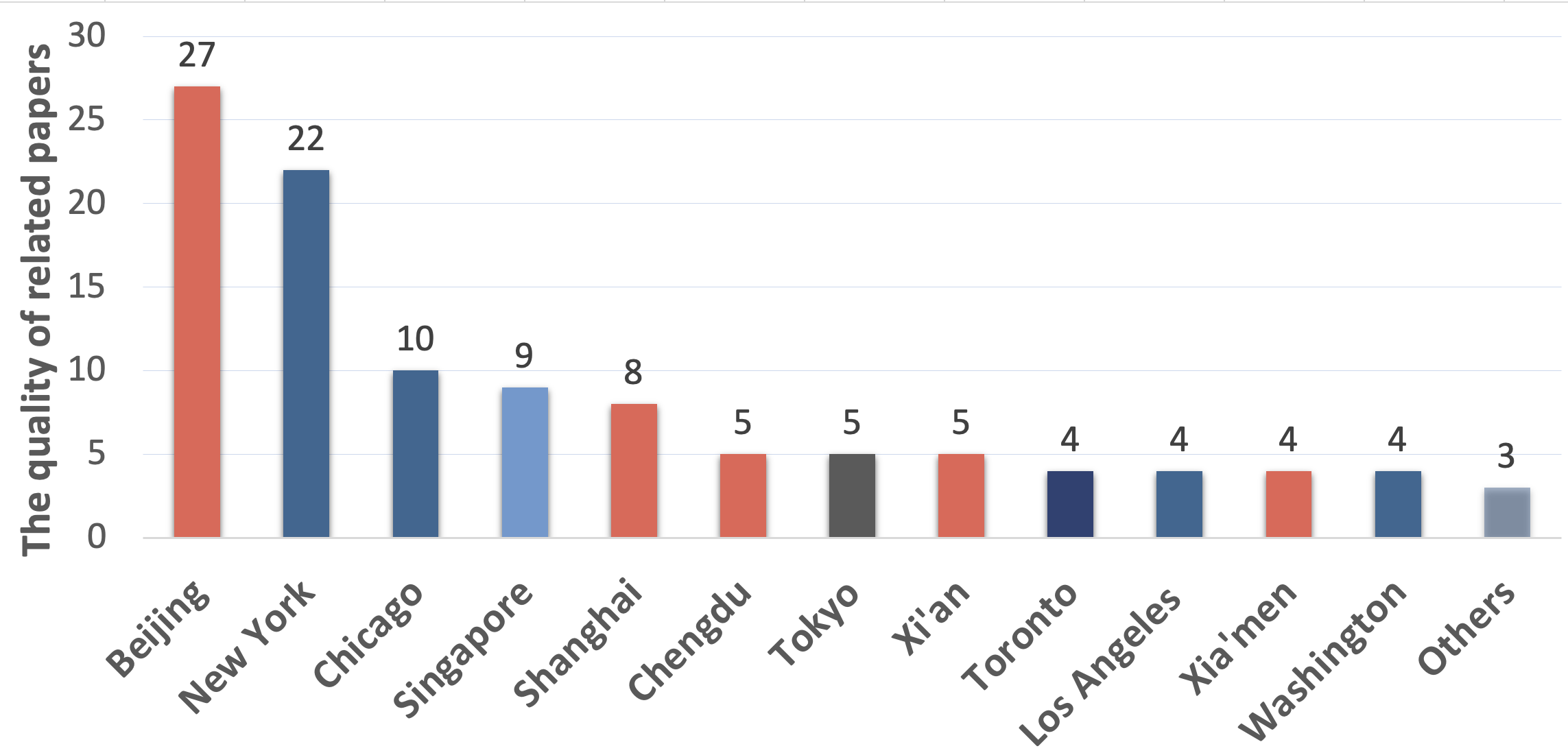}
    \vspace{-1em}
    \caption{Distribution of dataset usage frequency from different cities (i.e., bars) and countries (i.e., colors) across highly related papers within the scope of this survey. Note that the cities with less than four paper usage are omitted in this illustration for simplicity.}
    \label{data2}
\end{figure} 
Furthermore, we also investigate the geographic distribution of the datasets across various cities and countries. As depicted in Figure \ref{data2}, the bar chart indicates the frequency of dataset utilization in different cities (i.e., bars) and countries (i.e., colors). Distinctive colors of the bars denote different countries, indicating the popularity of the data focus in certain countries. Notably, datasets from Beijing and New York emerged as extensively utilized, followed closely by cities such as Chicago, Singapore, and Shanghai. Overall, the majority of datasets in the domain of cross-domain data fusion in urban computing originate from China and the United States.

\begin{table*}[htbp]
\caption{Taxonomy and summary of open-sourced dataset used for cross-domain data fusion in urban computing.}
\resizebox{\textwidth}{!}{%
\begin{tabular}{ccccll}
\toprule[2.0pt]
 &
   &
   &
   &
   \\
\multirow{-2}{*}{\textbf{Category}} &
  \multirow{-2}{*}{\textbf{Content}} &
  \multirow{-2}{*}{\textbf{Format}} &

  \multirow{-2}{*}{\textbf{Dataset}} &
  \multirow{-2}{*}{\textbf{Link}} &
  \multirow{-2}{*}{\textbf{Reference}} \\ \midrule[1.5pt]
 & 
&
   &
  ArcGIS &
  {\color[HTML]{2E86C1} \url{ https://developers.arcgis.com}} &
  \cite{liu2023knowledge} \\
 & &
   &
  PlanetScope&
  {\color[HTML]{2E86C1} \url{ https://developers.planet.com/docs/data/planetscope/}} &
  \cite{li2022predicting} \\
 & &
   &
  Google Earth  &
  {\color[HTML]{2E86C1} \url{ https://developers.google.com/maps/documentation/}} &
  \begin{tabular}[c]{@{}l@{}}\cite{jenkinsUnsupervisedRepresentationLearning2019}\end{tabular} \\
 & &
   &
  OpenStreetMap &
  {\color[HTML]{2E86C1} \url{ https://www.openstreetmap.org/}} &
  \cite{yin2023multimodal} \\
 &
  \multirow{-5}{*}{Satellite Image} &
  \multirow{-5}{*}{Image} &
  Baidu Maps &
  {\color[HTML]{2E86C1} \url{ https://lbsyun.baidu.com}} &
  \begin{tabular}[c]{@{}l@{}}\cite{yangDuAREAutomaticRoad2022,xiaoContextualMasterslaveFramework2023}\end{tabular} \\ \cmidrule{2-6} 
 & &
   &
  Baidu Map  &
  {\color[HTML]{2E86C1} \url{ https://lbsyun.baidu.com}} &
  \begin{tabular}[c]{@{}l@{}}\cite{liu2023knowledge,jiangITVInferringTraffic2021}\end{tabular} \\
 & &
   &
  Google Street &
  {\color[HTML]{2E86C1} \url{ https://developers.google.com/maps/}} &
  \begin{tabular}[c]{@{}l@{}}\cite{liu2023knowledge,alfarrarjeh2021exploring}\end{tabular} \\
 &
  \multirow{-3}{*}{\makecell[c]{Street-View \\Image}} &
  \multirow{-3}{*}{Image}&

  Tencent Map &
  {\color[HTML]{2E86C1} \url{ https://lbs.qq.com/tool/streetview/index.html}} &
  \cite{huang2023comprehensive}\\ \cmidrule{2-6} 
 & &
   &
  Tencent Map Service &
  {\color[HTML]{2E86C1} \url{ https://lbs.qq.com/getPoint/}} &
  \begin{tabular}[c]{@{}l@{}}\cite{xiFirstLawGeography2022b,ruan2022service}\end{tabular} \\
& &
   &
  WeChat POIs &
  {\color[HTML]{2E86C1} \url{ https://open.weixin.qq.com}} &
 \cite{wangSpatiotemporalUrbanKnowledge2021} \\
& &
   &
  Baidu Map POIs &
  {\color[HTML]{2E86C1} \url{ https://lbsyun.baidu.com}} &
  \begin{tabular}[c]{@{}l@{}}\cite{li2022predicting,liu2023characterizing,liuJointRepresentationLearning2019,huang2022ernie,xiaoContextualMasterslaveFramework2023}\end{tabular} \\
& &
   &
  NYC Open POIs &
  {\color[HTML]{2E86C1} \url{ https://opendata.cityofnewyork.us/}} &
  \begin{tabular}[c]{@{}l@{}}\cite{lin2019deepstn+,wang2021gsnet,bingPretrainedSemanticEmbeddings2023,zhangRegionEmbeddingIntra2023,wangTrafficAccidentRisk2023}\end{tabular} \\
& &
   &
  Foursquare &
  {\color[HTML]{2E86C1} \url{ https://developer.foursquare.com/docs/checkins/checkins}} &
  \begin{tabular}[c]{@{}l@{}}\cite{bingPretrainedSemanticEmbeddings2023,zhaoAnnotatingPointsInterest2016a,balsebreGeospatialEntityResolution2022,chenInformationCoverageLocation2015a,huangExploitingSpatialtemporalsocialConstraints2016b,jenkinsUnsupervisedRepresentationLearning2019}\end{tabular} \\
& &
   &
  Wikipedia POIs &
  {\color[HTML]{2E86C1} \url{ https://www.wikipedia.org}} &
  \cite{zhao2017photo2trip} \\
& &
   &
  AMap Service&
  {\color[HTML]{2E86C1} \url{ https://lbs.amap.com}} &
  \cite{bai2023geographic} \\
& &
   &
  Yelp POIs &
  {\color[HTML]{2E86C1} \url{ https://www.yelp.com/developers}} &
  \begin{tabular}[c]{@{}l@{}}\cite{balsebreGeospatialEntityResolution2022,zhao2020pgeotopic,zhao2016towards}\end{tabular} \\
& &
   &
  {Dianping POIs} &
  {\color[HTML]{2E86C1} \url{ https://api.dianping.com/}} &
  \begin{tabular}[c]{@{}l@{}}\cite{chandraNodeSense2VecSpatiotemporalContextaware2021,duGeofirstLawLearning2019}\end{tabular} \\
& &
   &
  Weibo POIs &
  {\color[HTML]{2E86C1} \url{ https://open.weibo.com/wiki/API}} &
  \begin{tabular}[c]{@{}l@{}}\cite{chandraNodeSense2VecSpatiotemporalContextaware2021,keerthi2020collective,gao2023dual}\end{tabular} \\
& &
   &
  Flickr POIs &
  {\color[HTML]{2E86C1} \url{ https://www.flickr.com/services/developer/api/}} &
  \cite{he2019joint}\\
\multirow{-20}{*}{\textbf{\makecell[c]{Geographical \\Data}}} & 
  \multirow{-13}{*}{POIs} &
   \multirow{-13}{*}{\makecell[c]{Point Vector}} &
  Bing Map POIs &
  {\color[HTML]{2E86C1} \url{ https://www.bingmapsportal.com}} &
  \begin{tabular}[c]{@{}l@{}}\cite{chenCrosscityFederatedTransfer2022}\end{tabular} \\ \midrule


 &&
   &
  Shenzhou UCar &
  {\color[HTML]{2E86C1} \url{ https://bit.ly/2MG47xz}} &
  \cite{guo2020force}\\
 &&

   &
  Chicago Transportation&
  {\color[HTML]{2E86C1} \url{ https://data.cityofchicago.org/}} &
  \begin{tabular}[c]{@{}l@{}}\cite{wang2021gsnet,wangTrafficAccidentRisk2023,jenkinsUnsupervisedRepresentationLearning2019}\end{tabular} \\
   &
  &
  &
  VED &
  {\color[HTML]{2E86C1} \url{ https://github.com/gsoh/VED}} &
  \cite{Oh2019VehicleED,zhang2022extended} \\ 
 &&
   &
  Taxi Shenzhen &
  {\color[HTML]{2E86C1} \url{ https://github.com/cbdog94/STL}} &
  \begin{tabular}[c]{@{}l@{}}\cite{huangComprehensiveUrbanSpace2023a,wuMiningSpatiotemporalReachable2017}\end{tabular} \\
 &&
   &
  NYC Open Taxi Data&
  {\color[HTML]{2E86C1} \url{ https://opendata.cityofnewyork.us/how-to/}} &
  \begin{tabular}[c]{@{}l@{}}\cite{zhangMultiviewJointGraph2021a,zhangRegionEmbeddingIntra2023}\end{tabular} \\
   &&
   &
  GeoLife&
  {\color[HTML]{2E86C1} \url{ http://urban-computing.com/index-893.htm}} &
  \begin{tabular}[c]{@{}l@{}}\cite{han2021graph,zheng2010geolife,zheng2009mining,zheng2008understanding,yu2015personalized}\end{tabular} \\
   &&
   &
  T-Drive Taxi&
  {\color[HTML]{2E86C1} \url{ http://urban-computing.com/index-58.htm}} &
  \begin{tabular}[c]{@{}l@{}}\cite{yuan2011driving,yuan2010t,pan2019urban,lv2018lc}\end{tabular} \\
   &
  &
 &
  DiDi Traffic&
  {\color[HTML]{2E86C1} \url{ https://outreach.didichuxing.com/research/opendata/}} &
  \cite{yuan2021effective,luoLetTrajectoriesSpeak2021,rajehModelingMultiregionalTemporal2023,yao2022trajgat,tedjopurnomo2021similar} \\ 
 &&
   &
  Xiamen Taxi &
  {\color[HTML]{2E86C1} \url{ https://data.mendeley.com/datasets/6xg39x9vgd/1}} &
  \begin{tabular}[c]{@{}l@{}}\cite{you2022panda,chenUVLensUrbanVillage2021,jiangITVInferringTraffic2021,chen2018radar}\end{tabular} \\

   &
     \multirow{-8}{*}{ \makecell[c]{Traffic \\Trajectory}} &
  \multirow{-8}{*}{\makecell[c]{Spatio-temporal\\Trajectory}} &
  Grab-Posisi &
  {\color[HTML]{2E86C1} \url{ https://goo.su/W3yD5m}} &
  \begin{tabular}[c]{@{}l@{}}\cite{yin2023multimodal,yinMultitaskLearningFramework2020a}\end{tabular} 


  \\ \cmidrule{2-6}

 &&
   &
  California-PEMS &
  {\color[HTML]{2E86C1} \url{ http://pems.dot.ca.gov}} &
  \cite{arslan2009grvs,sun2023battery}\\
 &&
   &
  METR-LA &
  {\color[HTML]{2E86C1} \url{ https://www.metro.net}} &
  \cite{lablack2023spatio,liu2023spatio}\\
 &&
   &
  Large-ST &
  {\color[HTML]{2E86C1} \url{ https://github.com/liuxu77/LargeST}} &
  \cite{liu2023largest}\\
  
 &
  &
  &
  MobileBJ &
  {\color[HTML]{2E86C1} \url{https://github.com/FIBLAB/DeepSTN/issues/4 }} &
  \cite{lin2019deepstn+, keerthi2020collective, chandraNodeSense2VecSpatiotemporalContextaware2021} \\

   &
  &
 &
  TaxiBJ &
  {\color[HTML]{2E86C1} \url{ https://goo.su/aQyjTAz}} &\cite{liang2021fine-grained,bai2019stg2seq,qu2022forecasting,jiSelfsupervisedSpatiotemporalGraph2023,zhangMultiviewJointGraph2021a,fu2019efficient}\\

   &
      \multirow{-6}{*}{Taffic Flow}&
    \multirow{-6}{*}{\makecell[c]{Spatio-temporal\\ Graph}}&
  BikeNYC &
  {\color[HTML]{2E86C1} \url{ https://citibikenyc.com/}} &
  \cite{lin2019deepstn+,bai2019stg2seq,qu2022forecasting,jiSelfsupervisedSpatiotemporalGraph2023} \\ 

  \cmidrule{2-6}

  &
  \multirow{2}{*}{Road Network} &
  \multirow{2}{*}{Spatial Graph}&
   \makecell[c]{OpenStreetMap}   &
  {\color[HTML]{2E86C1} \url{ https://www.openstreetmap.org}} &
  \cite{yinMultitaskLearningFramework2020a,balsebreGeospatialEntityResolution2022,luoLetTrajectoriesSpeak2021,yuan2021effective,gengMultimodalGraphInteraction2019} \\ 
  
 &&
   &
  US Census Bureau &
  {\color[HTML]{2E86C1} \url{ https://www.census.gov/data.html}} &
  \cite{zhangRegionEmbeddingIntra2023} \\ \cmidrule{2-6}


  \multirow{-17}{*}{\textbf{\makecell[c]{Traffic \\Data}}} 
  &
 &
  &
  LaDe &
  {\color[HTML]{2E86C1} \url{ https://cainiaotechai.github.io/LaDe-website/}} &
  \cite{wu2023lade} \\

  &
 \multirow{-2}{*}{Logistics}&
  \multirow{-2}{*}{\makecell[c]{Spatio-temporal\\Trajectory}}&
  JD Logistics &
  {\color[HTML]{2E86C1} \url{ https://corporate.jd.com/ourBusiness\#jdLogistics}} &
  \cite{ruan2022service} \\  \midrule

   &&
 
     &
    \multirow{2}{*}{\makecell[c]{Twitter}} &
  \multirow{2}{*}{{\color[HTML]{2E86C1} \url{ https://developer.twitter.com/en/docs}} }&
    \begin{tabular}[c]{@{}l@{}}\cite{bingPretrainedSemanticEmbeddings2023,zhaoAnnotatingPointsInterest2016a,zhao2016towards,yuan2017pred,vu2016geosocialbound,wu2015semantic,shen2018forecasting}\end{tabular} \\
  &&
     &
      &
        &
    \begin{tabular}[c]{@{}l@{}}\cite{wang2017computing,wang2016estimating,wang2016enhancing,wang2015citywide,miyazawa2019integrating,liu2020spatiotemporal}\end{tabular} \\
   &&
     &
    \multicolumn{1}{c}{Common Crawl} &
    {\color[HTML]{2E86C1} \url{ https://registry.opendata.aws/commoncrawl/}} &
    \cite{zhao2020pgeotopic} \\
   &&
     &
    \multicolumn{1}{c}{Yelp Reviews} &
    {\color[HTML]{2E86C1} \url{ https://www.yelp.com/dataset}} &
    \begin{tabular}[c]{@{}l@{}}\cite{zhao2020pgeotopic,zhao2016towards}\end{tabular} \\
   &
    \multirow{-4}{*}{Text} &
    \multirow{-4}{*}{Text} &
    \multicolumn{1}{c}{Weibo Traffic Police} &
    {\color[HTML]{2E86C1} \url{http://open.weibo.com/developers/}} &
    \cite{you2022panda} \\ \cmidrule{2-6} 

   &&
     &
    \multicolumn{1}{c}{\multirow{-1}{*}{YFCC100M}} &
    \multirow{-1}{*}{{\color[HTML]{2E86C1} \url{ https://goo.su/jzaDU}}} &
    \multirow{-1}{*}{\begin{tabular}[c]{@{}l@{}}\cite{zhao2017photo2trip,yin2021learning,he2019joint}\end{tabular}} \\

   &&
     &
    \multicolumn{1}{c}{\multirow{-1}{*}{NUS-WIDE}} &
    \multirow{-1}{*}{{\color[HTML]{2E86C1} \url{ https://goo.su/dWPQZcD}}} &
    \multirow{-1}{*}{\begin{tabular}[c]{@{}l@{}}\cite{yin2021learning,yin2019gps2vec}\end{tabular}} \\
   &
    \multirow{-3}{*}{\makecell[c]{Geo-tagged \\Image \& Video}} &
    \multirow{-3}{*}{Image\&Video} &
    \multicolumn{1}{c}{GeoUGV} &
    {\color[HTML]{2E86C1} \url{ https://qualinet.github.io/databases/video/}} &
    \cite{lu2016geougv} \\ \cmidrule{2-6} 
   &&
     &
    \multicolumn{1}{c}{Jiepang User Check-in} &
    {\color[HTML]{2E86C1} \url{ https://jiepang.app/}} &
    \cite{fu2019efficient} \\
   &&
     &
    \multicolumn{1}{c}{Gowalla User Location} &
    {\color[HTML]{2E86C1} \url{ http://konect.cc/networks/loc-gowalla_edges/}} &
    \begin{tabular}[c]{@{}l@{}}\cite{chenInformationCoverageLocation2015a,yuan2017pred}\end{tabular} \\
  \multirow{-10}{*}{\textbf{\makecell[c]{Social\\Media\\ Data}}} &
    \multirow{-3}{*}{Users’ Info} &
        \multirow{-3}{*}{Time Series} &
    \multicolumn{1}{c}{WeChat Mobility} &
     {\color[HTML]{2E86C1} \url{ https://open.weixin.qq.com/}} &
    \cite{wangSpatiotemporalUrbanKnowledge2021} \\ \midrule
&
  Crime&
  Time Series&
  NYC Crime
  &{\color[HTML]{2E86C1} \url{ https://opendata.cityofnewyork.us/}} &
  \cite{zhangMultiviewJointGraph2021a} \\ \cmidrule{2-6} 
&&
  &
  Land Use SG
  &{\color[HTML]{2E86C1} \url{ https://www.ura.gov.sg/Corporate/Planning/Master-Plan}} &
  \cite{liUrbanRegionRepresentation2023} \\
&\multirow{-2}{*}{Land Use}
  &
  \multirow{-2}{*}{Time Series}
  &Land Use NYC
  &{\color[HTML]{2E86C1} \url{ https://goo.su/puTuG}} &
  \cite{liUrbanRegionRepresentation2023} \\
  \cmidrule{2-6} 
\multirow{-4}{*}{\textbf{\makecell[c]{Demographic\\ Data}}} &
  Population &
  Time Series &
  WorldPop&
  {\color[HTML]{2E86C1} \url{ https://www.worldpop.org/}} &
  \begin{tabular}[c]{@{}l@{}}\cite{xiFirstLawGeography2022b,li2022predicting,bai2023geographic}\end{tabular} \\ \midrule
 &&
   &
  TipDM China Weather &
  {\color[HTML]{2E86C1} \url{ https://www.tipdm.org/}} &
  \cite{liu2022symbolic} \\
 &&
   &
  DarkSky Weather &
  {\color[HTML]{2E86C1} \url{ https://support.apple.com/en-us/102594}} &
  \cite{yuan2021effective} \\
 &&
   &
  WeatherNY &
  {\color[HTML]{2E86C1} \url{ https://opendata.cityofnewyork.us/}} &
  \cite{wang2021gsnet}\\
 &&
   &
  WeatherChicago &
  {\color[HTML]{2E86C1} \url{ https://data.cityofchicago.org/}} &
  \cite{wang2021gsnet}\\
 &&
   &
  Weather Underground &
  {\color[HTML]{2E86C1} \url{ https://www.wunderground.com/}} &
  \cite{you2022panda}\\

 &&
   &
  \makecell[c]{DidiSY} &
  {\color[HTML]{2E86C1} \url{ https://www.didiglobal.com/}} &
  \cite{bai2019spatio} \\
 &&
   &
  WD\_BJ weather &
  {\color[HTML]{2E86C1} \url{ https://goo.su/DmHFHd}} &
  \cite{ma2023histgnn}\\
 &
 \multirow{-8}{*}{Meteorology}  &
 \multirow{-8}{*}{Time Series}&
  WD\_USA weather &
  {\color[HTML]{2E86C1} \url{ https://goo.su/RVhBA}} &
  \cite{ma2023histgnn}\\
  \cmidrule{2-6}
 &
  Greenery &
  Time Series &
  Google Earth&
  {\color[HTML]{2E86C1} \url{ https://earth.google.com/}} &
  \cite{you2022panda} \\
   \cmidrule{2-6}
\multirow{-8}{*}{\textbf{\makecell[c]{Environment\\ Data}}} &
  \multirow{2}{*}{Air Quality} &
  \multirow{2}{*}{Time Series}&
  UrbanAir &
{\color[HTML]{2E86C1} \url{https://goo.su/hfzNB53}} &
  \cite{zheng2015Forecasting,zheng2013u,zheng2014cloud}\\ 
  &&
   &
  KnowAir &
  {\color[HTML]{2E86C1} \url{ https://github.com/shuowang-ai/PM2.5-GNN}} &
  \cite{wang2020pm2,yu2021spatial,zhang2022deep,xu2023dynamic}\\

  \bottomrule[2.0pt]
\end{tabular}%
}

\label{tab:dataset-table}
\end{table*}

\subsection{Geographical Data}
\begin{figure}[htbp]
    \centering
    \vspace{-1em}
    \includegraphics[width=0.46\textwidth]{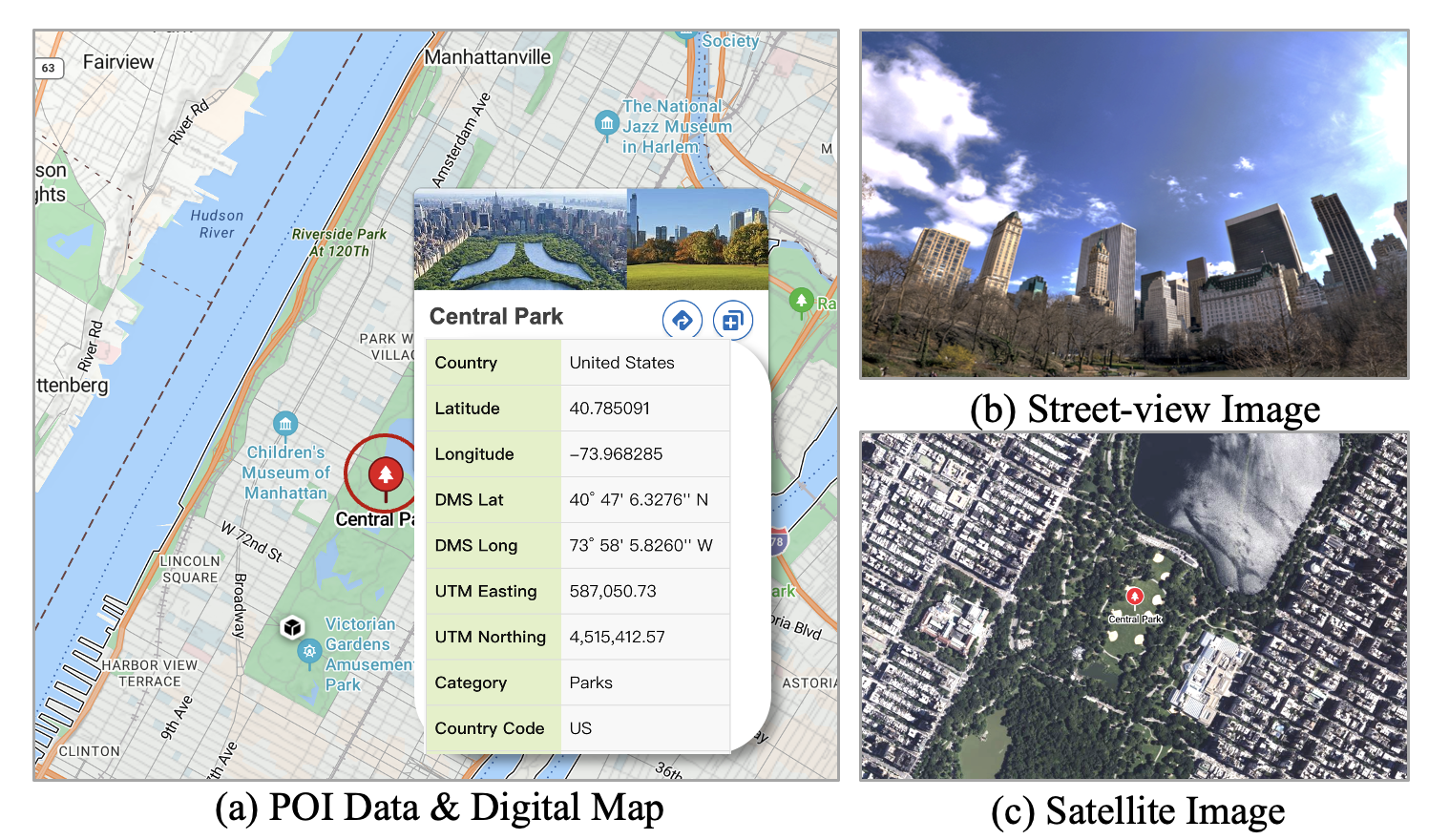}
    \vspace{-1em}
    \caption{Visualization of four types of geographical data collected at Central Park, New York, USA: (a) POI data and digital map; (b) street-view image; (c) satellite image.}.
    \label{fig:geodata}
\end{figure}

Geographical data plays a crucial role in modeling spatial relationships, contributing to the enhancement of cross-domain data fusion in urban computing by providing valuable insights into the geospatial context \cite{breunig2020geospatial,zhao2021deep,zhang2019functional}. Tobler's First Law of Geography, as stated by \citet{miller2004tobler}, indicates that "Everything is related to everything else, but near things are more related than distant things." This emphasizes the importance of geographical data, which serve as the basis for spatial modeling-based urban computing research.  Figure \ref{fig:geodata} shows the main four types of geographical data through a visualization approach.

\textbf{Points of Interest (POI) data}, as a cornerstone of cross-domain data fusion in urban computing, command significant attention and utility in the realm of geographical data \cite{lin2019deepstn+,wang2021gsnet,zhao2017photo2trip,balsebreGeospatialEntityResolution2022,fangTopkPOIRecommendation2023,caoPointsofinterestRecommendationAlgorithm2020,gao2022contextual,huangExploitingSpatialtemporalsocialConstraints2016b}.
POI data refers to a collection of data representing specific locations or sites that hold significance or interest, often encompassing businesses, landmarks, or other notable entities in a given geographical area  \cite{psyllidis2022points,liu2020urban}. Conventionally, POI data may include the location coordinates (i.e., longitude and latitude), address, categorization (e.g., restaurants, hotels, parks, etc.), address, phone number, operating hours, and so on. Recent cross-domain urban datasets may also contain user reviews, photos, and other individual information \cite{zhao2020pgeotopic,chandraNodeSense2VecSpatiotemporalContextaware2021,duGeofirstLawLearning2019,balsebreGeospatialEntityResolution2022}. Therefore, POI data is capable of representing the semantics of a specific area, encapsulating key locations and entities to provide a comprehensive understanding of the geographic context. For example, \citet{bingPretrainedSemanticEmbeddings2023} collected POI data of two cities: New York City and Tokyo from the Foursquare platform and observed that POIs with higher spatial similarity values often have similar semantics. Through AMaps Service Platform in Wuhan and Shanghai, \citet{bai2023geographic} collected approximately 2 million POI records and fused them with satellite visual embedding for socioeconomic downstream tasks. To model the spatial correlations between POIs,  \citet{fu2019efficient} developed a spatial network based on the multi-relation data among POIs, such as spatial distance and mobility connectivity.

\textbf{Satellite image data} holds significant importance in representing geo-context by providing a visual depiction of Earth's surface,  and offering advantages such as global coverage, real-time monitoring, and the ability to capture fine-scale details \cite{burke2021using,yuan2021review,neupane2021deep}. Satellite imagery is readily accessible and amenable to processing through various publicly available platforms (e.g., Google Earth \cite{jenkinsUnsupervisedRepresentationLearning2019}, OpenStreetMap \cite{yin2023multimodal}, Baidu Map \cite{yangDuAREAutomaticRoad2022,xiaoContextualMasterslaveFramework2023}, and PlanetScope \cite{li2022predicting}), facilitating multi-modal research in urban computing domain. 

Moreover, \textbf{street-view image data} plays a vital role in depicting geo-contextual information by providing immersive, ground-level perspectives \cite{biljecki2021street,gong2019classifying,kang2020review}. A significant amount of cross-domain urban research has collected data from public platforms such as Google Street \cite{liu2023knowledge,alfarrarjeh2021exploring}, Baidu Map \cite{liu2023knowledge,jiangITVInferringTraffic2021}, and Tencent Map \cite{huangComprehensiveUrbanSpace2023a}.

\subsection{Traffic Data}
\begin{figure}[!h]
    \centering
    \includegraphics[width=0.48\textwidth]{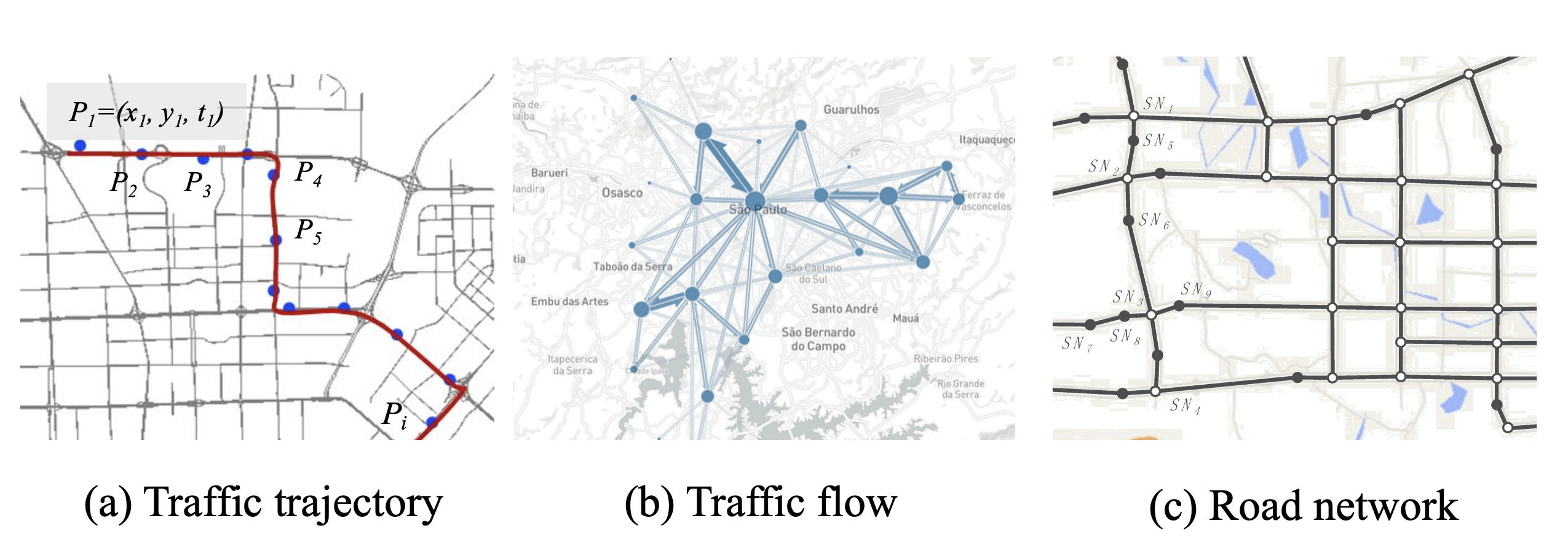}
    \vspace{-1em}
    \caption{Visualization of three primary types of traffic data: (a) traffic trajectory; (b) traffic flow; (c) road network \cite{wu2016updating}.}
    \label{fig:trafficdata}
\end{figure}

Traffic data accounts for the second largest proportion in Figure \ref{data1}. Different from geographical data, traffic data is generated through human activities and is directly associated with socio-economic factors, making it a distinct data type with significant implications.
Traffic data finds application in diverse downstream tasks, encompassing multiple domains within urban computing, which involve spatial and temporal dimensions, as well as dynamic and static aspects. Therefore, we classify traffic data into four distinct types, taking into account their characteristics and usage scenarios: trajectory data, traffic flow data, road network data, and other miscellaneous data. Figure \ref{fig:trafficdata} indicates the visualization of the first three types of traffic data.

\textbf{Human trajectory data} can be conceptualized as a sequential trace produced by an object in motion within geographical spaces. This trajectory is typically represented as an array of chronologically sequenced points, delineated as \( p_1 \rightarrow p_2 \rightarrow \cdots \rightarrow p_n \). Each point in this sequence comprises a set of geospatial coordinates coupled with a corresponding timestamp, formalized as
\begin{equation}\label{eqn:traj_definition}
     \mathbf{P} = (\mathbf{x}, \mathbf{y}, \mathbf{t}), 
\end{equation}
where \(\mathbf{x}\) and \(\mathbf{y}\) denote the spatial coordinates, typically representing longitude and latitude in a geodetic framework. \(\mathbf{t}\) represents the timestamp. This format facilitates the precise tracking and analysis of spatial dynamics over time.  


 Nowadays, location-based services, such as taxi and ride-hailing services, have generated enormous trajectory data. 
Additionally, many service providers, such as Uber and DiDi, have made their datasets publicly available to support research in this field.
For instance, \citet{zhangMultiviewJointGraph2021a} utilized taxi trip data during one month as a modality to assist in profiling urban regions.
PANDA \cite{you2022panda} demonstrated the significance of taxi trajectory in predicting road risk, which can aid in the extraction of road segments.
\citet{gengMultimodalGraphInteraction2019} forecasted ride-hailing demand on large-scale ride-hailing datasets in Beijing and Shanghai.

\textbf{Traffic flow data} provides critical insights into the movement patterns of vehicles and pedestrians within urban environments, essential for understanding and optimizing city dynamics. In the realm of urban computing, three datasets — \textit{MobileBJ}, \textit{BikeNYC}, and \textit{TaxiBJ} — emerge as key sources of insight for traffic mobility. Specifically, the \textit{MobileBJ} dataset, collected from a Chinese social network, has significantly contributed to three key areas of urban computing. Firstly, it enabled the development of DeepSTN+ \cite{lin2019deepstn+}, a deep learning model for predicting urban crowd flows, which integrates spatial dependencies, POIs, and temporal data. Additionally, it supported studies on high-dimensional spatio-temporal data in urban communities, employing advanced representation learning techniques \cite{keerthi2020collective,chandraNodeSense2VecSpatiotemporalContextaware2021}.

The \textit{BikeNYC} dataset, derived from New York City's bike-sharing system in 2014, provides rich data on trip durations, station locations, and times. It focuses on the last 14 days for testing purposes and includes 9 POI types. This dataset has been instrumental in developing a model for multi-step passenger demand forecasting \cite{bai2019stg2seq}, leveraging its detailed trip data to address complex spatio-temporal challenges in urban transportation planning \cite{jiSelfsupervisedSpatiotemporalGraph2023,qu2022forecasting,zhang2023adding}. Meanwhile, \textit{TaxiBJ} collects GPS traces from Beijing taxis, encompassing trip details, travel times, speeds, and specific pick-up and drop-off points. This dataset allowed for the development of a novel embedding strategy for urban regions \cite{fu2019efficient}, providing deeper insights into city dynamics and supporting sustainable urban development. It also enhanced traffic prediction accuracy by applying spatiotemporal graph neural networks and integrating self-supervised learning for improved long-term forecasting \cite{jiSelfsupervisedSpatiotemporalGraph2023}. Furthermore, \textit{TaxiBJ} demonstrated the effectiveness of a contrastive self-supervision method in inferring fine-grained urban flows, especially in environments with limited resources \cite{qu2022forecasting}. An innovative approach for computing trajectory similarities, \textit{TrajGAT}, was introduced using \textit{TaxiBJ}, significantly advancing the analysis of long trajectory data \cite{yao2022trajgat}.

\textbf{Road network data} is intimately related to human daily lives, as it serves as the foundation for various services such as navigation and food delivery.
The acquisition of road network data can be achieved through various methods.
The earliest on-site manual surveying was labor-intensive and required a large amount of resources. With the development of remote sensing technology, which can effectively reduce costs, the proportion of road network data drawing relying on on-site collection has been decreasing.
Another method to collect road network data is through UGC (User Generated Content), which collects road network information through anonymous terminal devices. The collected route can effectively help with updating road attributes and refinement of the road network.
Generally, road network data can be downloaded in shapefile format from different open-sourced platforms such as Open Street Map, GRIP global roads database, DIVA-GIS, etc.
For example, \citet{yuan2021effective} predicted travel demand and traffic flow based on taxi orders which is associated with trajectories on the road network.
\citet{zhu2020inferring} ranked region significance taking into account multi-source spatial data including trajectories on road networks.
Besides, the acquisition of street-view image datasets \cite{huang2023comprehensive,doi:10.1073/pnas.2220417120} also necessitated the sampling of collection points on road networks.

Other miscellaneous data include logistic data, transportation safety data, and transportation recommendation data. In the logistics field, LaDe \cite{wu2023lade} introduced the first industry-scale last-mile delivery dataset. LaDe includes detailed information about the courier trajectories and waybill information,  which can support massive spatio-temporal data mining tasks \cite{ruan2022service}. Transportation safety data is also of great significance. For example, in \cite{chen2018radar}, road obstacle data was used to develop RADAR, a real-time system for identifying road obstacles in urban areas during typhoon seasons. \citet{you2022panda} used a comprehensive event dataset, including data on road accidents, fallen trees, and ponding water, to successfully develop a framework for predicting road risks in post-disaster urban settings with high accuracy. Transportation recommendations have become an integral part of our daily lives, contributing significantly to various service enhancements. For instance, \citet{gaoDualgrainedHumanMobility2023b} utilized this data to develop GraphTrip, a groundbreaking framework that leverages spatio-temporal graph representation learning for trip recommendations. This framework was rigorously tested using datasets from Edinburgh, Glasgow, Osaka, Toronto, and Melbourne, showcasing notable improvements in travel planning accuracy. Similarly, \citet{guo2020force} and \citet{he2019joint} employed multi-source urban data, encompassing operational data, taxi GPS trajectories, and public transportation information, to create advanced recommendation models.

\subsection{Social Media Data}

Twitter, an online social media and social networking service allows registered users to post \textbf{geo-textual data} \cite{conger2023twitter}. Hence, the Tweet data, owing to its inherent geo-tagged information, is utilized by researchers as a modality representing user social states, and it is integrated with other models in multi-modal learning. For instance, \citet{zhaoAnnotatingPointsInterest2016a} collected English tweets using the Twitter API in two cities, New York City and Singapore, and then annotated whether a tweet was POI-related. Finally, they can model the association between the tweet and its most semantically related POI. To study periodic human mobility patterns, \citet{yuan2017pred} collected geo-annotated tweets from the most recent 3,200 tweets of Twitter users, and mapped them to the corresponding city by reverse geocoding. With a similar goal of understanding urban dynamics, \citet{miyazawa2019integrating} focused only on tweets regarding mobility and social activity; while \citet{wang2015citywide} are more interested in tweets concerning traffic events in Chicago. Furthermore, geo-textual data such as tweets can be used to learn user preferences to support personalized maps \cite{zhao2016towards}. Some work, such as \cite{vu2016geosocialbound} working on POI boundary estimation, also removed the content automatically created by other services like Twimight, Tweetbot, and so forth.

\textbf{Geo-tagged photos} are crucial in multi-modal learning as they provide spatial context, enriching the understanding of content by incorporating location information into the broader spectrum of data modalities \cite{nam2022realroi,anbalagan2022event,bui2023automatic}. The two most commonly used geo-tagged photo datasets are the Yahoo Flickr Creative Commons (YFCC) dataset \cite{thomee2015new} as well as the NUS-WIDE dataset \cite{chua2009nuswide}. It is noteworthy that the YFCC dataset is well-known as the largest public multimedia collection released, which consists of 100 million photos posted on Flickr with relevant meta information such as geo-location coordinates and the date taken. The NUS-WIDE dataset is a well-known web image dataset (with 269,648 images and the associated tags from Flickr) created by Lab for Media Search in the National University of Singapore. Both \citet{zhao2017photo2trip} and \citet{he2019joint} leveraged the geo-tagged photos to jointly learn a context-aware embedding for personalized tour recommendation. \citet{yin2019gps2vec} derived training labels from the geo-tagged documents from the NUS-WIDE dataset, and therefore generated GPS embeddings. In the work of \citet{yin2021learning} where multi context-aware location representations need to be learned, the geo-tagged photos from the NUS-WIDE dataset are used for image classification evaluation, whereas those from the YFCC dataset are leveraged to learn the semantic context.

With the ubiquity of sensor-rich smartphones, acquiring continuous video frames with spatial metadata has become practical. The public \textbf{geo-tagged mobile video data} comes mainly from two mobile platforms, MediaQ \cite{kim2014mediaq,mediaq} and GeoVid \cite{arslan2009grvs,geovid}. \citet{lu2016geougv} collected the geo-tagged video data from the aforementioned data platforms, and proposed the GeoUGV dataset consisting of two sets, videos and their geospatial metadata. Moreover, many social media platforms (e.g., WeChat \cite{wangSpatiotemporalUrbanKnowledge2021}, Gowalla \cite{chenInformationCoverageLocation2015a,yuan2017pred}, Baidu \cite{liu2023characterizing}, Jiepang \cite{fu2019efficient}) encompass valuable user information, serving as a fundamental modality for fusion.

\subsection{Demographic Data}
The inclusion of demographic datasets in spatio-temporal multi-modal learning is pivotal as it enhances the contextual understanding of a given human group \cite{song2019monitoring,wallin2019prevalence,stratton2021population}. The WorldPop organization plays a crucial role in global demographic research by providing high-resolution \textbf{population data}, enabling informed decision-making, and addressing various socio-economic and public health challenges worldwide \cite{worldpop1,worldpop2,worldpop3,worldpop4}. \citet{xiFirstLawGeography2022b} and \citet{li2022predicting} collected Beijing population and population density statistics from the WorldPop platform as one of the predicted socioeconomic indicators; whereas \citet{liUrbanRegionRepresentation2023} extracted those from Singapore and New York City. In terms of data post-processing, \citet{bai2023geographic} further estimated the population density per grid cell through the Random Forest-based redistribution method.

\textbf{Crime data} holds immense significance as it serves as a critical resource for understanding patterns and factors influencing criminal activities, enabling policymakers and researchers to develop effective strategies for crime prevention and public safety \cite{huang2018deepcrime,hajela2021multi,zhang2022interpretable}. For example, \citet{zhangMultiviewJointGraph2021a} collected crime data from the NYC Open Data website as ground truth values to be predicted by region embeddings, and there are 40 thousand crime records during one year in New York City. 

Besides, \textbf{land use data} is crucial in urban planning, providing valuable insights into the spatial distribution of human activities, and informing decision-makers to optimize land resources for various purposes \cite{steurer2020measuring,chen2021mapping,risal2020sensitivity}. It consists of property data (e.g., private residential property transactions), residential data (e.g., buying and renting properties), business data (e.g., renewal of business use), and so on. For instance, \citet{liUrbanRegionRepresentation2023} collected land use data of Singapore and New York City from Singapore Master Plan 2019 and NYC MapPLUTO, respectively, as a region representation evaluation benchmark.

\subsection{Environment Data}
Environment data, particularly \textbf{meteorological data}, offers essential insights into dynamic weather patterns and environmental conditions that are integral for understanding complex interactions in various urban domains \cite{he2020first,anderegg2020climate,perera2020quantifying,trisos2020projected}. The publicly available APIs can be utilized to gather meteorological data. For example, \citet{yuan2021effective} utilized the Dark Sky API integrated into Apple Weather to extract diverse weather characteristics for each region and used one-hot encoding to represent categorical attributes. \citet{you2022panda} extracted temporal contextual features including rainfall, temperature, humidity, dew point, and wind speed using Weather Underground API. \citet{wang2021gsnet} obtained temperature and sky condition data for both NYC and Chicago from NYC Open Data and Chicago Data Portal, respectively. By doing so, they were able to analyze the spatio-temporal correlations to predict the risk of traffic accidents. To precisely forecast the weather, \citet{ma2023histgnn} employed three real-world weather datasets:  the WD\_BJ dataset, which was gathered from 10 ground automatic weather stations in Beijing and contains nine meteorological variables; and the WD\_ISR dataset and WD\_USA dataset, both collected from OpenWeather and providing information on four weather conditions (namely temperature, humidity, wind speed, and atmospheric pressure) for Israel and the USA, respectively. In addition, enterprises also gather their own meteorological data, such as the DidiSY dataset which includes information on weather conditions, temperature, and wind speed specifically in Shenyang, a major city in China \cite{bai2019spatio}. Besides, open-source datasets from competitions can also be utilized, such as the Chinese weather dataset from the 7th Teddy Cup Data Mining Challenge, which includes 12 different types of weather \cite{liu2022symbolic}.

\textbf{Greenery data} holds significance in sustainable urban planning, providing valuable information about the distribution and density of vegetation, which is essential for assessing biodiversity, urban green spaces, and their impact on overall ecological health \cite{ye2019daily,chen2020quantifying,nourmohammadi2021mapping}. In particular, \citet{you2022panda} extracted the degree of tree coverage using AlexNet \cite{krizhevsky2012imagenet} based on satellite imagery obtained from the Google Earth platform. \textbf{Air quality data} is of paramount importance for environmental management, facilitating the identification of sources, and enabling the development of effective strategies to mitigate the adverse impacts of air pollution on both human health and the environment \cite{zhang2022linking,tang2022air,carozzi2023dirty}. For example, the UrbanAir system offers air quality data every hour from 2,296 stations in 302 Chinese cities, where each record consists of the concentration of six pollutants: $NO_2$, S$O_2$, $O_3$, $CO$, $PM2.5$ and $PM10$ \cite{zheng2015Forecasting}.

\section{Methodology Perspective}
\label{FusionMethod}
\definecolor{mycolor}{RGB}{215, 245, 200}

\tikzstyle{my-box}=[
    rectangle,
    draw=hidden-draw,
    rounded corners,
    text opacity=1,
    minimum height=1.5em,
    minimum width=5em,
    inner sep=2pt,
    align=center,
    fill opacity=.5,
    line width=0.8pt,
]
\tikzset{
leaf/.style={
my-box,
minimum height=1.5em,
fill=mycolor, 
text=black,
align=left,
font=\footnotesize,
inner xsep=2pt,
inner ysep=4pt,
line width=0.8pt
}
}
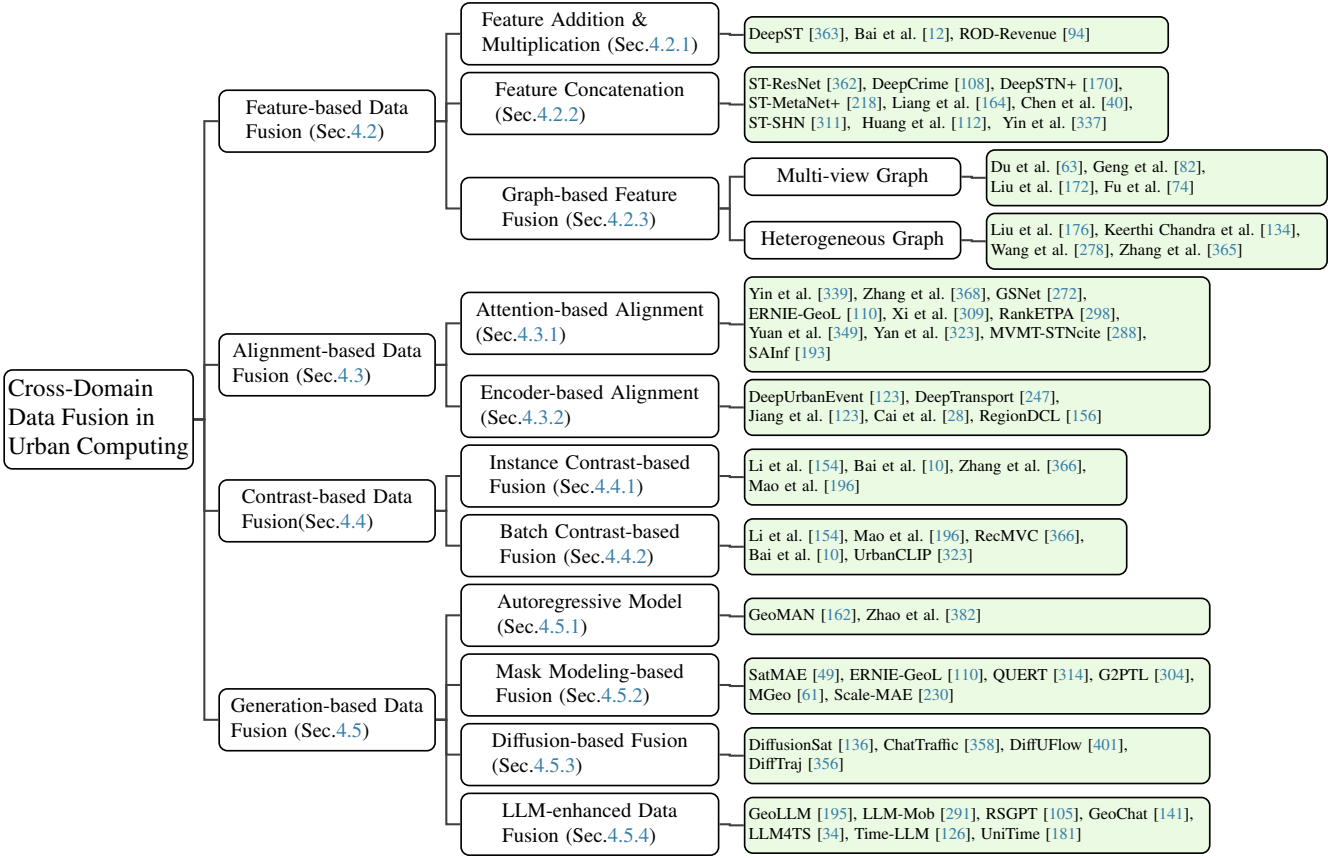
\begin{figure*}[t!]
    \centering
    \begin{adjustbox}{width=0.95\textwidth}
        \begin{forest}
            forked edges,
            for tree={
                grow=east,
                reversed=true,
                anchor=base west,
                parent anchor=east,
                child anchor=west,
                base=center,
                font=\large,
                rectangle,
                draw=hidden-draw,
                rounded corners,
                align=left,
                text centered,
                minimum width=5em,
                edge+={darkgray, line width=1pt},
                s sep=3pt,
                inner xsep=2pt,
                inner ysep=3pt,
                line width=0.8pt,
                ver/.style={rotate=90, child anchor=north, parent anchor=south, anchor=center},
            },
            where level=1{text width=10em,font=\normalsize,}{},
            where level=2{text width=12em,font=\normalsize,}{},
            where level=3{text width=10em,font=\normalsize,}{},
            where level=4{text width=7em,font=\normalsize,}{},
            where level=5{text width=7em,font=\normalsize,}{},
            [Cross-Domain \\Data Fusion in \\Urban Computing
               [Feature-based Data \\Fusion (Sec.\ref{sec:featurebased})
                    [Feature Addition \& \\Multiplication (Sec.\ref{sec: feature addition multiplication})
                        [DeepST \cite{zhang2016dnn}{, }\citet{bai2019spatio}{, }ROD-Revenue \cite{guo2019rod}, leaf, text width=20em]
                    ]
                    [Feature Concatenation \\(Sec.\ref{sec: feature concat})
                        [ST-ResNet \cite{zhang2017deep}{, }DeepCrime \cite{huang2018deepcrime}{, }DeepSTN+ \cite{lin2019deepstn+}{, } \\ 
                        ST-MetaNet+ \cite{pan2020spatio}{, }\citet{liang2021fine-grained}{, }\citet{chenUVLensUrbanVillage2021}{, }\\
                        ST-SHN \cite{xia2021spatial}{, } \citet{huang2023comprehensive}{, } \citet{yin2023multimodal}, leaf, text width=20em]
                    ]
                    [Graph-based Feature\\ Fusion (Sec.\ref{sec: graph fusion})
                        [Multi-view Graph
                        [\citet{duGeofirstLawLearning2019,geng2019spatiotemporal}{, }\\
                        \citet{liu2023characterizing,fu2019efficient}, leaf, text width=16em]
                        ]
                        [Heterogeneous Graph
                        [\citet{liu2019joint,keerthi2020collective}{, }\\
                        \citet{wang2021spatio,zhang2022multi}, leaf, text width=16em]
                        ]
                    ]
                ]
                [Alignment-based Data \\ Fusion (Sec.\ref{sec:alignment})
                    [Attention-based Alignment \\(Sec.\ref{sec:attention fusion})
                        [\citet{yinMultitaskLearningFramework2020a,zhangMultiviewJointGraph2021a}{, }GSNet \cite{wang2021gsnet}{, }\\
                        ERNIE-GeoL \cite{huang2022ernie}{, }\citet{xiFirstLawGeography2022b}{, }RankETPA \cite{wen2023enough}{, }\\
                        \citet{yuan2021effective,yan2023urban}{, }MVMT-STNcite \cite{wangTrafficAccidentRisk2023}{, }\\
                        SAInf \cite{ma2023sainf}, leaf, text width=22em]
                    ]
                    [Encoder-based Alignment\\ (Sec.\ref{sec:encoder fusion})
                        [DeepUrbanEvent \cite{jiang2019deepurbanevent}{, }DeepTransport \cite{song2019deep}{, }\\
                        \citet{jiang2019deepurbanevent,cai2023m}{, }RegionDCL \cite{liUrbanRegionRepresentation2023}, leaf, text width=22em]
                    ]
                ]
                [Contrast-based Data \\Fusion(Sec.\ref{sec:contrastive})
                    [Instance Contrast-based\\ Fusion
                        (Sec.\ref{sec: instance contrast})
                        [\citet{li2022predicting,bai2023geographic,zhangRegionEmbeddingIntra2023}{, }\\
                        \citet{mao2022jointly}, leaf, text width=18em]
                    ]
                    [Batch Contrast-based \\Fusion (Sec.\ref{sec: batch contrast})
                        [\citet{li2022predicting, mao2022jointly}{, }RecMVC \cite{zhangRegionEmbeddingIntra2023}{, }\\
                        \citet{bai2023geographic}{, }UrbanCLIP \cite{yan2023urban}, leaf, text width=18em]
                    ]
                ]
                [Generation-based Data \\Fusion (Sec.\ref{sec:generative})
                    [Autoregressive Model\\(Sec.\ref{sec:autoregressive})
                         [GeoMAN \cite{liang2018geoman}{, }\citet{zhao2021bounding}, leaf, text width=22em
                        ]
                    ]
                    [Mask Modeling-based\\Fusion (Sec.\ref{sec:maskmodeling})
                        [SatMAE \cite{cong2022satmae}{, }ERNIE-GeoL \cite{huang2022ernie}{, }QUERT \cite{xie2023quert}{, }G2PTL \cite{wu2023g2ptl}{, } \\
                        MGeo \cite{ding2023mgeo}{, }Scale-MAE \cite{reed2023scale}, leaf, text width=22em
                        ]
                    ]
                    [Diffusion-based Fusion\\(Sec.\ref{sec:diffusion})
                        [DiffusionSat \cite{khanna2023diffusionsat}{, }ChatTraffic \cite{zhang2023chattraffic}{, }DiffUFlow \cite{zheng2023diffuflow}{, }\\
                        DiffTraj \cite{zhu2023DiffTraj}, leaf, text width=22em
                        ]
                    ]
                    [LLM-enhanced Data\\Fusion (Sec.\ref{sec:llm})
                        [GeoLLM \cite{manvi2023geollm}{, }LLM-Mob \cite{wang2023would}{, }RSGPT \cite{hu2023rsgpt}{, }GeoChat \cite{kuckreja2023geochat}{, }\\
                        LLM4TS \cite{chang2023llm4ts}{, }Time-LLM \cite{jin2023time}{, }UniTime \cite{liu2023unitime}, leaf, text width=22em
                        ]
                    ]
                ]
            ]
        \end{forest}
    \end{adjustbox}
        \vspace{-4mm}
    \caption{Taxonomy of deep learning-based cross-domain data fusion methods in urban computing.}
    \label{fig:fusion tree}
\end{figure*} 
In this section, we begin by providing definitions and explanations of four types of multi-modal fusion. Then, we explore each type in detail, offering more specific categorizations and providing illustrative examples for each case. The comprehensive summary of fusion models can be found in Table \ref{tab:fusiontable}.

\subsection{Overview}
 The integration of deep learning techniques into urban computing has facilitated the development of various deep-learning-based data fusion methods. These techniques aim to leverage the inherent connections within diverse urban data streams. To better understand the differences in the underlying concepts of data fusion in these studies, \add{we consult prior surveys on methodologies in Urban Computing \cite{zheng2015methodologies,liu2020urban} and taxonomies of deep learning methods in other fields \cite{wang2020deep,yuan2021survey,deldari2022beyond,gao2022generative} and categorize the lasted researches into four distinct groups based on their underlying techniques,} as illustrated in Figure \ref{fig:fusion tree}. The definitions of each category are outlined as follows:

\definecolor{geo}{RGB}{149,149,149}
\definecolor{traffic}{RGB}{160,224,224}
\definecolor{social}{RGB}{245,201,200}
\definecolor{demo}{RGB}{251,235,101}
\definecolor{env}{RGB}{60,152,152}

\definecolor{sp}{RGB}{255,230,183}
\definecolor{visual}{RGB}{170,220,224}
\definecolor{text}{RGB}{082,143,173}
\definecolor{graph}{RGB}{030,070,110}

\newcommand{\spcicle}{\textcolor{sp}{\CircleSolid}}
\newcommand{\graphcicle}{\textcolor{graph}{\CircleSolid}}
\newcommand{\visualcicle}{\textcolor{visual}{\CircleSolid}}
\newcommand{\textcicle}{\textcolor{text}{\CircleSolid}}

\newcommand{\geoicon}{\textcolor{geo}{\TriangleUp}}
\newcommand{\trafficicon}{\textcolor{traffic}{\JackStarBold}}
\newcommand{\socialicon}{\textcolor{social}{\DiamondSolid}}
\newcommand{\demoicon}{\textcolor{demo}{\OrnamentDiamondSolid}}
\newcommand{\envicon}{\textcolor{env}{\FourClowerSolid}}

\begin{table*}
    
\caption{The summary of deep learning-based cross-domain data fusion models in urban computing. We denote different data source as follows: Traffic Data \trafficicon ; Geographical Data \geoicon ; Social Media Data \socialicon ; Demographical Data \demoicon ; Environmental Data \envicon . Notice that method names are assigned based on original reference model names if available; otherwise, they are named after the first authors.}
\centering
\resizebox{\textwidth}{!}{%

\begin{tabular}{ccccccccccccccc}
\toprule[2.0pt]
\specialrule{0.1em}{0.3em}{0.2em}
 &  &  & \multicolumn{9}{c}{\textbf{Modality}} &  &  &  \\ \cmidrule{4-12}
 &  &  & \multicolumn{5}{c}{\textbf{\textit{General Spatio-temporal}}}  & \multicolumn{2}{c}{\textbf{\textit{Visual}}} & \multicolumn{2}{c}{\textbf{\textit{Textual}}} &  &  &  \\ 
 \cmidrule(r){4-8} \cmidrule(r){9-10} \cmidrule(r){11-12}
\multirow{-3}{*}{\textbf{Category}} & \multirow{-3}{*}{\textbf{Method}} & \multirow{-3}{*}{\textbf{Data Sourc}e} & \textit{\makecell[c]{Time \\series}} & \textit{\makecell[c]{POI /\\Location}} & \textit{\makecell[c]{Trajectory/\\Road network}} & \textit{Mobility} & \multicolumn{1}{c}{\textit{\makecell[c]{ST \\events}}} & {  \textit{\makecell[c]{Satellite\\ image}}} & \multicolumn{1}{c}{{  \textit{\makecell[c]{Street-view\\ image}}}} & {  \textit{\makecell[c]{Social \\media text}}} & {  \textit{\makecell[c]{Geo-\\imformation\\ text}}} & \multirow{-3}{*}{\textbf{Application}} & \multirow{-3}{*}{\textbf{Institution}} & \multirow{-3}{*}{\textbf{Year}} \\ \midrule[1.5pt]

\cellcolor{lightgray!0}
&DeepST \cite{zhang2016dnn}&\trafficicon\geoicon\demoicon&\spcicle &&\spcicle& &&&&&&Transportation&Microsoft&2016\\

\cellcolor{lightgray!0}
&ST-ResNet \cite{zhang2018predicting}&\trafficicon\geoicon\demoicon\envicon&\spcicle &\spcicle&\spcicle& &&&&&&Transportation&Microsoft&2018\\

\cellcolor{lightgray!0}
&ST-MetaNet+ \cite{pan2020spatio} 
&\trafficicon\geoicon 
& \spcicle
& \spcicle
&  \spcicle
& 
& 
& 
& 
& 
& 
&Transportation 
&JD Research 
&2020 
\\\cellcolor{lightgray!0}

\cellcolor{lightgray!0}
&DeepCrime \cite{huang2018deepcrime} 
&\geoicon\socialicon\demoicon 
& \spcicle
& \spcicle
&  \spcicle
& 
& 
& 
& 
& 
& 
&Social 
&JD Research 
&2021 
\\\cellcolor{lightgray!0}

\cellcolor{lightgray!0}
&STUKG \cite{wang2021spatio}&\geoicon\socialicon& &\spcicle&\spcicle& &&&&&&Transportation&THU&2021\\

\cellcolor{lightgray!0}
&DeepSTN+ \cite{lin2019deepstn+}&\trafficicon\geoicon& \spcicle&\spcicle&& &&&&&&Transportation&THU&2019\\

\cellcolor{lightgray!0}
&DeepTP \cite{yuan2021effective}&\trafficicon\geoicon& &\spcicle& &\spcicle &&&&&&Transportation&THU&2021\\

\cellcolor{lightgray!0}
&\citet{guo2019rod}&\trafficicon\socialicon
&\spcicle &\spcicle& \spcicle&&&&&&&Transportation&BUAA&2019\\

\cellcolor{lightgray!0}
&Photo2Trip \cite{zhao2017photo2trip} 
&\trafficicon\geoicon\socialicon 
& 
&\spcicle 
&  
& 
& 
& 
& \visualcicle
& 
& 
&Transportation 
&SU/RU/UCA 
&2017 
\\

\cellcolor{lightgray!0}
&ST-SHN \cite{xia2021spatial} 
&\geoicon\demoicon 
& \spcicle
&\spcicle 
&  
& 
& 
& 
& 
& 
& 
&Public Safety 
&SCUT/HKU 
&2021 
\\

\cellcolor{lightgray!0}
&GeoMAN \cite{liang2018geoman} 
&\geoicon\envicon 
& \spcicle
&\spcicle 
&  
& 
& 
& 
& 
& 
& 
&General 
&XDU 
&2018 
\\

\cellcolor{lightgray!0}
&\citet{huang2023comprehensive} 
&\trafficicon\geoicon 
& 
& 
&  
& \spcicle
& 
& \visualcicle
& \visualcicle
& 
& 
&Urban planning 
&PKU 
&2023 
\\\cellcolor{lightgray!0}

&\citet{liang2021fine-grained} 
&\trafficicon 
& \spcicle
& \spcicle
&  \spcicle
& 
& 
& 
& 
& 
& 
&Transportation 
&NUS 
&2021 
\\\cellcolor{lightgray!0}

&\citet{balsebreGeospatialEntityResolution2022} 
&\geoicon 
& 
& \spcicle
&  
& 
& 
& 
& 
& 
& \textcicle
&Urban Planning 
&NTU 
&2022 
\\\cellcolor{lightgray!0}

&\citet{ruan2022service} 
&\geoicon \trafficicon
& \spcicle
& \spcicle
&  
& 
& 
& 
& 
& 
& 
&Transportation 
&NTU 
&2022 
\\\cellcolor{lightgray!0}

&\citet{liu2023characterizing} 
&\geoicon \socialicon
& 
& \spcicle
&  
& 
& \spcicle
& 
& 
& 
& 
&Economy 
&HKUST(GZ) 
&2023 
\\\cellcolor{lightgray!0}

&PANDA \cite{you2022panda} 
&\trafficicon \socialicon\demoicon\envicon
& \spcicle
& 
&  \spcicle
& 
& 
& 
& 
& 
& 
&Public Safety 
&XMU 
&2022 
\\\cellcolor{lightgray!0}

&UVLens \cite{chenUVLensUrbanVillage2021} 
&\trafficicon \geoicon \demoicon
& 
& 
&  \spcicle
& \spcicle
& 
& \visualcicle
& 
& 
& 
&Urban Planning 
&XMU 
&2021 
\\\cellcolor{lightgray!0}

&\citet{miyazawa2019integrating} 
& \geoicon \socialicon
& 
& \spcicle
&  \spcicle
& 
& 
& 
& 
& \textcicle
& 
&Transportation 
&SUSTech 
&2019 
\\\cellcolor{lightgray!0}

&NodeSense2Vec \cite{chandraNodeSense2VecSpatiotemporalContextaware2021} 
& \trafficicon\geoicon
& 
& \spcicle
&  
& 
& \spcicle
& 
& 
& \textcicle
& 
&Social 
&UCF 
&2021 
\\\cellcolor{lightgray!0}

&\citet{keerthi2020collective} 
& \trafficicon\geoicon\socialicon
& 
& \spcicle
&  
& \spcicle
& 
& 
& 
& 
& 
&Urban Planning 
&UCF 
&2020 
\\\cellcolor{lightgray!0}

&\citet{fu2019efficient} 
& \trafficicon\geoicon\socialicon
& 
& \spcicle
&  
& 
& \spcicle
& 
& 
& 
& 
&General 
&UCF 
&2019 
\\\cellcolor{lightgray!0}

&\citet{liu2020spatiotemporal} 
& \socialicon
& 
& \spcicle
&  
& 
& 
& 
& 
& \textcicle
& 
&Social 
&Gatech 
&2022 
\\\cellcolor{lightgray!0}

&\citet{yuan2021effective} 
& \geoicon
& \spcicle
& \spcicle
&  
& 
& 
& 
& 
& 
& 
&Transportation 
&RMIT 
&2021 
\\\cellcolor{lightgray!0}

&\citet{bai2019spatio} 
& \trafficicon\envicon
& \spcicle
& 
&  
& 
& 
& 
& 
& 
& 
&Transportation 
&Shanghai AI Lab 
&2019 
\\\cellcolor{lightgray!0}

&\citet{ke2021joint} 
& \trafficicon
& \spcicle
& \spcicle
&  
& 
& 
& 
& 
& 
& 
&Transportation 
&Alibaba 
&2021 
\\\cellcolor{lightgray!0}

&\citet{gengMultimodalGraphInteraction2019} 
& \trafficicon
& 
& \spcicle
&  \spcicle
& 
& 
& 
& 
& 
& 
&Transportation 
&Alibaba 
&2019 
\\\cellcolor{lightgray!0}

&\citet{yao2018deep} 
& \trafficicon
& \spcicle
& \spcicle
&  
& 
& 
& 
& 
& 
& 
&Transportation 
&PSU 
&2018 
\\\cellcolor{lightgray!0}

&\citet{gao2022contextual} 
& \trafficicon
& \spcicle
& \spcicle
&  \spcicle
& \spcicle
& \spcicle
& 
& 
& 
& 
&Transportation 
&SWJTU 
&2022 
\\\cellcolor{lightgray!0}

&DeepMob \cite{songDeepMobLearningDeep2017} 
& \trafficicon\demoicon
& \spcicle
& \spcicle
&  \spcicle
& \spcicle
& 
& 
& 
& 
& \textcicle
&Public Safety 
&SUSTech 
&2017 
\\\cellcolor{lightgray!0}

\multirow{-30}{*}{\cellcolor{lightgray!0}\textbf{\textit{\makecell[c]{Feature\\Based\\Data\\Fusion}}}}
&\citet{geng2019spatiotemporal} 
& \trafficicon\geoicon
& 
& \spcicle
&  \spcicle
& \spcicle
& 
& 
& 
& 
& 
&Transportation 
&HKUST 
&2019 
\\

\midrule
\cellcolor{lightgray!0}
&\citet{xiFirstLawGeography2022b} 
& \geoicon
& 
& \spcicle
& 
& 
& 
& \visualcicle
& 
& 
& 
& General 
&THU 
&2022 
\\\cellcolor{lightgray!0}

&\citet{zhangMultiviewJointGraph2021a} 
& \trafficicon\geoicon\demoicon
& 
& \spcicle
& 
& \spcicle
& 
& 
& 
& 
& 
& General 
&THU 
&2021 
\\\cellcolor{lightgray!0}

&\citet{yuan2021effective} 
& \trafficicon\geoicon\envicon
& 
& \spcicle
& 
& \spcicle
& 
& 
& 
& 
& 
& Transportation 
&THU 
&2021 
\\\cellcolor{lightgray!0}

&\citet{yinMultitaskLearningFramework2020a} 
& \trafficicon\geoicon\demoicon\envicon
& 
& 
& \spcicle
& 
& 
& 
& 
& 
& 
& Urban Planning 
& NUS 
&2020 
\\\cellcolor{lightgray!0}

& GSNet \cite{wang2021gsnet} 
& \trafficicon\geoicon\envicon
& \spcicle
& \spcicle
& \spcicle
& 
& 
& 
& 
& 
& 
& Public Safety 
&BJTU 
&2021 
\\\cellcolor{lightgray!0}

& \citet{hashem2023urban} 
& \geoicon\demoicon
& 
& \spcicle
& 
& 
& 
& 
& 
& 
& 
& General 
& NTU 
&2023 
\\\cellcolor{lightgray!0}

& TrajGAT \cite{yao2022trajgat} 
& \trafficicon
& 
& \spcicle
& \spcicle
& 
& 
& 
& 
& 
& 
& Transportation 
& NTU 
&2022 
\\\cellcolor{lightgray!0}

& RADAR \cite{chen2018radar} 
& \trafficicon
& \spcicle
& 
& \spcicle
& 
& 
& \visualcicle
& 
& 
& 
& General 
& XMU 
&2018 
\\\cellcolor{lightgray!0}

& \citet{wang2021traffic} 
& \trafficicon\geoicon\envicon
& 
& \spcicle
& \spcicle
& 
& 
& 
& 
& 
& 
& General 
& CSU 
&2021 
\\\cellcolor{lightgray!0}

& \citet{tedjopurnomo2021similar} 
& \trafficicon
& \spcicle
& \spcicle
& 
& 
& 
& 
& 
& 
& 
& Transportation 
& RMIT 
& 2021 
\\\cellcolor{lightgray!0}

& ERNIE-GeoL \cite{huang2022ernie} 
& \geoicon
& 
& \spcicle
& 
& 
& 
& 
& 
& 
& \textcicle
& General 
& Baidu 
& 2022 
\\\cellcolor{lightgray!0}

& SAInf \cite{ma2023sainf} 
& \trafficicon\geoicon\envicon
& \spcicle
& 
& \spcicle
& 
& \spcicle
& 
& 
& 
& 
& Transportation 
& JD Research 
& 2023 
\\\cellcolor{lightgray!0}

\multirow{-13}{*}{\textbf{\textit{\makecell[c]{Alignment\\Based\\Data\\Fusion}}}}
& \citet{gao2023dual} 
& \trafficicon\geoicon
& \spcicle
& \spcicle
& 
& \spcicle
& \spcicle
& 
& 
& 
& 
& Transportation 
& SWJTU 
& 2023 
\\

\midrule
\cellcolor{lightgray!0}
&\ KnowCL \cite{liu2023knowledge} 
&\geoicon\socialicon\demoicon 
& 
& \spcicle
&  
& 
& 
& \visualcicle
& \visualcicle
& 
& 
&Economy 
&THU 
&2023 
\\\cellcolor{lightgray!0}

&\ \citet{li2022predicting} 
&\geoicon\socialicon\demoicon 
& 
& 
&  \spcicle
& 
& 
& \visualcicle
& \visualcicle
& 
& 
&Economy 
&THU 
&2022 
\\\cellcolor{lightgray!0}

&\ MMGR \cite{bai2023geographic} 
&\geoicon\socialicon\demoicon 
& 
& \spcicle
&  
& 
& 
& \visualcicle
& 
& 
& 
&General 
&NTU 
&2023 
\\\cellcolor{lightgray!0}

&\ ReMVC \cite{zhangRegionEmbeddingIntra2023} 
& \trafficicon\geoicon\demoicon 
& 
& \spcicle
&  
& \spcicle
& 
& 
& 
& 
& 
&Urban Planning 
&NTU 
&2022 
\\\cellcolor{lightgray!0}

&\ HMTRL \cite{liu2023unified} 
& \trafficicon 
& \spcicle
& 
&  \spcicle
& 
& 
& 
& 
& 
& 
&Transportation 
&UCF 
&2023 
\\\cellcolor{lightgray!0}

&\ \citet{mao2022jointly} 
& \trafficicon 
& \spcicle
& 
&  \spcicle
& 
& 
& 
& 
& 
& 
&Transportation 
&Shanghai AI Lab 
&2022 
\\\cellcolor{lightgray!0}

\multirow{-8}{*}{\textbf{\textit{\makecell[c]{Contrast\\Baed\\Data\\Fusion}}}}
&\ UrbanSTC \cite{qu2022forecasting} 
& \trafficicon 
& \spcicle
& 
&  
& \spcicle
& 
& 
& 
& 
& 
&Transportation 
&JD Research 
&2022 
\\\cellcolor{lightgray!0}

&\ UrbanCLIP \cite{yan2023urban} 
& \geoicon\socialicon 
& 
& 
&  
& 
& 
& \visualcicle
& 
& 
& \textcicle
&General 
&HKUST(GZ) 
&2023 
\\

\midrule
\cellcolor{lightgray!0}
&\ SG-GAN \cite{zhang2020enhanced} 
& \geoicon 
& 
& \spcicle
&  
& 
& 
& \visualcicle
& 
& 
& 
&Urban Planning 
&NUS 
&2020 
\\\cellcolor{lightgray!0}


&\ ActSTD \cite{yuan2022activity} 
& \geoicon\socialicon 
& \spcicle
& 
&  \spcicle
& 
& 
& 
& 
& 
& 
&Transportation 
&THU 
&2022 
\\\cellcolor{lightgray!0}

&\ DiffSTG \cite{wen2023diffstg} 
&  \trafficicon\geoicon
& \spcicle
& 
&  \spcicle
& 
& 
& 
& 
& 
& 
&General 
&BJTU 
&2023 
\\\cellcolor{lightgray!0}

&\ CP-Route \cite{wen2023modeling} 
& \trafficicon\geoicon 
& \spcicle
& 
&  \spcicle
& 
& 
& 
& 
& 
& 
&Transportation 
&BJTU 
&2023 
\\\cellcolor{lightgray!0}

&\ G2PTL \cite{wu2023g2ptl} 
&  \trafficicon\geoicon\socialicon
& 
& \spcicle
&  \spcicle
& 
& 
& 
& 
& 
& \textcicle
&Transportation 
&Cainiao 
&2023 
\\

\cellcolor{lightgray!0}
&\ DiffUFlow \cite{zheng2023diffuflow} 
&  \trafficicon\geoicon
& 
& \spcicle
&  \spcicle
& 
& 
& 
& 
& 
& 
&Transportation 
&CSU 
&2023 
\\

\cellcolor{lightgray!0}
&\ DP-TFI \cite{xu2023diffusion} 
& \trafficicon\geoicon\envicon 
& \spcicle
& 
&  \spcicle
& 
& 
& 
& 
& 
& 
&Transportation 
&UESTC 
&2023 
\\

\cellcolor{lightgray!0}
&\citet{wang2021deep} 
& \trafficicon\geoicon\demoicon 
& 
& \spcicle
&  
& \spcicle
& \spcicle
& 
& 
& 
& 
&Urban Planning 
&UCF 
&2021 
\\

\cellcolor{lightgray!0}
&Chattraffc\cite{zhang2023chattraffic} 
& \trafficicon 
& 
& 
&  
& \spcicle
& 
& 
& 
& 
& \textcicle
&Transportation 
&BJUT 
&2023 
\\

\cellcolor{lightgray!0}\multirow{-11}{*}{\textbf{\textit{\makecell[c]{Generation\\Based\\Data\\Fusion}}}}
&\ MGEO \cite{ding2023mgeo} 
&  \geoicon
& 
& \spcicle
&  
& 
& 
& 
& 
& 
& \textcicle
&General 
&Alibaba 
&2023 
\\

\bottomrule[2.0pt]

\end{tabular}%
}

\label{tab:fusiontable}
\end{table*}

\textbf{Feature-Based Data Fusion} is a straightforward fusion strategy, involving the combination of features from diverse sources such as sensor, visual, and textual data. The core idea is to consolidate raw or processed data features from various sources, forming comprehensive characteristics of studied urban objects. This integration is crucial for enhancing the predictive capabilities of urban models, facilitating complex analyses such as traffic projections and socioeconomic pattern recognition. The distinctive feature of this category lies in directly merging features using methods such as addition, multiplication, concatenation, or graph-based operations. \add{This type of fusion has relatively low computational complexity, primarily depending on the feature dimensions and the specific fusion operation, typically linear \(O(n)\). Here, \(n\) represents the number of feature dimensions involved in the fusion process, indicating total number of features being combined from various sources.}

\textbf{Alignment-based Data Fusion} aims to identify a shared feature space or structure among diverse data representations, enabling their alignment or integration. This involves the model learning to transform information from one source to another for semantic consistency. For instance, in tasks merging images and text, the model must align visual features with textual descriptions, often achieved through a multi-modal embedding space \cite{rasenberg2020alignment,baltruvsaitis2018multimodal}. The widely used attention mechanism \cite{vaswani2017attention} exemplifies alignment-based data fusion, assigning weights to different input parts to prioritize them. In tasks like image captioning, cross-modal attention aligns and weights specific image regions with the textual description \cite{luo2021dual,zhang2023cross,zohourianshahzadi2022neural}. \add{The computational complexity could be higher due to extensive matrix operations, typically quadratic \(O(n^2)\), where \(n\) is the input sequence length.}

\textbf{Contrast-based Data Fusion} employs the contrastive learning framework to enhance feature discriminability at the sample level, contrasting with alignment-based data fusion that focuses on feature-level alignment \cite{deldari2022beyond,nakada2023understanding,le2020contrastive}. By training the model to differentiate categories or samples, it identifies key distinguishing features. In urban computing, this involves comparing traffic patterns or environmental variables across different areas, time points, and modalities. The pairing of positive and negative samples enables the model to learn and excel in distinguishing complex urban situations, refining the acuity of computational tools \cite{zhang2023spatio,tang2023spatio,pan2023spatial}. \add{Utilizing contrastive learning to enhance feature discriminability, this method requires significant computation for sample pairing and similarity calculations, with complexity often quadratic \(O(n^2)\). Here, \(n\) represents the number of samples involved in the contrastive learning process, indicating the total number of data points being compared.}

\textbf{Generation-based Data Fusion} utilizes deep learning's creative capacity to generate one urban modality under the condition of the same or other modalities \cite{liu2023review,fisch2013knowledge,tran2021generative}. In urban computing, this approach proves beneficial for simulating diverse scenarios, such as crafting traffic patterns under various circumstances and assessing urban planning outcomes \cite{yuan2023learning,amirian2019data}. The generative methods involved in urban multi-modal fusion encompass mask modeling, diffusion, and LLM-enhanced techniques. \add{Employing generative models such as GANs and VAEs to simulate urban scenarios, this approach involves highly complex training and optimization processes, leading to very high computational complexity, also generally quadratic \(O(n^2)\) or higher. Specifically, \(n\) indicates the number of data points or parameters involved in the training and optimization process of the generative model, indicating the scale of the data used to simulate urban scenarios.}

\subsection{Feature-based Data Fusion}
\label{sec:featurebased}
Feature-based fusion integrates information from different modalities through methods like \textit{feature addition}, emphasizing equal importance, and \textit{feature multiplication}, highlighting joint significance (Sec.\ref{sec: feature addition multiplication}). Additionally, techniques such as \textit{feature concatenation} (Sec.\ref{sec: feature concat}) and \textit{graph-based fusion} (Sec.\ref{sec: graph fusion}) enhance multi-modal understanding by combining feature vectors or leveraging graph structures to represent inter-modal connections.
\add{This fusion approach is characterized by its simplicity and efficiency, making it particularly advantageous for applications requiring high real-time performance, such as transportation. However, it is relatively coarse and lacks the ability to capture and align details accurately, which may lead to underperformance in tasks necessitating the amalgamation of diverse datasets, such as urban planning.}
\subsubsection{Feature Addition and Multiplication}
\label{sec: feature addition multiplication}
Element-wise addition is a fundamental operation among multi-modal fusion techniques, where corresponding elements of vectors or matrices from different modalities are summed to create a fused representation. Given two vectors \( \mathbf{X} = [x_1, x_2, \ldots, x_n] \) and \( \mathbf{Y} = [y_1, y_2, \ldots, y_n] \) representing features from different modalities, the element-wise addition operation results in a new vector \( \mathbf{Z} = [z_1, z_2, \ldots, z_n] \), where \( z_i = x_i + y_i \) for \( i = 1, 2, \ldots, n \). Mathematically, the element-wise addition operation is expressed as
\begin{equation}\label{eqn:addition}
     \mathbf{Z} = \mathbf{X} + \mathbf{Y}. 
\end{equation}
Element-wise multiplication varies from the resulting new vector $\mathbf{Z} = [x_1 y_1, x_2 y_2, \ldots, x_n y_n]$. Mathematically, such an operation can be expressed as 
\begin{equation}\label{eqn:multiplication}
    \mathbf{Z} = \mathbf{X} \odot \mathbf{Y},
\end{equation}
where $\odot$ denotes element-wise multiplication. These fusion approaches provide a simple yet effective way to integrate information, preserving the individual contributions of each modality in the combined representation.

For example, \citet{guo2019rod} developed the ROD-Revenue framework to predict driver revenue using a linear regression model, given features extracted from multi-source urban data. The basic features from multiple datasets, including ride-on-demand (RoD) service, taxi service, and POI information, are used to construct composite features in a product form. Likewise, \citet{bai2019spatio} fused the historical passenger demand, meteorological data, and time meta to learn a joint representation for citywide passenger demand prediction.

\subsubsection{Feature Concatenation}
\label{sec: feature concat}
Concatenation is a widely used technique for multi-modal fusion, involving the combination of feature vectors or matrices from different modalities by appending them along a specified axis. Let vectors \( \mathbf{X} = [x_1, x_2, \ldots, x_m] \) and \( \mathbf{Y} = [y_1, y_2, \ldots, y_n] \) represent features from two modalities. Concatenating these vectors along the concatenation axis results in a new vector \( \mathbf{Z} = [x_1, x_2, \ldots, x_m, y_1, y_2, \ldots, y_n] \), forming a fused representation. Mathematically, the concatenation operation is denoted as 
\begin{equation}\label{eqn:concat}
    \mathbf{Z} = \text{concat}(\mathbf{X}, \mathbf{Y}).
\end{equation}
This method allows the preservation of individual modality information while creating a comprehensive representation for downstream tasks.

\begin{figure}[!tbh]
  \centering
  \includegraphics[width=0.45\textwidth]{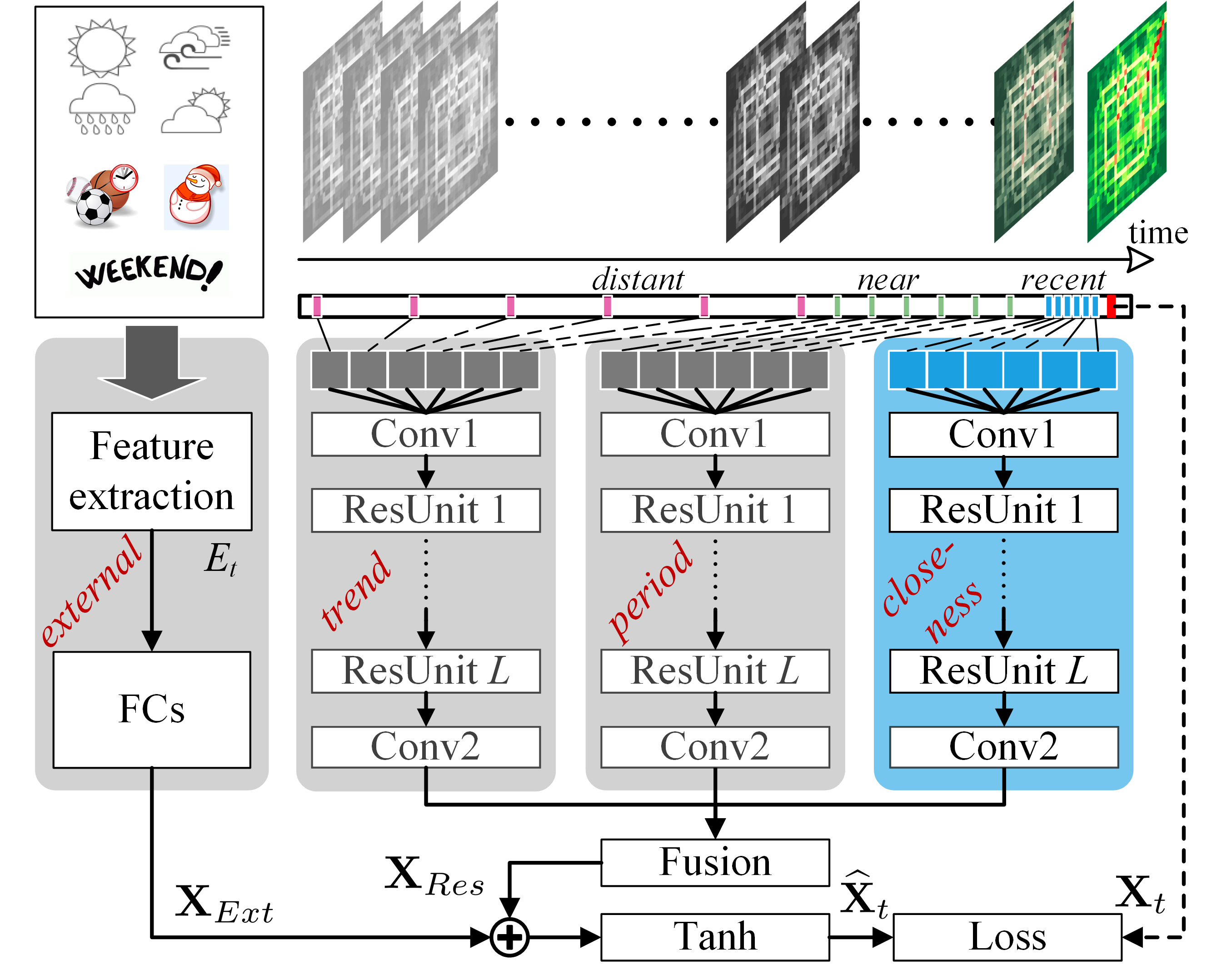}
  \vspace{0em}
  \caption{ST-ResNet dynamically fused the results from three distinct residual temporal networks, allocating varied weights to diverse branches and regions. This fusion process was additionally integrated with external factors, including meteorological information. \cite{zhang2017deep,zhang2018predicting}.}
  \label{fig: feature ST-ResNet}
\end{figure}

\begin{figure*}[!t]
    \centering
    \vspace{-1em}
    \includegraphics[width=0.8\textwidth]{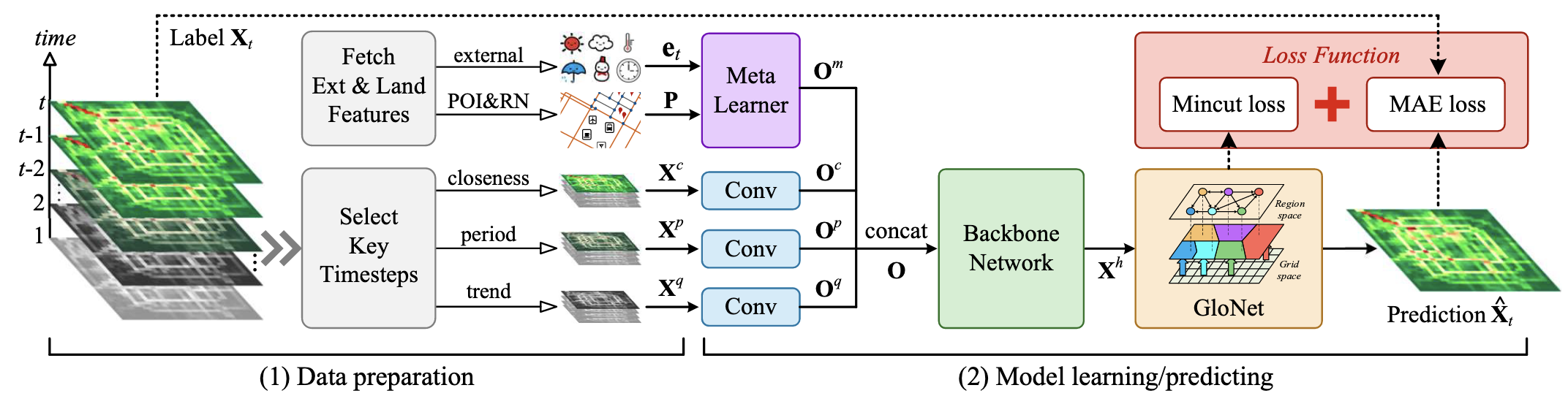}
    \vspace{-1em}
    \caption{The feature-based data fusion framework proposed by \citet{liang2021fine-grained}. Features extracted from the temporal sequence are concatenated together with external features from POI and road networks.}
    \label{fig:yliang}
\end{figure*}

For instance, ST-ResNet \cite{zhang2017deep,zhang2018predicting} dynamically aggregated the outputs of three different residual temporal networks, assigning different weights to different branches and regions. The aggregation was further combined with external factors such as meteorological information (illustrated in Figure \ref{fig: feature ST-ResNet}). Similarly, \citet{lin2019deepstn+} proposed a DeepSTN+ framework to forecast urban crowd flows via long-range spatial dependence modeling and the introduction of prior knowledge such as POI distribution. The three categories of historical crowd flow maps - closeness, period and trend, are concatenated firstly, followed by convolution as the fusion of different kinds of information. Some studies \cite{liang2019urbanfm,ouyang2020fine,sun2020predicting} followed this concatenation scheme to fuse multimodal information (i.e., external factors) as well. Furthermore, \citet{liang2021fine-grained} appended the same three temporal sequences with meta features representing POI and road network (illustrated in Figure \ref{fig:yliang}), which serve as fine-grained knowledge for urban flow prediction. 

In addition, concatenation operation is often used in visual feature fusion. \citet{huangComprehensiveUrbanSpace2023a} concatenated the high-dimensional features of satellite imagery, street-level imagery, and taxi trajectory time series together, and then passed them through a softmax layer to distinguish between urban and non-urban villages. \citet{yin2023multimodal} also concatenated latent representations of satellite imagery and GPS trace as a fusion method for missing road attribute inference. In addition, the features of household capacity, human mobility, and commercial hotness can be merged together to estimate the population \cite{chenUVLensUrbanVillage2021}.

\subsubsection{Graph-based Data Fusion}
\label{sec: graph fusion}

Traditional deep learning-based feature extraction and fusion methods have brought revolutionary advancements to various urban computing tasks in recent years. These tasks typically involve data represented in the Euclidean space. However, there is a growing need for addressing tasks where data is generated from non-Euclidean domains and represented as graphs with intricate relationships between objects such as POI networks. In urban computing, graphs serve as representations of intricate networks that encompass various elements such as roads, buildings, and social interactions. In general, a graph can be represented as
\begin{equation}\label{eqn:graph1}
    \mathbf{G} = (\mathbf{V}, \mathbf{E}) \text{ (General form),}
\end{equation}
or 
\begin{equation}\label{eqn:graph2}
    \mathbf{G}^{(t)} = (\mathbf{V}, \mathbf{E}, \mathbf{X}^{(t)}) \text{ (Spatio-temporal form),}
\end{equation}
where \textbf{V} represents a set of nodes and \textbf{E} is the set of edges. For a spatio-temporal graph, where the node attributes vary dynamically over time, $\textbf{G}^{(t)}$ is the graph representation at time step $t$ and $\textbf{X}^{(t)}$ is the node feature matrix of graph $\textbf{G}$ at the same time step. Deep learning models based on spatio-temporal graphs have gained significant importance and achieved remarkable success across various applications \cite{yu2017spatio,li2017diffusion,jain2016structural,wu2019graph}.

An intuitive approach for graph-based data fusion is \textbf{multi-view graph network-based data fusion}, which represents diverse data through a set of multi-view graphs \cite{liu2023characterizing,fu2019efficient,duGeofirstLawLearning2019,geng2019spatiotemporal}. In this framework, each view is associated with different edge vectors, denoted as $\mathbf{E^i}$, which represent the relationships between each node (such POI), denoted as $\mathbf{V}$, under a particular feature view $\mathbf{i}$ (distance, mobility, semantic, etc.). For graphs from different views, graph nodes $\mathbf{V}$ usually serve as connectors, linking them together for feature fusion. Each view in these frameworks captures a unique perspective or aspect of the underlying data, allowing for a comprehensive understanding of a more complex system.

\begin{figure}[!b]
    \centering
    \includegraphics[width=0.47\textwidth]{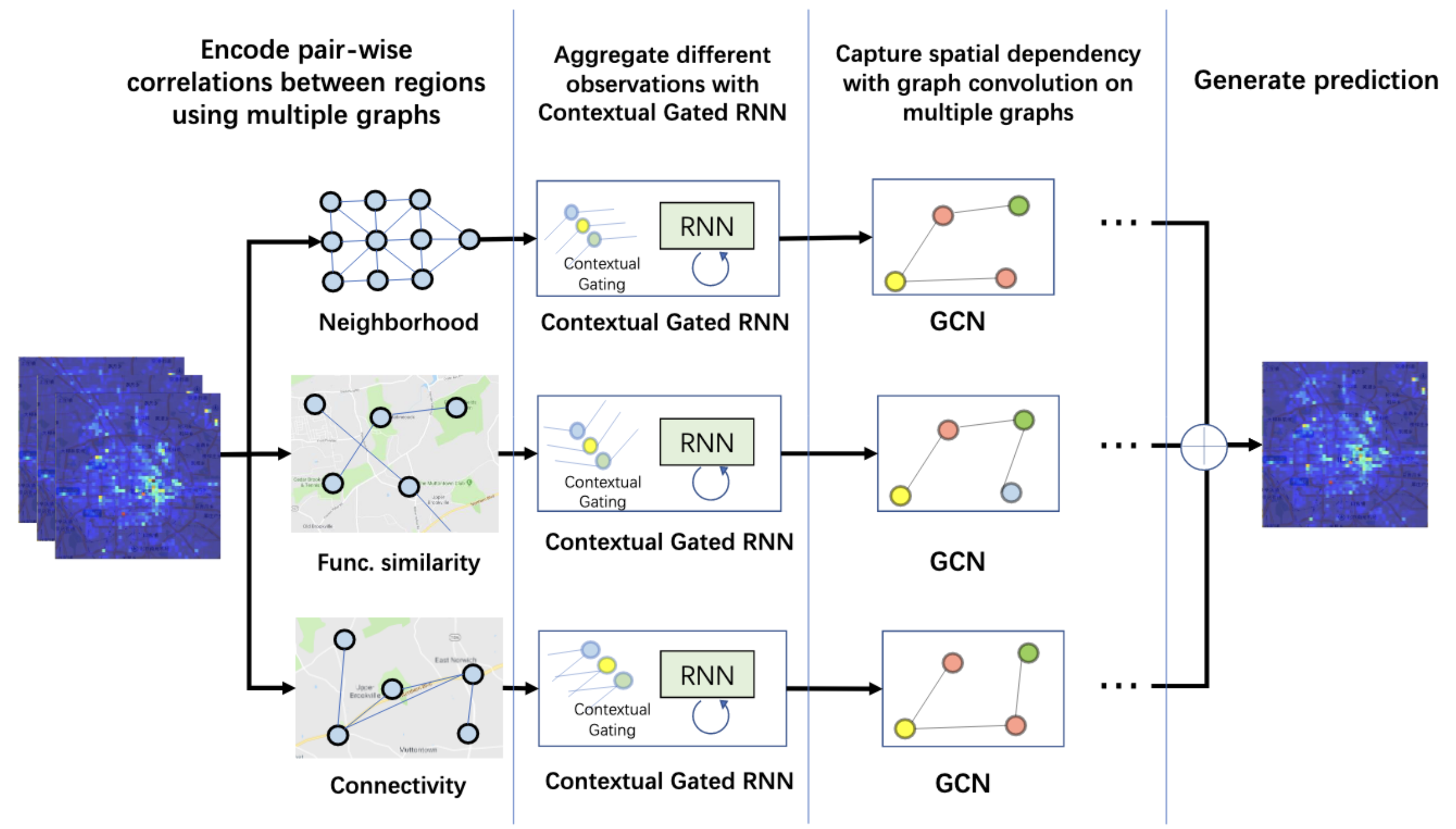}
    \caption{The overall architecture of the spatio-temporal multi-graph convolution network proposed by \citet{geng2019spatiotemporal}.}
    \label{fig:genspatio}
\end{figure}

Based on the conception of a multi-view graph, \citet{fu2019efficient} proposed multi-view POI graph networks that incorporate various geo-features including regions, distances, and human mobility connectivity. \citet{duGeofirstLawLearning2019} developed a group of POI networks to characterize both static and dynamic features through the variable graph edges. Both of these work effectively integrate and fuse information from various data sources and achieve great success in downstream tasks. \citet{liu2023characterizing} conducted research on characterizing urban vibrancy by employing a multi-view graph framework and demonstrated the effectiveness of multi-view graphs in capturing the intricate relationship between urban vibrancy and urban spatiotemporal dynamics. In addition, different regions in a city can be represented through multi-view graphs as well. \citet{geng2019spatiotemporal} proposed a multiple graph framework to encode pair-wise correlations between regions in urban areas. As shown in Figure \ref{fig:genspatio}, each region square, measuring ($1km \times 1km$), is represented as a node in the graphs. By utilizing graph convolutional networks to observe and fuse these graphs, the framework can capture various region correlations and achieve impressive results in forecasting ride-hailing demand.

\begin{figure}[!h]
    \centering
    \includegraphics[width=0.4\textwidth]{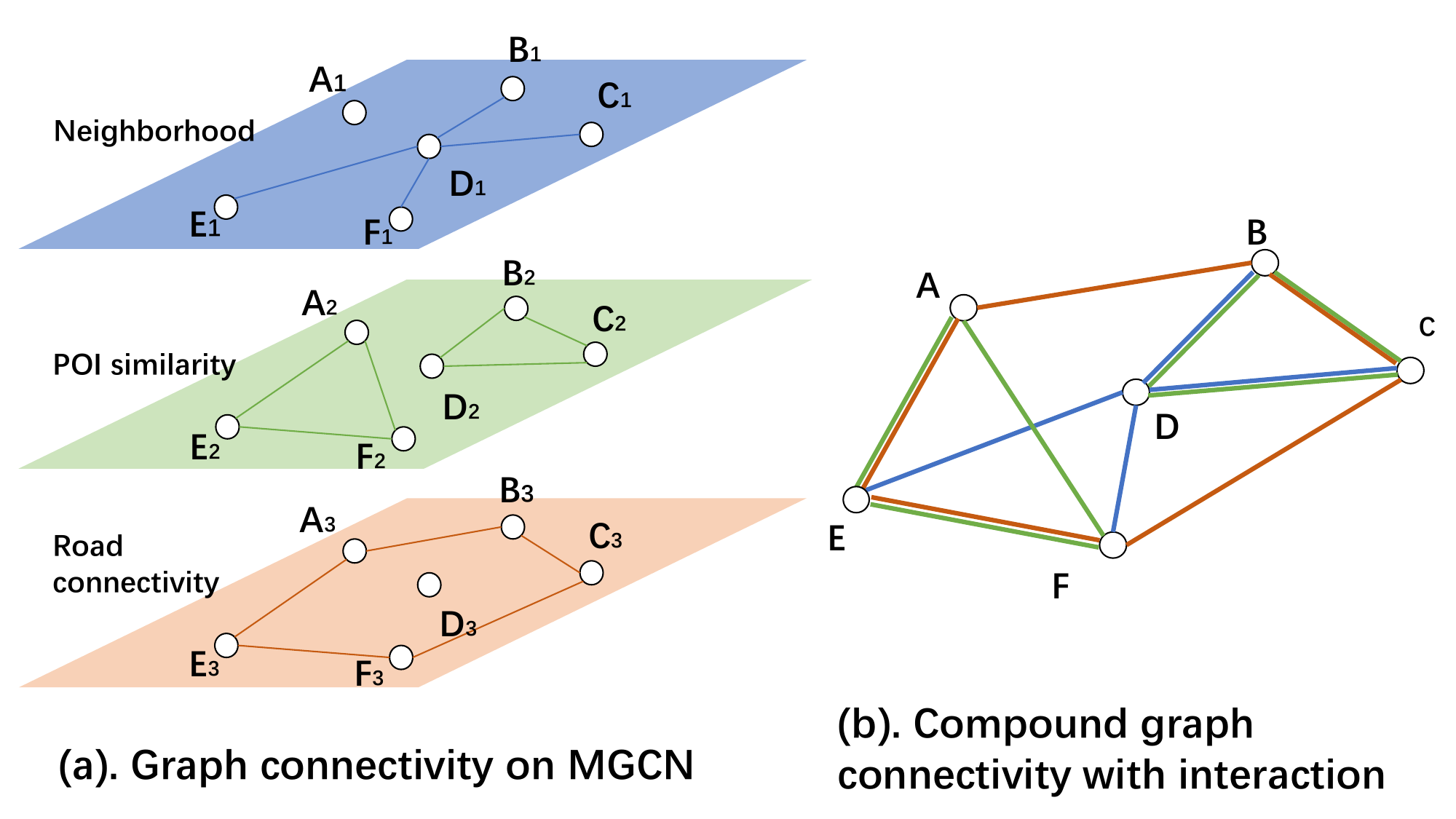}
    \caption{In (a), the multi-view graph connectivity is shown for each individual graph. Vertices represent regions, and the weighted edges represent region-wise relationships. There is no interaction between the graphs.
    In (b), the compound graph connectivity is represented as a multi-linear graph. Vertices are connected if there is an edge in any of the traditional multi-view graphs. This allows for a more comprehensive understanding of the interdependencies among regions across all graphs \cite{zhang2022multi}.}
    \label{fig:xugeng}
\end{figure}

While it is efficient to incorporate features from different data sources using multi-view graphs, describing high-dimensional cross-modal correlation between different items can be challenging as there is no direct interaction between nodes from each graph. To address these difficulties, researchers proposed spatiotemporal heterogeneous networks (SHNs) to encapsulate the complex cross-modal relationships among different urban entities. Subsequently, \textbf{heterogeneous graph-based data fusion} is considered as a promising approach to the fusion of various data in one graph and tackles the above challenges \cite{liu2019joint,zhang2022multi,wang2021spatio,keerthi2020collective,zou2024learning,wang2024hypergraph}. \citet{liu2020spatiotemporal} proposed a hierarchical embedding framework that aimed to establish connections between nodes in the city activity graph and user interaction graph. Although the specific emphasis on SHNs may not be explicitly stressed, this attempt, along with the successful cross-modal node connection, has demonstrated significant success in capturing and representing cross-modal connections. \citet{zhang2022multi} proposed a compound multi-linear relationship graph network to converge various edge connectivity with interaction between multi-view graphs. In their methodology, which is shown in Figure \ref{fig:xugeng}, any random walk path in the compound graph is a compound of various kinds of relationships for different views. This work strongly develops the generality of graph neural networks in cross-domain data fusion.

\citet{chandraNodeSense2VecSpatiotemporalContextaware2021} designed a SHNs embedding framework that takes the POIs as nodes and human mobility between POIs as weighted links. The SHNs in their research are represented as 
\begin{equation}\label{eqn:graph3}
    \mathbf{G} = (\mathbf{V}, \mathbf{E}, \mathbf{\phi}, \mathbf{\psi}),
\end{equation}
where $\phi:\textbf{V}\rightarrow\mathcal{L}$ is a mapping function for nodes and $\psi:\textbf{E}\rightarrow\mathcal{R}$ is a edge type mapping function. Within this presentation based on SHNs, the framework could analyze urban mobility data from multiple sources in a uniform model space.

In addition to the commonly researched SHNs, \citet{liu2019joint} proposed a multi-modal transportation graph consisting of nodes for users, transport modes, and origin-destination pairs. This approach aimed to develop a multi-modal planning methodology in a transport recommend system. Similarly, \citet{wang2021spatio} constructed an urban knowledge graph that effectively integrates features and knowledge from various trajectory data sources to model users' mobility patterns.

\subsection{Alignment-based Data Fusion}
\label{sec:alignment}
\begin{figure}[!htb]
    \centering
    \includegraphics[width=0.38\textwidth]{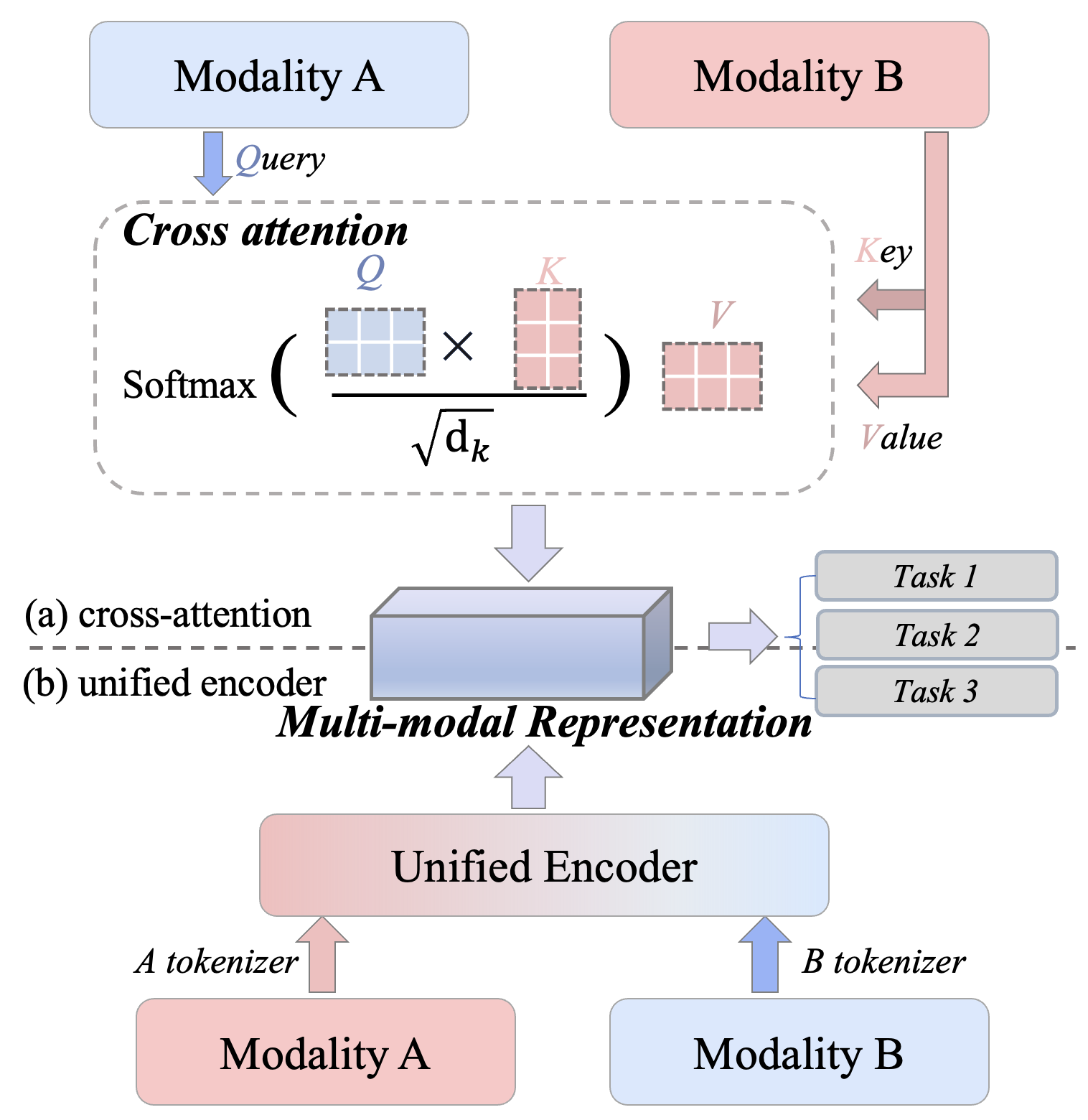}
    \caption{The general framework of alignment-based cross-domain data fusion in urban computing: (a) cross-attention framework for attention-based alignment; (b) unified encoder framework for encoder-based alignment.}
    \label{fig:alignment}
\end{figure}

Alignment fusion ensures that semantically related content across modalities is effectively combined. As illustrated in Figure \ref{fig:alignment}, methods for multi-modal alignment span categories such as \textit{cross-modal attention mechanism} (Sec.\ref{sec:attention fusion}) and \textit{multi-modal encoder-based fusion} (Sec.\ref{sec:encoder fusion}), enabling understanding and collaboration between disparate sources of information. 
\add{Alignment-based approaches can achieve more precise modal alignment and exhibit flexibility, making them suitable for various general scenarios. However, they generally require higher computational resources, which makes them more appropriate for offline tasks in transportation and urban planning.}

\subsubsection{Attention-based Alignment}
\label{sec:attention fusion}
\begin{figure}[!h]
    \centering
    \includegraphics[width=0.33\textwidth]{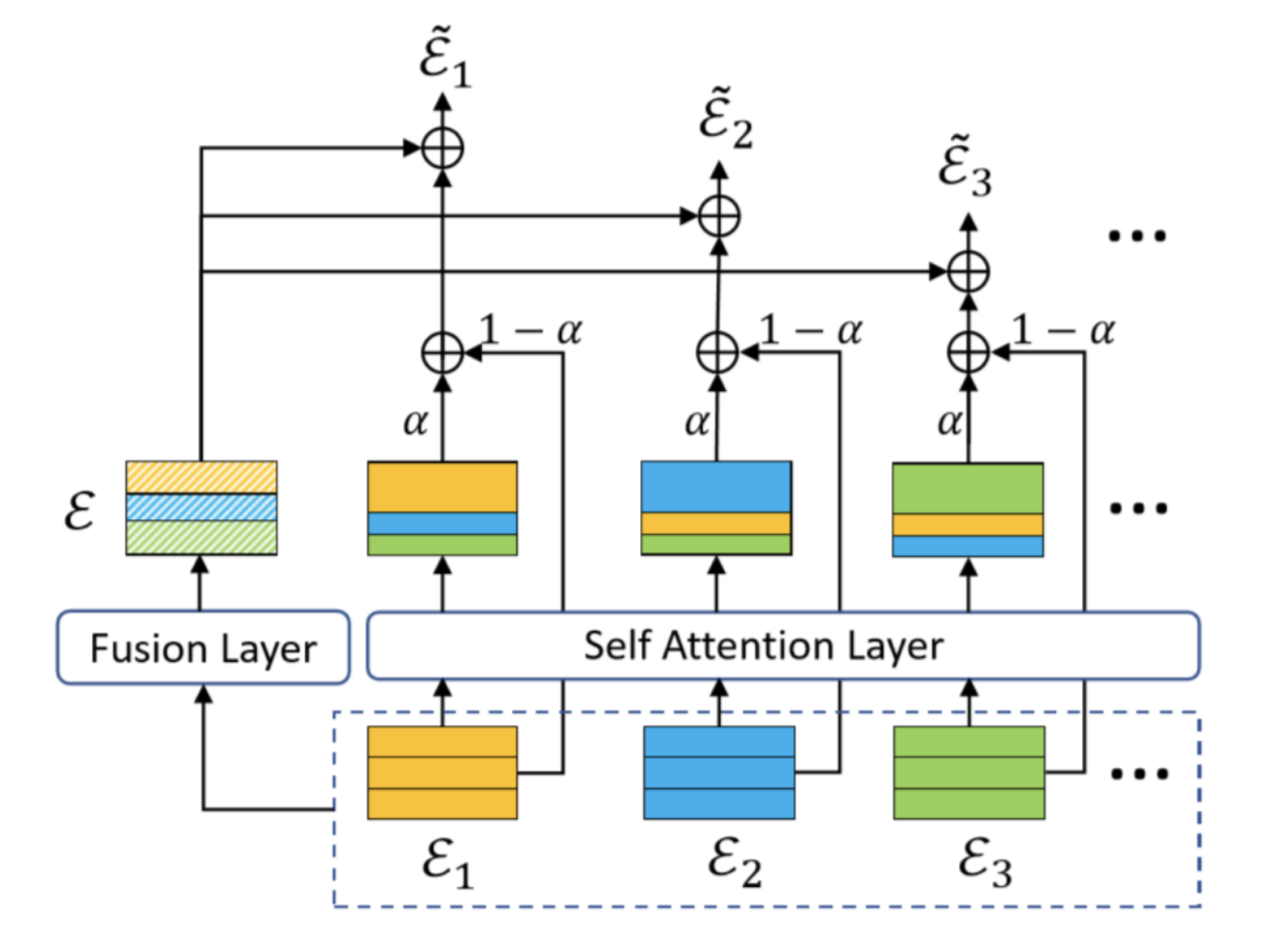}
    \caption{The architecture of multi-view joint learning module, consisting of a self-attention layer and a fusion layer\cite{zhangMultiviewJointGraph2021a}. \add{\( \mathcal{E}_{i}^{\prime} \) is the representation for the \( i \)-th view of global information, and \( \alpha \) is the weight of global information.}}
    \label{fig:liyong}
\end{figure}

Attention mechanism \cite{vaswani2017attention}, especially multi-modal cross-attention is a fusion technique crucial for integrating information across diverse modalities, such as text and images \cite{wei2020multi,ji2020sman,yang2023multimodal}. The cross-attention mechanism can be summarized in the following three steps. First, project the feature vectors of modalities \textbf{X} and \textbf{Y} into query (\textbf{Q}), key (\textbf{K}), and value (\textbf{V}) spaces using learnable parameter matrices \(\mathbf{W}_Q\), \(\mathbf{W}_K\), and \(\mathbf{W}_V\), respectively:
\begin{equation}\label{eqn:cross attention step1}
    \textbf{Q}_X = \textbf{X} \textbf{W}_{Q_X}, \quad \textbf{K}_Y = \textbf{Y} \textbf{W}_{K_Y}, \quad \textbf{V}_Y = \textbf{Y}  \textbf{W}_{V_Y}.
\end{equation}
Second, compute the initial attention scores by taking the dot product of the query vectors of modality \textbf{X} (i.e., \(\mathbf{Q}_X\)) and the key vectors of modality Y (i.e., \(\mathbf{K}_Y\)), divided by the square root of the dimensionality of the key vectors $\sqrt{d_k}$. Subsequently, we apply the softmax function to obtain normalized attention weights, ensuring that the weights sum up to 1:
\begin{equation}\label{eqn:cross attention step2}
    \textbf{A} = \text{Softmax}(\frac{\textbf{Q}_X\textbf{K}_Y^T}{\sqrt{d_k}}).
\end{equation}
Third, use the attention weights to compute the weighted sum of values for each modality:
\begin{equation}\label{eqn:cross attention step3}
   \textbf{Z}_X = \textbf{A}  \textbf{V}_Y, \quad \textbf{Z}_Y = \textbf{A}^T  \textbf{V}_Y.
\end{equation}
In recent years, the urban computing community leveraged such fusion mode and its variants for comprehensive urban modality alignment. For example, to model the relations among the target road attributes, \citet{yinMultitaskLearningFramework2020a} generated task-specific fused representations by applying attention-based feature fusion of location, bearing, speed, and map context. \citet{zhangMultiviewJointGraph2021a} applied a GAT-based attention mechanism in learning region representations from two views of the built correlations (i.e., human mobility view and region attribute view), and a joint learning module to fuse multi-view embeddings. In this proposed multi-view joint learning module shown in Figure \ref{fig:liyong}, the self-attention layer enables information sharing across all views and the fusion layer is responsible for combining multi-view representations via adaptive weights.

The MVMT-STN model \cite{wangTrafficAccidentRisk2023} included a multi-view GCN to capture the global semantic dependency between POI, road network, and risk information. \citet{huang2022ernie} used a transformer-based aggregation layer to model the graph structure containing both toponym and spatial knowledge. \citet{qiang2023modeling} designed a transformer encoder to represent the features of each package by integrating those of other candidate packages. The RankETPA model \cite{wen2023enough} was designed for package pick-up arrival time prediction, and there is an attention module to model the interaction between the package's features and the courier's features.

The other similar attentional fusion work includes \cite{xiFirstLawGeography2022b} (POI-view and geographic-view representations), \cite{yuan2021effective} (region-level and inter-traffic correlations), \cite{wang2021gsnet} (geographical and semantic spatio-temporal representations) and \cite{yan2023urban} (satellite visual and textual representations).

\subsubsection{Encoder-based Alignment}
\label{sec:encoder fusion}

Encoder-based fusion entails the integration of multiple modalities into a shared encoder architecture, as opposed to employing separate encoders for each modality. This approach leverages a unified deep learning model (e.g., RNN variants \cite{song2016deeptransport, jiang2019deepurbanevent} and self-attention methods \cite{cai2023m,liUrbanRegionRepresentation2023}) to collectively process and extract meaningful representations from diverse sources of information. By cohesively injecting multi-modal data into a single encoder, the model is inherently encouraged to capture intricate inter-modal relationships and dependencies. 

For example, \citet{song2016deeptransport} developed the DeepTransport model for mobility simulation and transportation mode prediction, where heterogeneous data including GPS records and transportation information are fed into a deep LSTM learning architecture. Such multi-layer LSTM has been demonstrated to be able to learn at different time scales over the input \cite{hermans2013training}. The DeepUrbanEvent system \cite{jiang2019deepurbanevent} contained a ConvLSTM \cite{shi2015convolutional} encoder module for simultaneous multi-step forecasting of crowd density and crowd flow. The ConvLSTM extends the fully connected LSTM (FC-LSTM) to have convolutional structures in both input-to-state and state-to-state transitions. The RegionDCL framework \cite{liUrbanRegionRepresentation2023} leveraged the Transformer encoder with average-pooling as the region-level encoder of building features and POI information. \citet{cai2023m} proposed a multi-level graph encoder equipped with a GAT encoding module to capture couriers' both high-level transfer modes between AOIs and low-level transfer models between locations.

\subsection{Contrast-based Data Fusion}
\label{sec:contrastive}
Contrastive learning, a pivotal paradigm in machine learning, can be categorized based on contrast creation \cite{le2020contrastive,nakada2023understanding}. The categorization encompasses methods such as instance contrast, batch contrast, and temporal contrast, which each focuses on contrasting representations through augmentations, batch comparisons, and temporal shifts, respectively.
\add{Compared to alignment-based methods, the contrast-based data fusion method enhances discrimination through negative sample augmentation. However, it imposes stringent requirements on the selection of negative samples and batch size. Due to its demand for large datasets, it is well-suited for downstream tasks where data is abundant or easily accessible, such as those related to transportation, environment and economy.}

The InfoNCE (Noise-Contrastive Estimation) loss, a common objective function in contrastive learning, aims to maximize the similarity between positive pairs and minimize the similarity between negative pairs \cite{oord2018representation}. Mathematically, the InfoNCE loss is formulated as follows:
\begin{equation}\label{eqn:InfoNCE}
\small
    \mathcal{L}_{\text{InfoNCE}}(\mathbf{X}, \mathbf{Y}) = -\log\left(\frac{\exp(\text{sim}(\mathbf{X}, \mathbf{Y}))}{\exp(\text{sim}(\mathbf{X}, \mathbf{Y})) + \sum_{k=1}^{K} \exp(\text{sim}(\mathbf{X}, \mathbf{N}_k))}\right),
\end{equation}
where \(K\) is the total number of negative samples, \(\text{sim}(\mathbf{X}, \mathbf{Y})\) is the similarity measure between positive pairs, and \(\mathbf{N}_k\) represents negative samples.

\begin{figure}[!htb]
    \centering
    \includegraphics[width=0.38\textwidth]{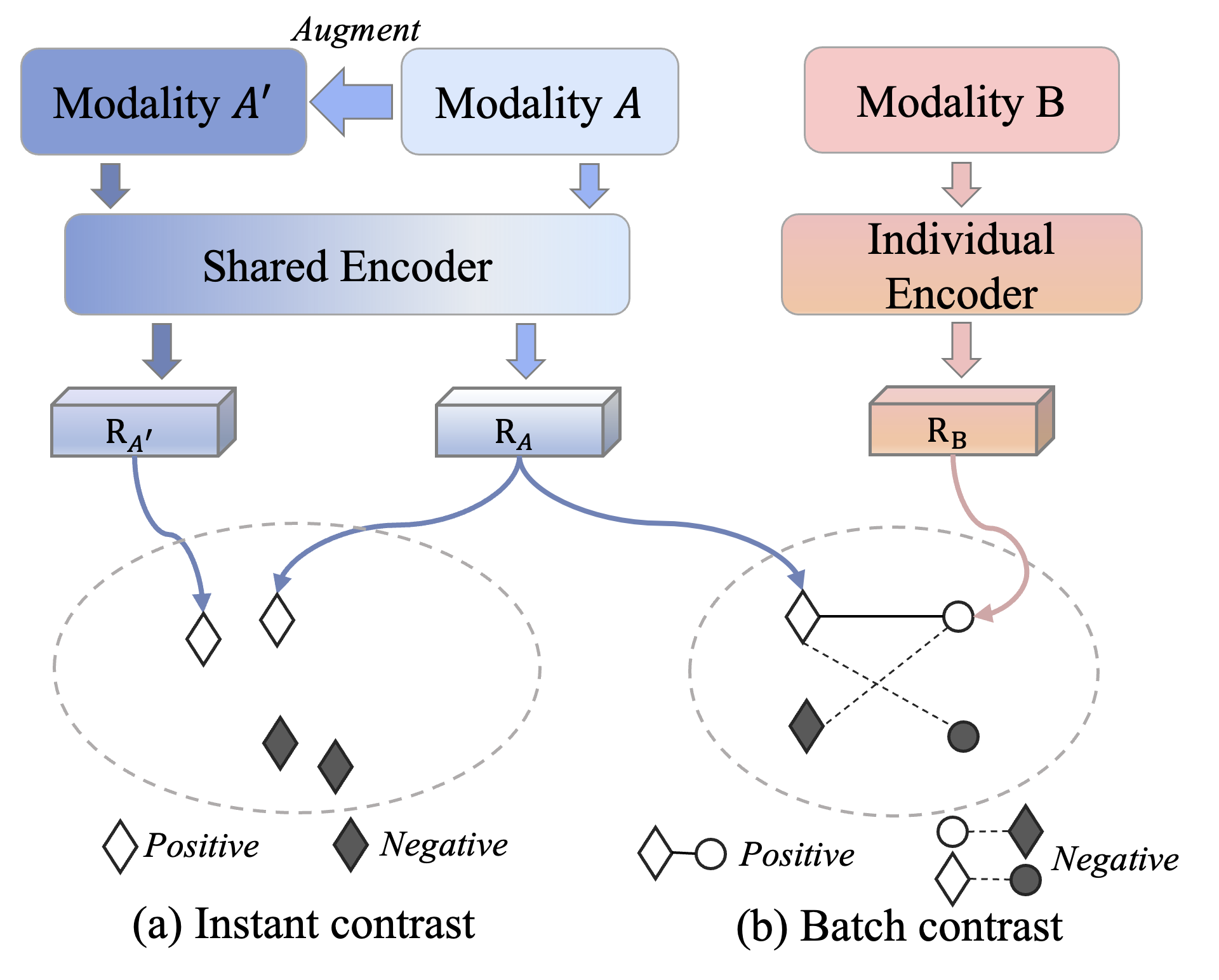}
    \caption{The general framework of contrast-based cross-domain data fusion in urban computing: (a) instant contrast; (b) batch contrast.}
    \label{fig:contrast}
\end{figure}

CLIP (Contrastive Language-Image Pre-training) is a typical model in contrastive learning, designed to concurrently learn representations of images and text \cite{radford2021learning}. CLIP can be considered a form of instance contrast because it encourages the model to align representations of a given instance across different modalities. The model achieves this by minimizing the negative logarithmic probability of similarity between positive image-text pairs while contrasting against negative pairs. Mathematically, the CLIP loss function (\( \mathcal{L}_{\text{CLIP}} \)) is expressed as:
\begin{equation}\label{eqn:CLIP}
\small
    \mathcal{L}_{\text{CLIP}}(\mathbf{I}, \mathbf{T}) = -\log\left( \frac{\exp(\text{sim}(\mathbf{I}, \mathbf{T}))}{\exp(\text{sim}(\mathbf{I}, \mathbf{T})) + \sum_{k=1}^{K} \exp(\text{sim}(\mathbf{I}, \mathbf{N}_k))} \right).
\end{equation}
Here, \(\mathbf{I}\) and \(\mathbf{T}\) represent image and text representations, respectively. In our survey, contrastive data fusion can be categorized into \textit{instance contrast-based fusion} (Sec.\ref{sec: instance contrast}) and \textit{batch contrast-based fusion} (Sec.\ref{sec: batch contrast}), as illustrated in Figure \ref{fig:contrast}.

\subsubsection{Instance Contrast-based Fusion}
\label{sec: instance contrast}

In instance contrast, the model aims to fuse the knowledge between different views of the same instance. It is effective for capturing intricate details within individual data points, promoting the model's ability to recognize fine-grained patterns and features \cite{le2020contrastive,deldari2022beyond,liu2021self}.

Inspired by the computer vision domain \cite{xu2023comprehensive, chengchuang2021review,zoph2020learning}, urban multi-modal research started to construct self-augmented data as contrastive pairs. For example, \citet{li2022predicting} explored self-similarity across urban images to construct contrastive samples via data augmentation methods including rotation, gray-scale, and flipping. Similarly, \citet{bai2023geographic} learned the physical properties of the geographies via intra-modal contrast among very high resolution (VHR) image augmentations. In addition to imagery augmentation, \citet{zhangRegionEmbeddingIntra2023} designed three POI-level augmentations for intra-view contrastive fusion: random insertion, random deletion, and random replacement. Furthermore, \citet{mao2022jointly} introduced domain-specific augmentations for road-road contrast and trajectory-trajectory contrast separately, i.e., road segment with its contextual neighbors and trajectory with its detour replaced and dropped alternatives.

\subsubsection{Batch Contrast-based Fusion}
\label{sec: batch contrast}

Batch contrast involves contrasting samples within the same batch. It introduces a form of global context, where the model learns to distinguish features not just within instances but also in relation to the entire batch \cite{le2020contrastive,deldari2022beyond,liu2021self}.

Geographical similarity-guided contrastive learning is pivotal in urban computing because urban images adhere to Tobler's First Law of Geography \cite{miller2004tobler}. For instance, \citet{li2022predicting} further enhanced the contrastive learning method by taking into account geographical similarity and minimizing the feature distance between two images that are geo-adjacent.

\begin{figure}[!ht]
  \centering
  \vspace{-1em}
  \includegraphics[width=0.48\textwidth]{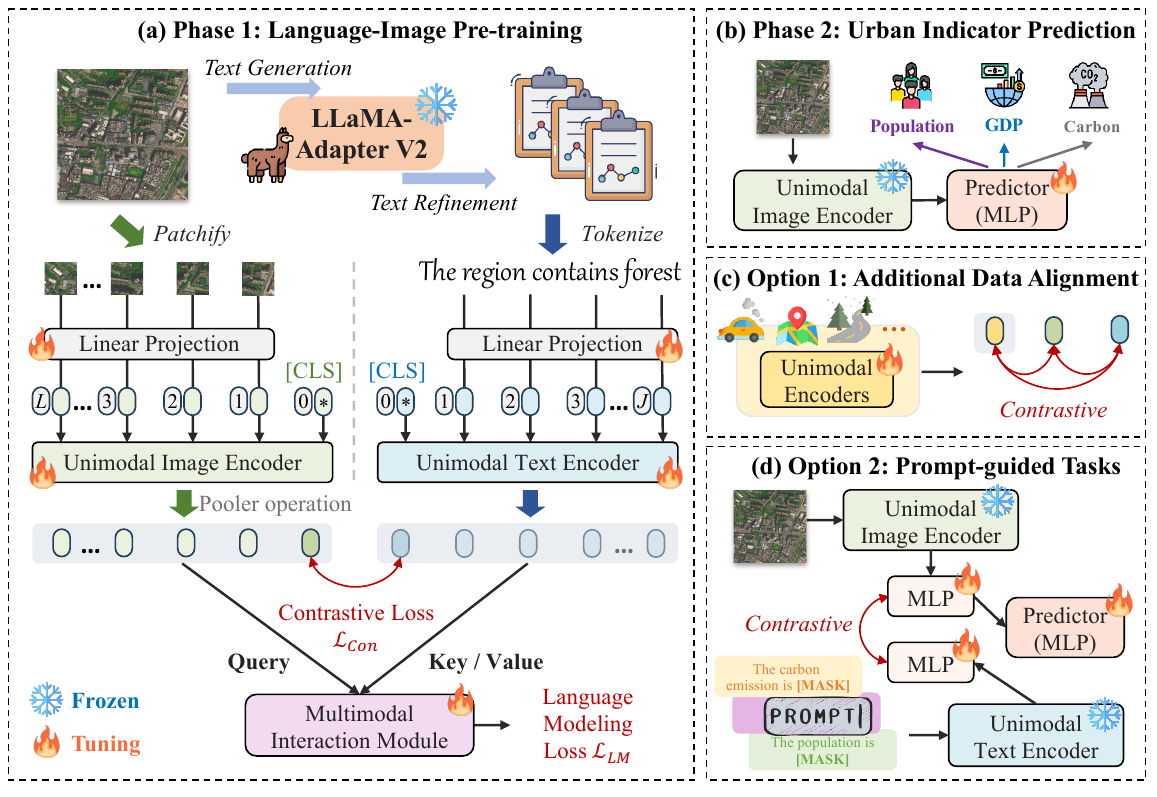}
  \vspace{-2em}
  \caption{The framework of UrbanCLIP, the first-ever LLM-enhanced framework that integrates the knowledge of textual modality into urban imagery profiling. It utilizes the CLIP rationale for language-image pertaining \cite{yan2023urban}.}
  \label{fig: contrast urbanclip}
\end{figure}

Besides, the CLIP-based paradigm has been explored in urban computing in recent years. For instance, \citet{liu2023knowledge} proposed the KnowCL model for socioeconomic prediction, which is the first solution that introduces the regional knowledge graph-based semantics and its associated imagery representation as contrastive pairs. Except for self-augmented contrast, \citet{bai2023geographic} also bridged the socio-economic semantic gap through inter-modal contrast between VHR images and POIs. The RecMVC model of \citet{zhangRegionEmbeddingIntra2023} included an inter-view contrastive learning module between POI and mobility, serving as a soft co-regularizer to transfer knowledge across multi-views. \citet{liu2023unified} designed a trajectory contrastive learning paradigm, where hub and link representations in the same trajectory are enforced to have a higher correlation with one another. Likewise, \citet{mao2022jointly} introduced road-trajectory cross-scale contrast to bridge the two scales by maximizing the total mutual information. This contrast is elaborately tailored via novel positive sampling and adaptive weighting strategies. Following the conventional CLIP setting, \citet{yan2023urban} leveraged satellite imagery and associated LLM-generated description as positive pairs, to learn a robust visual embedding for urban region profiling (depicted in Figure \ref{fig: contrast urbanclip}). Subsequent work, UrbanVLP~\cite{hao2024urbanvlp}, has broadened the contrastive learning paradigm to encompass multi-granularity visual clues, including satellite and street-view images. Concurrently, it also proposes effective methods to guarantee the generated text quality.

\subsection{Generation-based Data Fusion}

\label{sec:generative}
The primary objective of a generation-based model is to  produce the desired data based on specified input conditions.
Both input and output data may manifest in diverse formats
\cite{yan2023urban,zhu2023DiffTraj,zhang2023chattraffic}, 
rendering it a favorable choice for data fusion.
\add{By generating new content that correlates with input data, generative models are driven to discern intricate correspondences between multimodal information, thereby facilitating efficient information aggregation. However, these models are marked by significant computational complexity and considerable training challenges. Consequently, they are not suitable for tasks requiring high real-time performance. Instead, they excel with large datasets~\cite{kaplan2020scaling}, making them particularly well-suited for industries such as transportation, economics, and energy.}
Compared to the general Computer Vision (CV) or Natural Language Processing (NLP) area, 
the field of urban computing exhibits a relatively limited volume of generative research contributions.
However, in recent years, there has been a growing number of generative works emerging, particularly influenced by the rise of LLMs \cite{liu2023pre}. Based on the specific methods of generation, here we categorize the generation-based data fusion into four types, including autoregressive (Sec.\ref{sec:autoregressive}), mask modeling (Sec.\ref{sec:maskmodeling}), diffusion-based (Sec.\ref{sec:diffusion}), and LLM-enhanced models (Sec.\ref{sec:llm}).

\subsubsection{Autoregressive Model}
\label{sec:autoregressive}

The autoregressive model was first developed for the Language Modeling (LM) tasks in NLP, which predict future data based on historical data. 
Given a text sequence, $x_{1:T} = [x_1, x_2, ..., x_T]$, the learning objective of a language model is to maximize the probability of a sequence, which can be mathematically formulated as:
\begin{equation}
\mathcal{L}_{AR} = {\rm max}_\theta\sum{}{\rm log}P_\theta(x_t|x_{t-k},...,x_{t-1}),
\end{equation}
where $k$ is the size of the sliding window.

Gradually, the definition of input data and output data for autoregressive model has expanded from text to encompass multi-modal data. Owing to the multi-modal nature of the attention mechanism \cite{vaswani2017attention}, whereby queries, keys, and values can stem from different modalities, it has facilitated the exploration of multi-modal fusion.
VirTex \cite{desai2021virtex} proposed that compared with contrastive learning which uses classification labels as a learning signal, captions can provide a more semantically dense learning signal. 
SimVLM \cite{wang2021simvlm} reformulated the typical encoder-decoder architecture for end-to-end training with a single prefix language modeling objective, achieving the efficient utilization of weakly aligned image-text pairs.
CoCa \cite{yu2022coca} is a follow-up work to ALBEF \cite{li2021align} and SimVLM \cite{wang2021simvlm}, which creatively integrated attentional pooling methods in a creative manner, combining contrastive and generative approaches to achieve exceptional performance.

In the field of urban computing \cite{zheng2014urban}, the generative decoder has played a significant role, even before the era of transformer models \cite{vaswani2017attention}.
GeoMAN \cite{liang2018geoman} first introduced a multi-layer attention mechanism for spatio-temporal data prediction based on the encoder-decoder architecture. The generative decoder combined LSTM and Temporal Attention to predict the future performance of sensors. It achieved excellent performance in applications such as water quality prediction and air quality.
\citet{zhao2021bounding} implemented a bottom-up and top-down framework to do street-view image classification, which utilized RNN units to make predictions based on visual elements and contextual information.
Besides, in urban computing area \cite{zheng2014urban}, the limited availability and diverse nature of domain data, as well as the challenges in its acquisition, have led to increased interest in multi-modal pretraining in recent years.

\subsubsection{Mask Modeling-based Fusion}
\label{sec:maskmodeling}
The concept of masking modeling task has its origins in the field of NLP, contributing to the success of BERT \cite{devlin2018bert}.
MAE \cite{he2022masked} is the first work that introduces masking modeling into CV. Compared with masking modeling works in NLP, MAE \cite{he2022masked} has a larger mask ratio due to the fact that the images have less information density than natural language \cite{he2022masked}. 
GraphMAE \cite{hou2022graphmae} unleashes the power of mask modeling for graph structure. Different from most graph autoencoders' efforts in structure reconstruction, it proposes to focus on feature reconstruction with both a masking strategy and scaled cosine error.
In the context of mask modeling structure for multi-modal data fusion, various modalities are typically provided as input, followed by the masking of a portion or all of the modal data at varying proportions. Subsequently, information from other modalities is integrated to reconstruct the masked data \cite{kwon2022masked,geng2022multimodal,ding2023mgeo}.

\begin{figure}[!hb]
  \centering

  \includegraphics[width=0.48\textwidth]{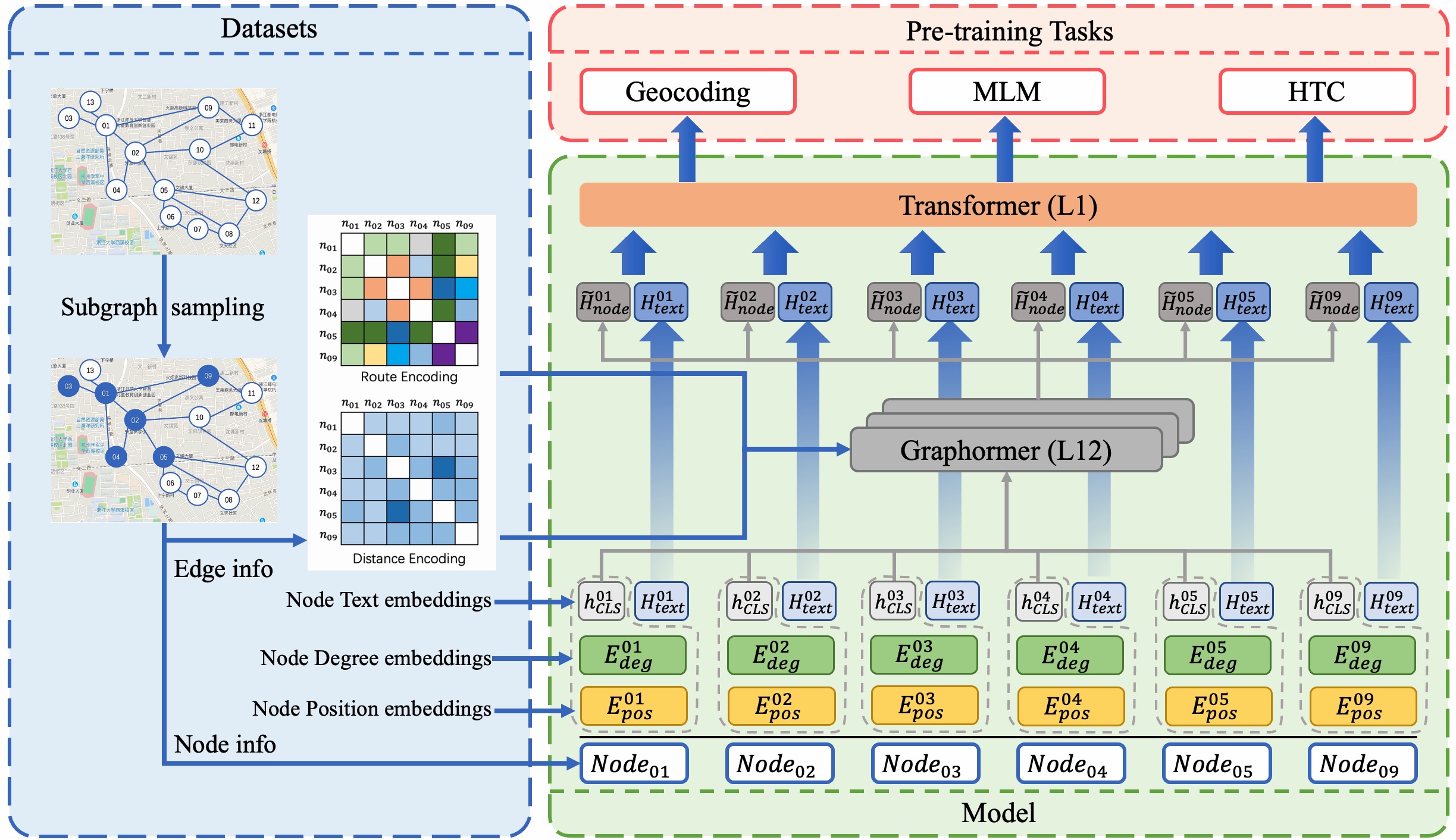}
  \vspace{-2em}
  \caption{The framework of G2PTL, a Geography-Graph Pre-Trained model for delivery address in the Logistics field, which uses the MLM task to learn the semantic information in the address \cite{wu2023g2ptl}.}
  \label{fig: generation g2ptl}
\end{figure}

MGeo \cite{ding2023mgeo} combined text and geolocation to implement location embedding for query-POI matching, which utilized two unimodal masked language modeling (MLM) loss and one Multi-Modal MLM loss to enforce information interaction.
As illustrated in Figure \ref{fig: generation g2ptl}, G2PTL \cite{wu2023g2ptl} is a graph-based pre-trained model for text-based delivery address embedding.
It leveraged MLM as one of the pre-training tasks to simulate input noise in missing or incorrect address cases in real-world scenarios.
QUERT \cite{xie2023quert} adapted pre-trained language models to specific Travel Domain Search applications, which designed a Geography-aware masking strategy to force the pre-trained model to pay more attention to the geographical location phrases.
ERNIE-GeoL \cite{huang2022ernie} was dedicated to integrating toponym knowledge and spatial knowledge. By utilizing a Masked Language Modeling training procedure, it can acquire four types of toponym knowledge, including relationships between POI name, address, and type.

In the remote sensing area, despite the similarities between remote sensing images and RGB images, they also possess distinctions, with remote sensing images containing geographical information and temporal tags, as well as spectral multiple channels, and diverse resolutions and scales.
SatMAE \cite{cong2022satmae} discussed the domain-specific temporal and spectral characteristics of satellite imagery,
and device's temporal encoding and spectral positional encoding respectively to adapt MAE structure to satellite imagery representation learning. 
Scale-MAE \cite{reed2023scale} was designed to address another challenge in remote sensing satellite imagery: the variation in image scale due to data from multi-scale sensors.
The pretraining process aims to learn a better representation of multiscale tasks by reconstructing low and high
frequency features at different scales.

\subsubsection{Diffusion-based Fusion}
\label{sec:diffusion}

In recent years, diffusion models \cite{yang2023diffusion}, as an emerging and potent type of deep generative models, have demonstrated state-of-the-art performance across various modalities such as images, speech, and video \cite{cao2022survey}.
Diffusion models can serve as an appropriate framework for data fusion, enabling the seamless incorporation of new modalities due to the existence of conditions \cite{zheng2023diffuflow,khanna2023diffusionsat}. Simultaneously, the integration of multiple modalities can also enhance the quality of generation.

Urban imagery, as the primary provider of visual information, has borrowed numerous techniques from the general computer vision area and achieved successful applications.
DiffusionSat \cite{khanna2023diffusionsat}, inspired by Stable Diffusion \cite{rombach2022high} and ControlNet \cite{zhang2023adding}, provided the first large-scale generative foundation model for satellite imagery. It combined data from different modalities including geospatial metadata (latitude, longitude,ground-sampling distance), timestamp, and texts, achieving promising performance on a series of tasks such as super-resolution, temporal generation, and in-painting.
The generation of street-view images \cite{gao2023magicdrive,wang2024customizing} for spatio-temporal applications, however, remains an underexplored area, which represents a promising research direction in the future.

Aside from common modalities like image, text, and audio, the spatio-temporal modality data covers a broader range of data formats in urban computing, including traffic situations \cite{zhang2023chattraffic}, urban flow \cite{zheng2023diffuflow,xu2023diffusion,wen2023diffstg}, and trajectories \cite{zhu2023DiffTraj,yuan2022activity}. In this context, the flexibility of diffusion models allows them to potentially exert a more considerable influence.

ChatTraffic \cite{zhang2023chattraffic} creatively introduced fine-grained text in the Text-to-Traffic Generation (TTG) task to adapt to unusual events. Besides, when enhanced with GCN to incorporate the spatial information inherent in the road network, it achieved more accurate and realistic long-term prediction.

\begin{figure*}[ht!]
    \centering
    \includegraphics[width=0.85\textwidth]{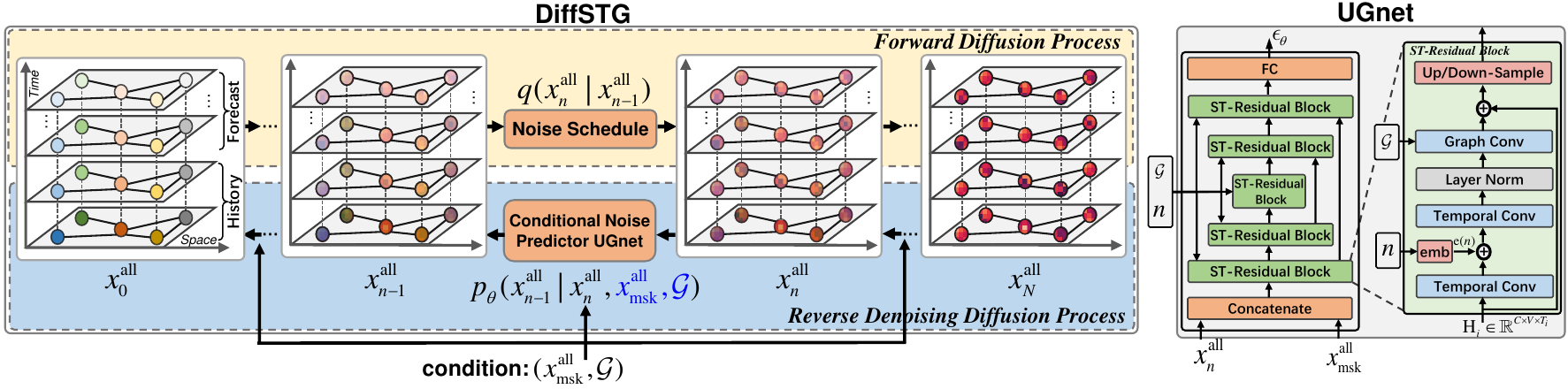}
    \caption{Illustration of DiffSTG and denoising network UGnet, which leverages an Unet-based architecture to capture multi-scale temporal dependencies and the Graph Neural Network (GNN) to model spatial correlations \cite{wen2023diffstg}.}
    \label{fig: generation diffstg}
\end{figure*}

DiffSTG \cite{wen2023diffstg} is the first work that generalizes the diffusion model DDPM to spatio-temporal graphs. As illustrated in Figure \ref{fig: generation diffstg}, it combined the spatio-temporal learning capabilities of STGNNs with the uncertainty measurements of diffusion models. The collection of fine-grained urban flow data is widely recognized as challenging due to the high costs associated with deployment and maintenance, as well as the presence of noise.
Diffusion models are suitable tools to generate fine-grained flow maps from the coarse-grained ones \cite{zheng2023diffuflow,xu2023diffusion}.
DiffUFlow \cite{zheng2023diffuflow} is the first generative approach for fine-grained urban flow inference. An ELFetcher module is proposed to utilize various external factors and land features as conditional guidance for the reverse denoising process. 

Trajectory data represents another essential type of spatio-temporal data. However, it often raises privacy concerns due to the inclusion of personal geolocation information. One promising solution for this challenge is trajectory generation \cite{zhu2023DiffTraj,yuan2022activity}, which aims to generate high-fidelity, privacy-free trajectories.
The diffusion model, as a more reliable and robust method of generation than canonical methods, begins to be increasingly explored.
DiffTraj \cite{zhu2023DiffTraj} is the first exploration of trajectory generation by the diffusion model, which proposed a Traj-UNet architecture to predict the noise of each diffusion time step. It proved that the step-by-step denoising process of the diffusion model is an appropriate choice due to the stochastic and uncertain characteristics of human activities.

\subsubsection{LLM-enhanced Data Fusion}
\label{sec:llm}

\begin{figure}[!htb]
    \centering
    \includegraphics[width=0.5\textwidth]{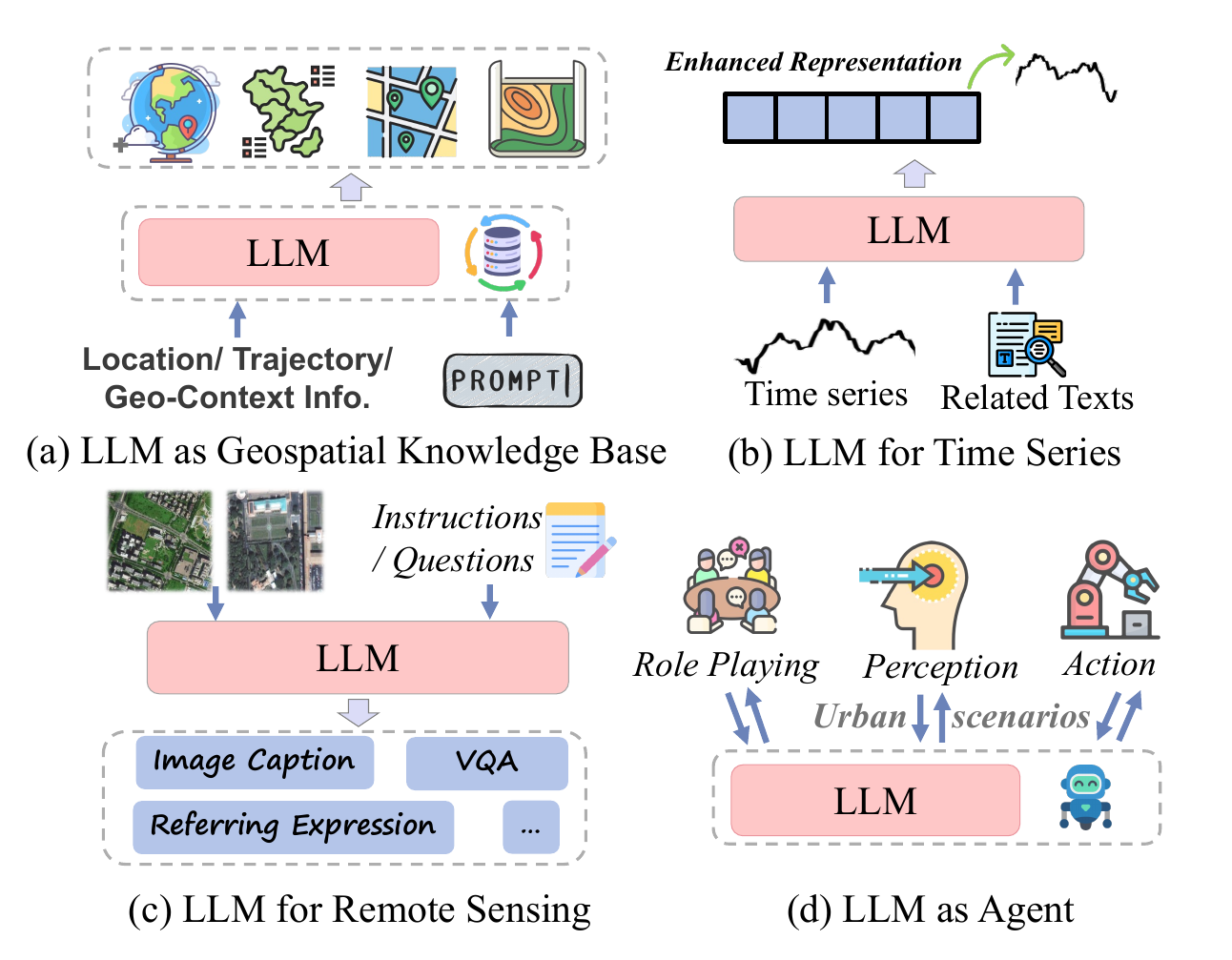}
    \caption{\add{Categories of LLM-enhanced data fusion.}}
    \label{fig:llm}
\end{figure}

With the rise of GPTs \cite{brown2020language,ouyang2022training,bubeck2023sparks}, the remarkable capabilities of LLMs across a broad spectrum of fields have garnered significant attention, sparking a wave of research paradigm shift and shedding light on artificial general intelligence (AGI).
The academia usually terms “large language models (LLM)” for these large-sized PLMs \cite{zhao2023survey} due to the scaling laws \cite{kaplan2020scaling} of Transformer \cite{vaswani2017attention} architecture.
LLM-enhanced data fusion can be considered as a specific instance of encoder-based alignment, employing LLMs for feature encoding and information interaction. The extensive parameter size of LLMs endows it with potent alignment capabilities.
LLMs have multifaceted impacts on the field of urban computing, including works that utilize LLMs' geospatial capabilities \cite{roberts2023gpt4geo,manvi2023geollm,wang2023would,xue2022leveraging}, applications in remote sensing \cite{hu2023rsgpt,kuckreja2023geochat}, works in the time series domain \cite{chang2023llm4ts,jin2023time,cao2023tempo,liu2023unitime,zhou2023one}, etc, as illustrated in Figure~\ref{fig:llm}.

LLMs are able to compress and store geospatial knowledge within the training data.
Exploring how to effectively utilize this compressed knowledge and design prompts that stimulate it are areas of current research interest \cite{roberts2023gpt4geo,manvi2023geollm,wang2023would,xue2022leveraging}.
GPT4GEO \cite{roberts2023gpt4geo} provided a comprehensive investigation of the extent to which GPT-4 \cite{bubeck2023sparks} has mastered factual geographic knowledge and its ability to use this knowledge for reasoning.
GeoLLM \cite{manvi2023geollm} proposed that constructing the right prompt is key to extracting geospatial knowledge. Through providing geo-context information near a specific location, LLMs can be fine-tuned to achieve state-of-the-art performance on a variety of large-scale geospatial datasets for tasks such as predicting population density, house price, women’s education, etc.
LLM-Mob \cite{wang2023would} demonstrated the first attempt to apply LLM to modeling human mobility, which reformulated mobility data through historical stays and context stays to introduce long-term and short-term dependencies for prediction and reasoning.

Large Multi-modal Models (LMMs) \cite{dai2305instructblip,liu2023improved,ye2023mplug,zhang2023internlmxcomposer} demonstrate substantial efficacy in information fusion and modal alignment. In the domain of urban computing, the utilization of remote sensing imagery is prevalent for integrating visual elements from a bird's-eye view perspective.
Inspired by InstructBLIP \cite{dai2305instructblip}, RSGPT \cite{hu2023rsgpt} utilized frozen Image Encoder and LLM, while training a lightweight Q-Former \cite{li2023blip} to align the two, achieving state-of-the-art remote sensing image captioning and remote sensing visual question answering downstream tasks.
Developed based on LLaVA-v1.5 \cite{liu2023improved}, GeoChat \cite{kuckreja2023geochat} introduced multi-modal instruction-tuning into the remote sensing domain and proposed the first versatile remote sensing Large Vision-Language Model with multitask conversational capabilities.

\begin{figure}[!h]
    \centering
    \includegraphics[width=0.48\textwidth]
    {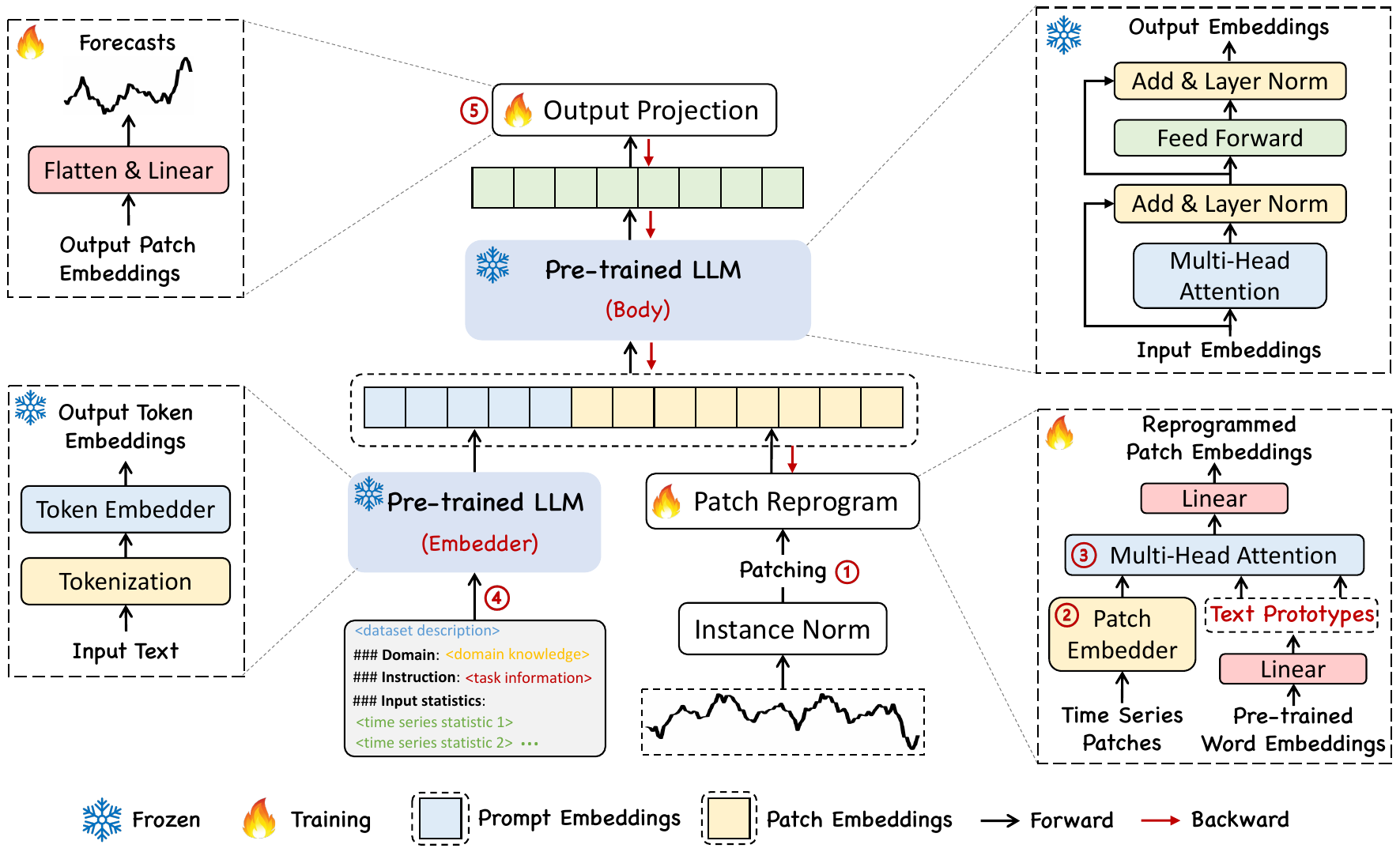}
    \caption{Time-LLM begins by reprogramming the input time series with text prototypes before feeding it into frozen LLM to align the two modalities \cite{jin2023time}.}
    \label{fig: generation timellm}
\end{figure}

LLMs have the potential to revolutionize time series analysis~\cite{jin2024position,liang2024foundation}. Following the success of large foundation models in NLP and CV, there are desires in the time series forecasting area to utilize pre-trained LLMs as powerful representation learners \cite{chang2023llm4ts,jin2023time,cao2023tempo,liu2023unitime,zhou2023one}.
LLM4TS \cite{chang2023llm4ts} combined patching and channel-independence techniques with temporal encoding, which unlocked the flexibility of pre-trained LLMs without introducing large parameter overhead.
As depicted in Figure \ref{fig: generation timellm}, Time-LLM \cite{jin2023time} solved the alignment of time series data and natural language modality to unleash the power of LLMs for time series forecasting through Prompt-as-Prefix or Patch-as-Prefix, which reprogrammed time series into text prototype representations.

\add{LLMs approach real-world cross-domain data fusion in the form of agents~\cite{jin2024position,xi2023rise}. In human perception of the world, diverse modalities ultimately converge into language, serving as the medium for expression and communication \cite{shao2023allspark}.
Based on essential language capabilities, LLMs can store knowledge and process cross-modal information, thereby continuously receiving feedback and interacting with the environment, demonstrating potential in spatiotemporal domains for comprehensive multimodal data integration.}
\citet{zhou2024large} first explores urban planning through multi-agent collaboration framework, partially demonstrated performance surpassing that of human experts. LLMLight~\cite{lai2023large} transcends the previous paradigm of using LLMs as assistants to enhance decision-making, by directly employing LLMs as Traffic Signal Control (TSC) agents for decision-making.

\section{Application Perspective}
\label{ApplicationSection}
\begin{figure*}[!t]
    \centering
    \includegraphics[width=0.95\textwidth]{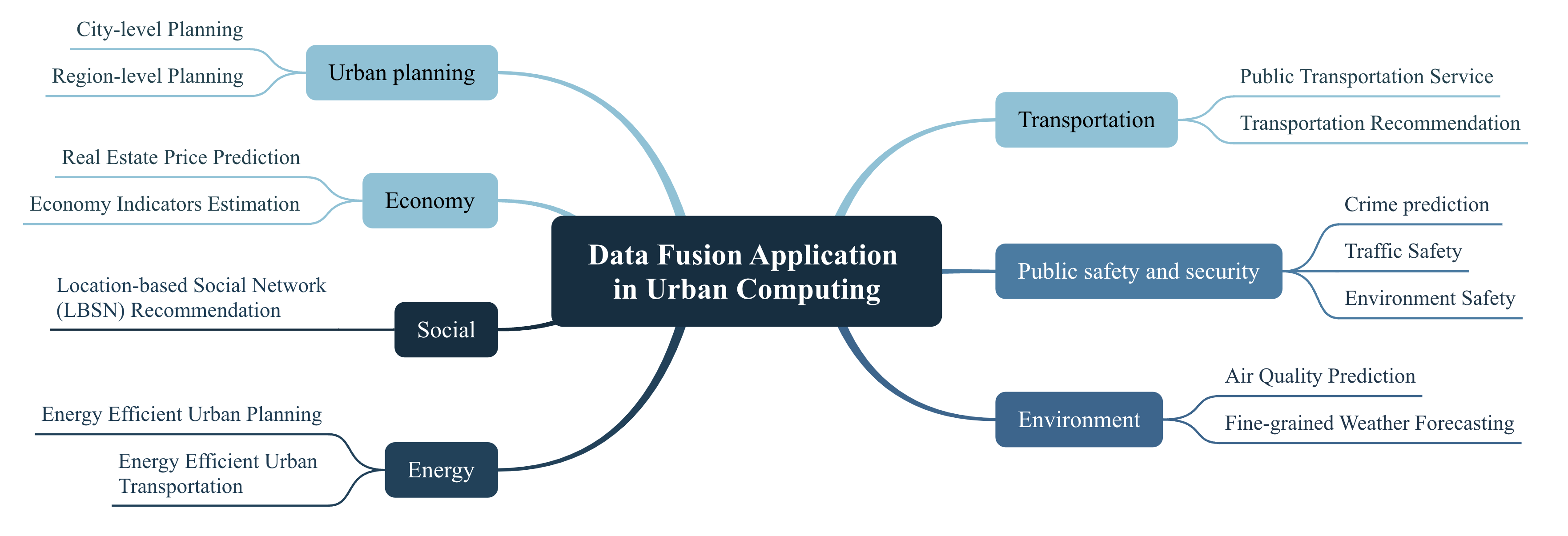}
    \caption{Taxonomy of application (category) and common downstream tasks (sub-category) for cross-domain data fusion in urban computing.}
    \label{fig:tax-application}
\end{figure*}

In this section, we categorize and summarize related works from the application perspective. Figure \ref{fig:tax-application} illustrates applications from seven domains, including urban planning, transportation, economy, public safety and security, society, environment, and energy. Subsequently, we provide a comprehensive exposition of these applications, delving into intricate details, while also highlighting the pivotal role that deep learning-based data fusion methods have played in facilitating these tasks.

\subsection{Urban Planning}
Sensing a city and making effective planning and governing is of great importance to its development. Every political planning and strategy necessarily needs to be formulated under strong computational support which covers lots of factors, such as road network structures, geographical limitations, transportation situation, human mobility, and society. In earlier years, to formulate a decent planning or strategy for a city, planners always need to sensor the city through various labor-intensive surveys. Besides, the processing of these surveys from different sources was also tough to conduct. 

\begin{figure}[!h]
    \centering
    \includegraphics[width=0.42\textwidth]{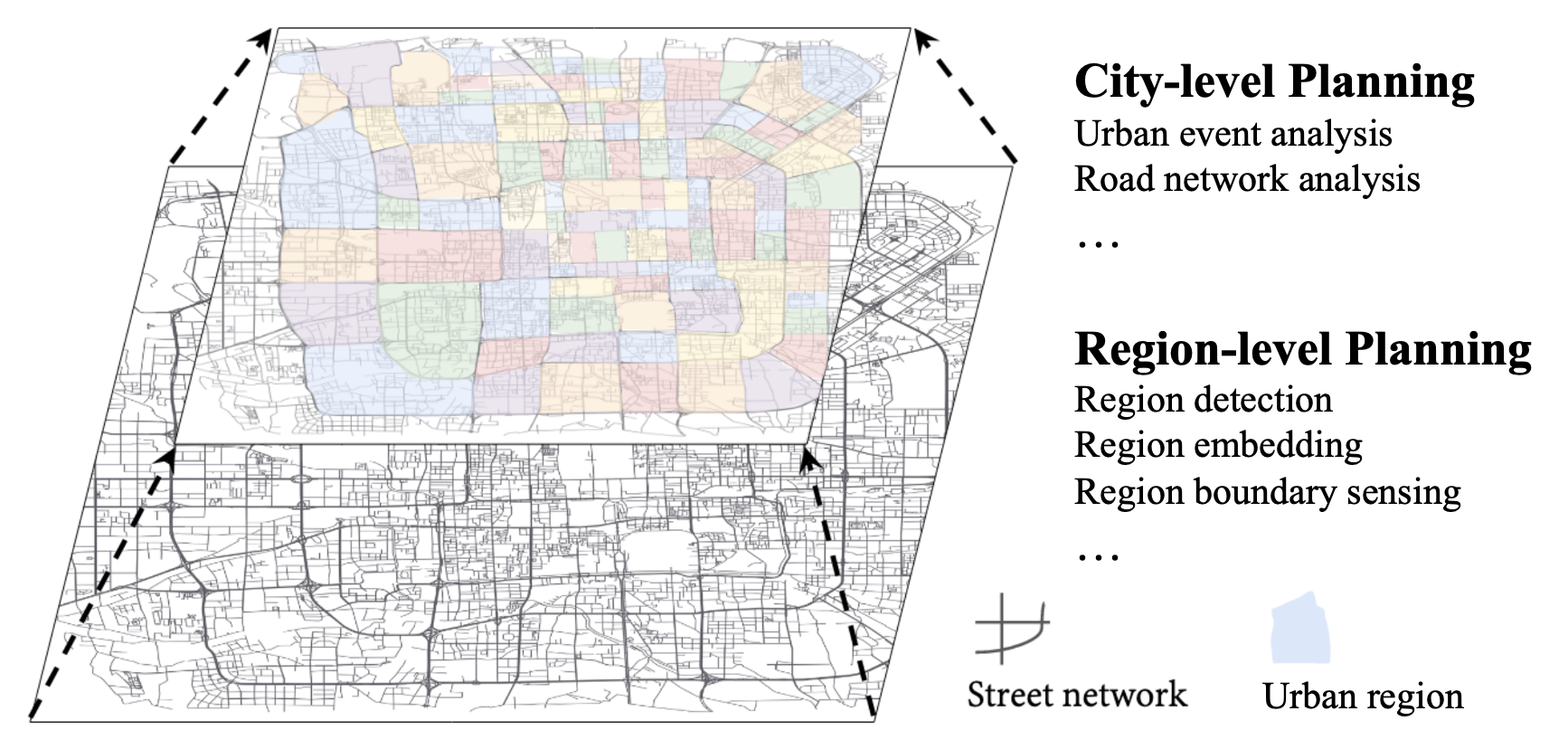}
    \caption{A city exhibits a multi-level structure for urban planning: city-level planning and region-level planning \cite{li2022predicting}.}
    \label{fig:urbanplanning}
\end{figure}

With the development of city infrastructures, most urban data and factors can be directly collected from various sensors and platforms. However, processing and understanding such big data from multiple sources becomes a challenge due to the poor performance of traditional data analysis methods on big data. Deep learning models provide researchers with opportunities to handle big and multi-source data with better efficiency and deeper understanding. Therefore, for most urban planning tasks leveraging the advantages of multi-source data, deep learning-based data fusion methods have become a strong support from urban planners in both city-level planning and region-level planning (shown in Figure \ref{fig:urbanplanning}).

\subsubsection{City-level Planning}
Multi-sources urban data provides comprehensive information about the operation and variation of cities. Deep learning-based data fusion could utilize the data to sense the city-level variation and understand it as well. This enables urban planners and researchers to understand urban dynamics, such as sensing social events \cite{jiang2019deepurbanevent, zhaoSpatiotemporalEventForecasting2022,guoNonparametricModelEvent2016}, region's prosperity, city's vibrancy \cite{liu2020urban} and any important change happened to the city \cite{balsebreGeospatialEntityResolution2022}. To sense city-level dynamics, \citet{liu2020urban} conducted research for understanding and predicting urban vibrancy evolution based on three data sources: mobile check-in data, POI data, and geographical data. \citet{balsebreGeospatialEntityResolution2022} proposed a framework named \textit{Geo-ER} to match geo-spatial entities from various data sources. 
\cite{zheng2023road,zheng2023spatial} introduced deep reinforcement learning to autonomously generate road layouts, aiming to connect various locations at the lowest construction expense.


\textbf{Urban event analysis} is crucial for urban planning as variations around a city usually come with urban events.  \citet{10.1145/2783258.2783377} proposed a multi-task learning framework to sense and predict urban events from the variation on social media. This framework could effectively train forecasting models for multiple locations simultaneously with shared information by restricting all positions to select a common set of features. \citet{jiang2019deepurbanevent} proposed a recurrent neural network (RNN) based online system named \textit{DeepUrbanEvent} to understand the crowd dynamic variation for social events. By extracting the deep trend from the current momentary observations on the historical GPS data from citizens, this system could generate an effective prediction for the crowd dynamic trend during a short future time at a big event in a city.

Focusing on the advantages of multi-source data, \citet{zhaoSpatiotemporalEventForecasting2022} summarized the previous research in event prediction and stressed four challenges for variation understanding and event prediction from multi-source data: 1) geographical hierarchies; 2) hierarchical missing; 3) feature sparsity; 4) difficulty in update with incomplete multiple data. Subsequently, they proposed a multi-source feature learning model based on an \textit{N}th-order geo-hierarchy and fused-overlapping group Lasso to handle these challenges. In their methodology, models can be instantly updated from new data from every data source with affordable computational cost, so the urban planner can respond to variation faster with this system.

\textbf{Road network analysis} plays a crucial role in the realm of urban planning, demanding significant attention from professionals in the field. The development of cities leads to continuous variation in the road conditions around a city. It is noteworthy that traditional road maps are unable to cope with these variations. Additionally, conventional surveying methods and database management techniques are inadequate for capturing and managing the rapidly evolving road networks. 
\citet{yinMultitaskLearningFramework2020a} pioneered the utilization of multi-source information for the automatic derivation of road attributes.
They decided on a multi-task learning framework to extract low-level feature embedding from every data source and applied attention-based fusion to fuse the representations. Based on this work, they could combine the information from GPS trajectory and existing map data and achieve a significant improvement in classification on the OpenStreetMap data in Singapore. Based on the conception of data fusion, \citet{yangDuAREAutomaticRoad2022} proposed a \textit{DuARE} system for large-scale automatic extraction by leveraging the GPS trajectory and the satellite images. Driven by its promising performance, DuARE has been deployed in China and has been updating the national road network by 100,000 km every month.

\subsubsection{Region-level Planning}

Cities serve as the foundation for societal functions in modern society. Based on the planning efforts of urban managers or the natural social operating principles, cities always develop into distinct functional regions, such as industrial zones, residential areas, and financial districts. These regions are sometimes referred to as Areas of Interest (AOIs). Simultaneously, within these areas, there are diverse locations that cater to the specific needs and demands of residents, commonly known as Points of Interest (POIs). These functional regions and points fulfill the daily requirements of citizens, enabling efficient urban operations. Accurately perceiving the status of AOIs and POIs in a city and understanding them is crucial for urban planners and administrators in making informed decisions. 
This encompasses activities such as the selection of appropriate park locations, the planning of transportation routes, and the development of regional policies.

\textbf{Region detection} is the first step for us to understand region-level information. For POI or AOI detection, \citet{xiaoContextualMasterslaveFramework2023} proposed a \textit{Contextual Master-Slave Framework (CMSF)} that unitizing graph neural network to fusion the POI information and satellite image to detect urban villages. \citet{huang2023comprehensive} fused satellite image, and street-level image with mobility data in their study, and the extra vision data were embedded through a \textit{Vision-LSTM} network and were demonstrated crucial for region detection through comprehensive ablation study. Their work achieved an overall accuracy of 91.6\% in identifying urban villages in Shenzhen, China. \citet{zhaoAnnotatingPointsInterest2016a} carried out a research on integrating information from social media such as Twitter as well as real-world locations. They designed a supervised Bayesian Model (sBM) to analyze the textual information, spatial features, POI information, and user behaviors. Based on this work, researchers are able to find user interests in special regions and understand the regions' properties.

\begin{figure}[htbp]
    \centering
    \includegraphics[width=0.45\textwidth]{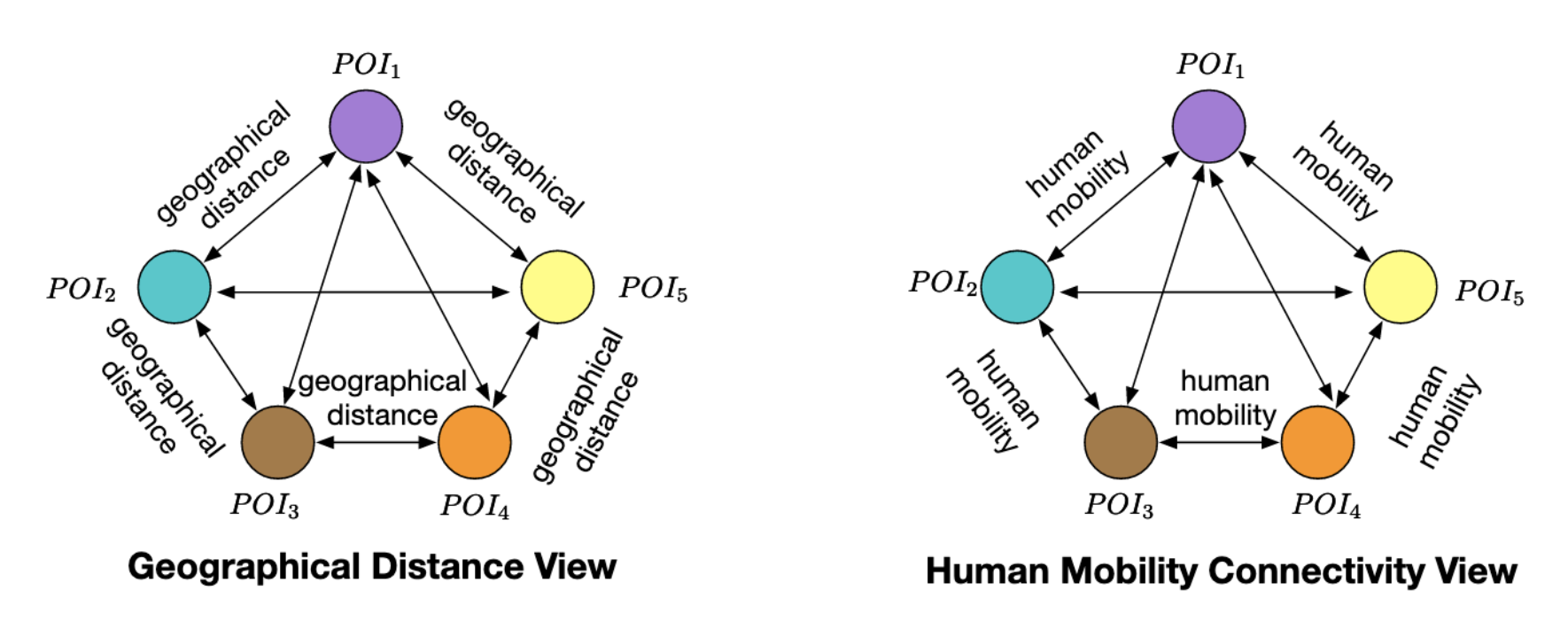}
    \vspace{-1.5em}
    \caption{An example of the multi-view POI graph networks proposed by \citet{fu2019efficient}.}
    \label{fig:regionembedding}
\end{figure}

\textbf{Region embedding} based on multi-source data is the foundation of region-level urban computing. \citet{fu2019efficient} proposed a multi-view POI network by varying the representation of edges to fusion the geographical distance information and human modality. There are 2 kinds of POI-POI graph networks (shown in Figure \ref{fig:regionembedding}) in this specialized multi-view framework: 1) distance-based POI-POI networks: \textit{the weight of an edge represents the distance between two associated POIs}; 2)mobility connectivity-based POI-POI networks:\textit{ the weight of an edge represents the mobility connectivity between two associated POIs.} \citet{zhangMultiviewJointGraph2021a} fused human mobility information with region attributes information with a multi-view joint learning module to infer the land usage of a city and \citet{zhangRegionEmbeddingIntra2023} designed a  multi-task learning framework called \textit{ReMVC} to classify regions by land usage and \citet{liUrbanRegionRepresentation2023} conducted a research on fusion publicly available building footprint information on \textit{OpenStreetMap} and POI data for region representation learning. They carried out an experiment in Singapore and New York and demonstrated its efficiency. 

\citet{bai2023geographic} also carried out research on geographic representation learning. In their methodology, satellite images and POI information are fused to solve multiple geographic mapping tasks. Besides, \citet{bingPretrainedSemanticEmbeddings2023} presented a POI embedding method named CatEM, which jointly considered the spatial information and mobility information of POIs and achieved an impressive performance on POI classification tasks. Through efficient region embedding, \citet{wang2021deep,wang2023human} proposed a generative framework to generate land-use configuration of a trageted region for urban planning.

\textbf{Region boundary sensing} plays a crucial role in region-level urban planning as it encompasses not only the political boundaries of the city but also functional or cultural regions within the city. Traditional administrative boundaries, often determined by policies and planning, may not accurately reflect the actual natural boundaries of a city. 

Only a few deep learning researches were proposed for region boundary sensing in recent years \cite{wangMappingUrbanBoundary2020,chenCrosscityFederatedTransfer2022,chenUVLensUrbanVillage2021,vu2016geosocialbound}. The natural boundaries of a city are constantly evolving and adapt to the real needs and functions of the urban area. In fact, accurately sensing these boundaries is crucial for formulating regional policies and adjusting administrative planning to align with the true dynamics of the city. \citet{wangMappingUrbanBoundary2020} conducted a study on mapping the urban boundary of Zhengzhou City, China with its satellite images and POI data and found the result is in great agreement with the boundary ground truth. 

\citet{chenCrosscityFederatedTransfer2022} designed a cross-city federated transfer learning framework named \textit{CcFTL} that can deal with multi-source data including POIs, road networks, and population density. In their methodology, this framework could not only deal with information from a single city but also transfer within various cities with great robustness. Besides,  \citet{chenUVLensUrbanVillage2021} proposed a boundary sensing framework for urban villages that fuse the information from satellite images, mobility from bike-sharing drop-off data, and POIs. This platform was successfully deployed on the government data platform of Xiamen City, China to serve both urban planners and citizens. Some researchers focused on sensing the boundaries of POIs, \citet{vu2016geosocialbound} proposed a boundary estimation frameworks by pairing the geo-tagged text information on tweets with the real name of POI and analyzing their spatial correlations. This work demonstrated that the spatial distribution of relevant tweets on the platform could reflect the social boundary of the targeted POI.

\subsection{Transportation}
Transportation serves as the arteries of a city, acting as bridges that connect different urban entities \cite{doi2015cities}. As a result, the comprehensive utilization of multi-modal data focuses a significant proportion of attention on downstream transportation-related tasks \cite{miller2016public,bwire2020comparison,sinha2019sustainable,10.1145/3532611}.

\subsubsection{Public Transportation Service}
The advancement of a city is primarily demonstrated by enhancements in its transportation systems \cite{doi2015cities, guo2020context, dai2015personalized}.
Public transportation, in comparison to private transportation, is recognized for its superior environmental sustainability and energy efficiency \cite{miller2016public}.
Guided by government supervision, public transportation embodies the features of public goods.
In addition to fundamental passenger transportation, there are also freight transportation services, along with emerging services such as food delivery and ride-hailing, all of which contribute to enhancing urban convenience.
Different from \cite{zheng2014urban}, where transportation systems were categorized based on the type of vehicles used, this classification of services is based on their specific application scenarios. 
Public transportation applications can primarily be categorized into three main groups: Safety, Flow Control, and Efficiency.

\begin{figure}[!h]
    \centering
    \includegraphics[width=0.45\textwidth]{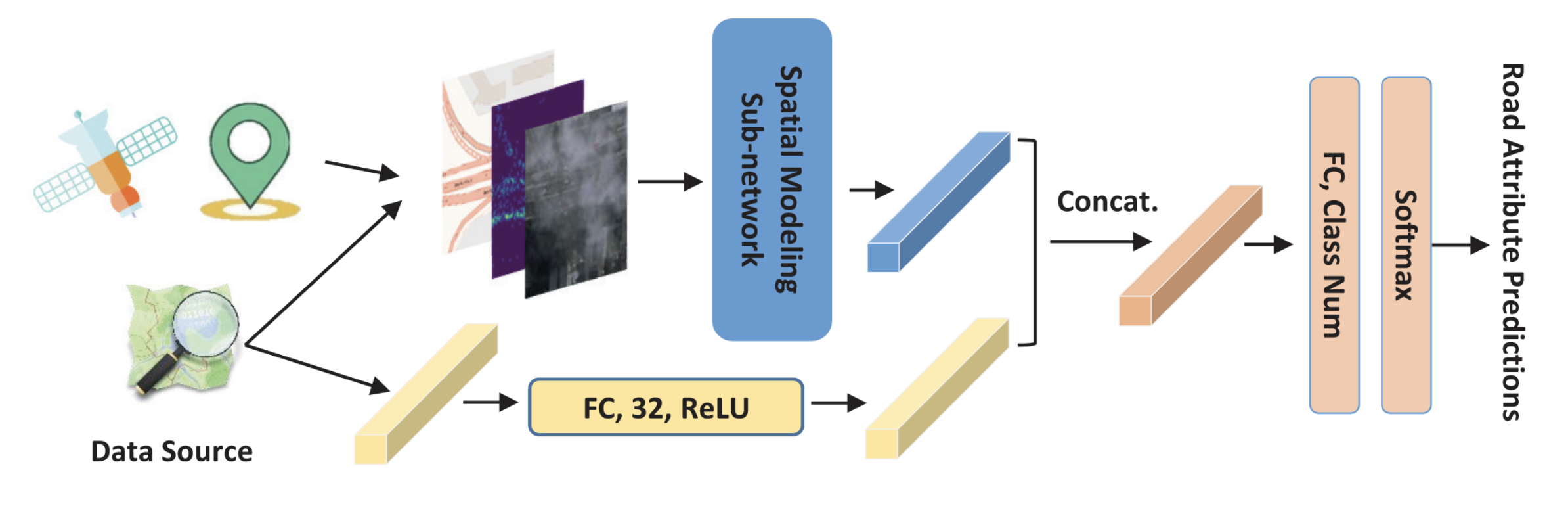}
    \vspace{-1em}
    \caption{Network architecture of the proposed multi-modal fusion network by \citet{yin2023multimodal} for robust road attribute detection.}
    \label{fig:traffic2}
\end{figure}

\textbf{Safety} is the most fundamental and urgent requirement. Consequently, roads, being high-risk areas for public transportation safety, have prompted numerous research efforts focusing on road safety.
PANDA \cite{you2022panda} predicted road risks after natural disasters by combining trajectories data and road event data.
To identify road obstacles such as fallen trees and ponding water, RADAR \cite{chen2018radar}  utilized co-training and active learning to fuse the heterogeneous features from trajectory data and environmental sensing data.
\citet{yin2023multimodal} proposed a multi-modal fusion network for inferencing missing road attributes. In their methodology (shown in Figure \ref{fig:traffic2}), the robustness of the network is largely enhanced by the pixel-level fusion of GPS traces and satellite images.

With the rapid urbanization of numerous countries, abnormal traffic incidents, have emerged as a substantial threat to public health and development. \cite{wang2021traffic} used a Multi-View Multi-Task Spatio-Temporal Network to capture the various context features such as weather, POIs, and road conditions, which result in the occurrence of traffic accidents. The multi-task learning framework is effective in modeling features of different granularity levels, thanks to the tie of spatial associations.

Human factors play a critical role in transportation safety and can be considered a primary factor in accidents \cite{negash2023driver,mozaffari2020deep}.
SAX-DF \cite{liu2022symbolic} is a Symbolic Aggregate approximation Data Fusion model to comprehensively utilize multi-source and heterogeneous data, which can effectively improve the driver behavior detection performance with the help of a Positive Danger Mapping algorithm.
The rise of autonomous driving has stimulated the demand for real-time driver behavior detection. \citet{huang2022real} proposed a deformable inverted residual network to adaptively detect real-time driver behavior. 

\textbf{Flow control} is an important downstream application that holds great significance for public policies such as resource scheduling and urban planning \cite{gerla1980flow}.
For instance, utilizing urban flow data enables governments to implement flow control measures during events such as New Year's Eve at Shanghai's Bund or Times Square in New York.
By managing the crowd flow and directing subway passengers to nearby stations, the government can prevent potential dangers and promote efficient social functioning, thus yielding social benefits. 
Private ride-hailing platforms like DiDi and Uber allocate more transportation resources to densely populated areas, which allows them to gain better economic benefits.

Indeed, flow control can be broadly categorized into two main areas: i) Traffic Flow Control \cite{bellemans2002models}, which involves regulating and optimizing the flow of vehicles on roads and highways.
ii) Mobility Prediction \cite{zhang2018mobility,barbosa2018human}, which endeavors to forecast and comprehend patterns of mobility behavior exhibited by humans or vehicles, is intended to enhance the planning and management of urban transportation and resources.

DeepSTN+ \cite{lin2019deepstn+} utilized a spatio-temporal network to predict inflow and outflow regarding POIs and historical data. 
To infer urban flow, UrbanSTC \cite{qu2022forecasting} implemented contrastive pretraining in both spatial and temporal to learn robust features in both two modalities with different pretasks.
Likewise, CSST \cite{ke2023spatio} made use of a pretrain-finetuning paradigm to align low-quality GPS reports and external factors with the crowd flow.

For mobility prediction,
\cite{miyazawa2019integrating} combined GPS traces and geo-tagged tweets to model human crowd flow.
GraphTUL \cite{gao2022contextual} solved the Trajectory user linking (TUL) problem by training an adversarial network in a semi-supervised way.

\textbf{Efficiency} primarily involves the promotion of efficiency in the allocation of transportation resources. It mainly relates to taxi/ride-hailing demand prediction, delivery time prediction, and passenger demand prediction.

\begin{figure}[!hb]
    \centering
    \includegraphics[width=0.48\textwidth]{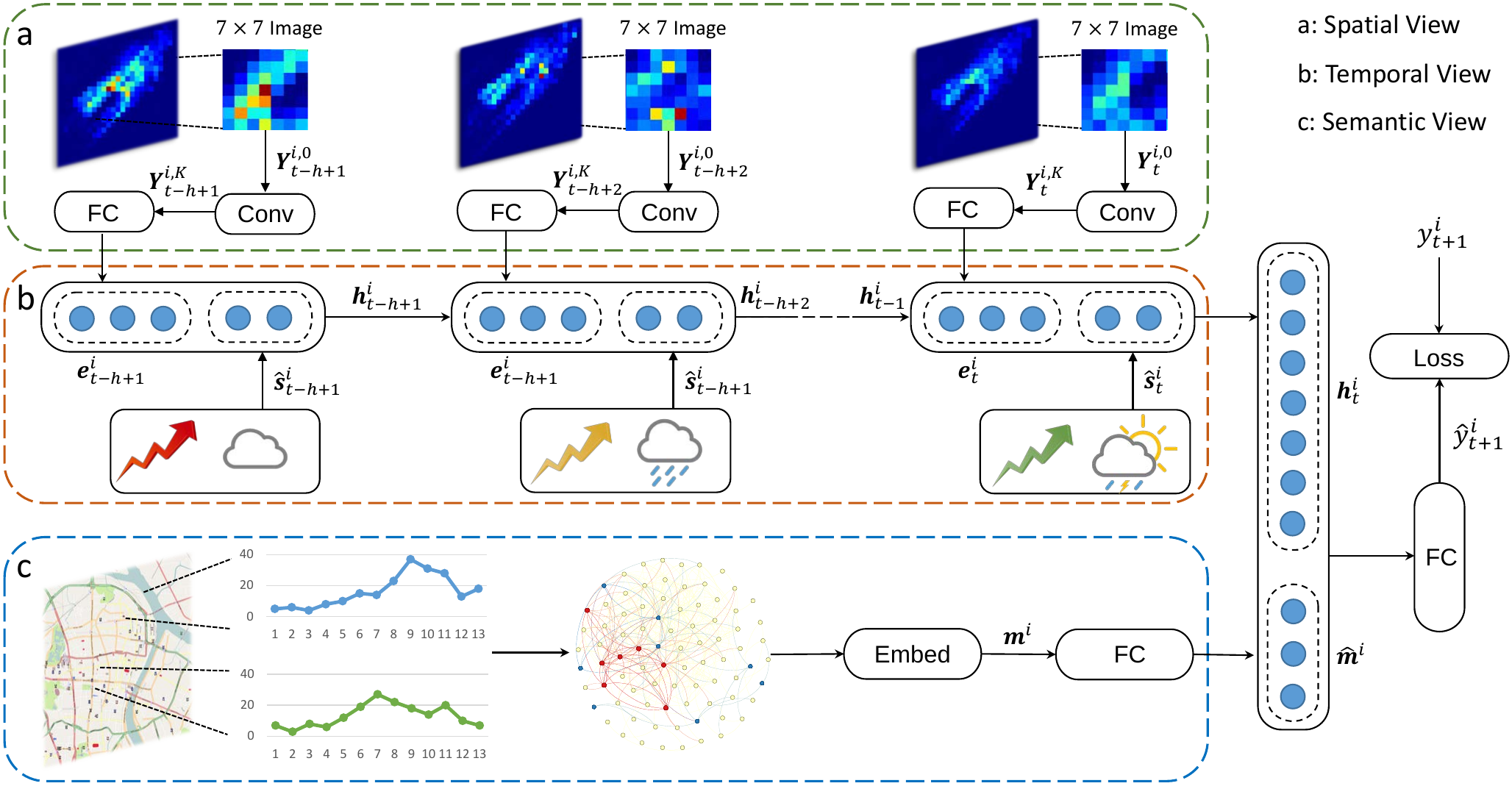}
    \caption{The architecture of DMVST-Net \cite{yao2018deep}, which is designed to incorporate spatial, temporal, and semantic components for taxi demand prediction.}
    \label{fig:efficiency}
\end{figure}

DMVST-Net \cite{yao2018deep}, shown in Figure \ref{fig:efficiency}, combined multi-view information to implement taxi demand prediction. It splits the process into spatial, temporal, and semantic views, which are modeled by local CNN, LSTM, and semantic graph embedding, respectively.
To address the challenge of predicting travel demands in city regions for future time intervals demands, DeepTP \cite{yuan2021effective} proposed to encode and capture three key properties from traffic data: Region-Level Correlations, Temporal Periodicity, and Inter-Traffic Correlations.
To jointly predict demands for various ride-hailing service modes like solo and shared rides, \citet{ke2021joint} combined multi-graph convolutional networks with two multi-task learning structures, enhancing prediction accuracy for diverse ride-hailing services in urban areas.
For delivery tasks, MetaSTP \cite{ruan2022service} introduced a meta-learning-based neural network model for predicting service time in last-mile delivery. Utilizing a Transformer-based layer and location prior knowledge, MetaSTP addresses complex delivery scenarios, showing significant improvements in prediction accuracy and practical deployment in JD Logistics.

The prediction of passenger demand is also of great significance, which can assist transportation departments and related enterprises in enhancing the planning and management of passenger traffic.
\citet{bai2019spatio} proposed a deep learning framework combining graph convolutional recurrent neural networks and LSTM networks. This framework aims to accurately predict citywide passenger demand in ride-sharing platforms by analyzing historical demand data and external factors like weather and time. STG2Seq \cite{bai2019stg2seq} solved the challenges for multi-step passenger demand forecasting in cities. It combines a Graph Convolutional Network (GCN) with an innovative encoder and attention-based output module, effectively capturing spatio-temporal correlations in-demand data and outperforming traditional methods in real-world tests.

\subsubsection{Transportation Recommendation}

\begin{figure*}[ht]
    \centering
    \includegraphics[width=0.85\textwidth]{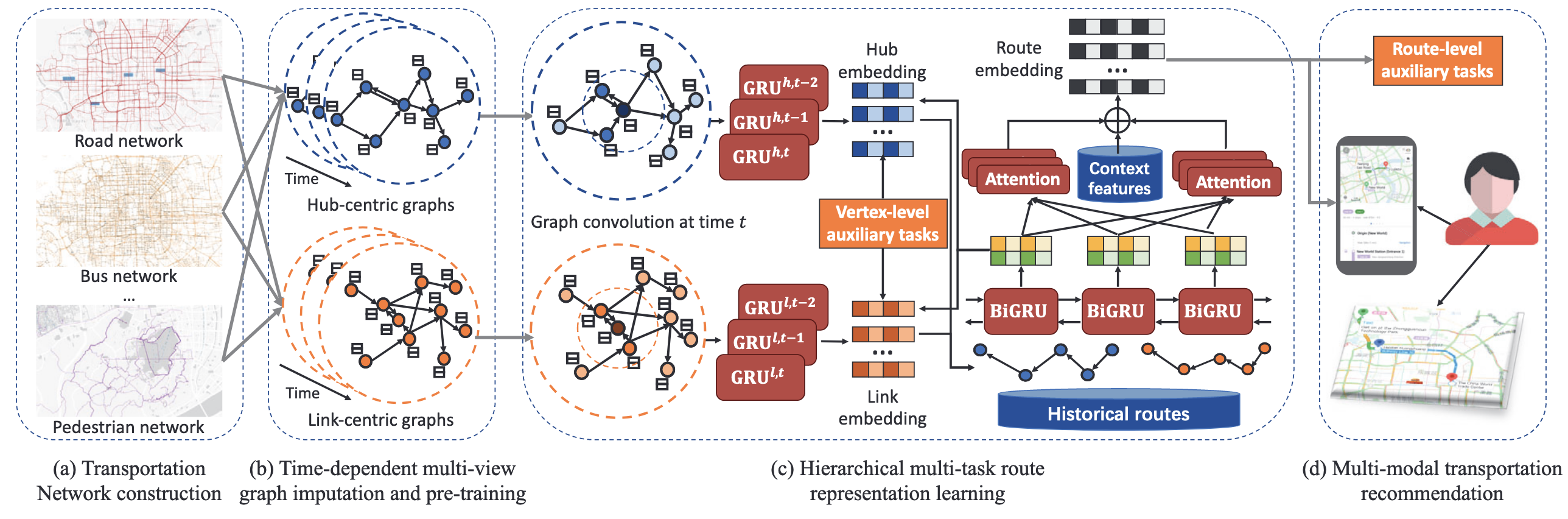}
    \caption{An overview of unified route representation learning for multi-modal transportation recommendation \cite{liu2023unified}. }
    \label{fig:recommondation}
\end{figure*}

Different from public services, \mo{\textbf{transportation recommendations} are proposed for specific personalized needs, significantly facilitating human society's daily life, thus receiving increasing attention in recent years.} 
\citet{solotravel2023} demonstrated that the prevalence of private travel has been on the rise since 2016, with 83 million Americans looking to take a solo trip in 2023.
With the progress of human society and the development of urbanization,
transportation recommendations are expected to create more economic and social benefits in the future.
Among the various applications of transportation recommendations, route Recommendations, multi-modal transportation recommendations, and trip recommendations have garnered the most research attention. In this section, we will primarily introduce them respectively.

\mo{\textbf{Route recommendation} plays a core role in many applications such as taxi services like RoD (ride-on-demand) and navigation \cite{li2017pare, dai2015personalized}.} 
\citet{guo2020force} presented a novel force-directed algorithm for improving route recommendations in ride-on-demand services. This method, inspired by electrostatic principles, utilizes urban data to align vehicle distribution with passenger demand, thereby enhancing route efficiency.

Given the origin and destination, \mo{\textbf{multi-modal transportation recommendations} offer travel plans consisting of diverse transportation modes} (e.g., driving, cycling, and public transit), as well as instructions for making seamless transitions between different modes. Doing so improves individuals' convenience and well-being, particularly for older adults and children.
\citet{liu2019joint} proposed Trans2Vec, a model that enhances transportation mode recommendations by integrating heterogeneous data to capture user and location preferences, thereby improving accuracy in suggesting multi-modal transportation options.
\citet{liu2023unified} developed a novel framework (shown in Figure \ref{fig:recommondation} for multi-modal transportation recommendations. The approach leverages a hierarchical multi-task route representation learning technique that integrates spatio-temporal autocorrelation modeling and coherent-aware attentive route learning. Enhanced by spatio-temporal pre-training strategies, the framework effectively utilizes time-dependent multi-view transportation graphs, offering a more nuanced and accurate transportation recommendation system.

\mo{\textbf{Trip Recommendation} aims to propose a series of POIs for individuals with specific travel preferences, such as the number of attractions, starting and ending points, and so on.}
\citet{he2019joint} introduced a novel context-aware POI embedding model that jointly learns the impact of POI popularity, user preferences, and co-occurring POIs.
Uniquely integrating visual features, Photo2Trip \cite{zhao2017photo2trip} utilized geo-tagged photos with collaborative filtering models to enhance personalized tour recommendations.
GraphTrip \cite{gao2023dual} addressed the sparsity of data and the need for more personalized trip recommendations by effectively integrating diverse knowledge domains. The paper introduces a dual-grained mobility learning approach that uses spatio-temporal graph representation.

\subsection{Economy}

The economy serves as a critical indicator of urban development, and data collected from cities inherently reflects the economic conditions of the regions. For instance, the density of POIs can indicate the popularity of an area, satellite images can provide insights into the urbanization level in a region, and human mobility data can contribute to the estimation of the economic vitality in regions.

\textbf{Real estate price} is a vital indicator for a city which is affected by various factors. To manage citywide real estate information, \citet{duGeofirstLawLearning2019} designed a dual-level collective learning (\textit{DLCL}) framework  to collectively learn spatial representation through intra-region and inter-region geographical structure knowledge which contain POI information as well as geographical and human mobility information for accurate prediction of the real estate price of Beijing. In their framework, intra-region POI and other side information are constructed through multi-view POI-POI networks. The constructed intra-region structure and inter-region autocorrelation are embedded through the Adversarial AutoEncoder.

\citet{jenkinsUnsupervisedRepresentationLearning2019} proposed a model named \textit{RegionEncoder} for various data sources and conducted experiments on Chicago and New York City for the perdition of real estate prices as well. They added satellite image data to the input of the model as a supplementary and gained impressive results in the experiment.

In addition to real estate price prediction, the study conducted by \citet{xiFirstLawGeography2022b} introduced an attention model that combines satellite imagery with POI information to \textbf{estimate various economic indicators}. Their methodology addresses the limitations of previous studies on satellite image data by leveraging the complementary nature of POI data in capturing human activity factors. \citet{li2022predicting} conducted the first research to organize multi-view urban images (including satellite images and street-view images) by leveraging city structural information. The effective combination of the street-view image data contributed to a significant improvement of 10\% in urban economic indicators predicting compared to previous studies before 2022. Besides, \citet{liu2023knowledge} also emphasized the necessity of integrating street-view images with satellite images in their research about urban economy.

\subsection{Public Safety and Security}
In the process of daily urban operations, ensuring public safety and security is a critical responsibility of cities and a standing focus for city administrators and researchers. The activities of humans and the movement of vehicles often encounter various safety hazards. Additionally, natural events like landslides and air pollution can also pose threats to people's lives. The various urban big data provide us with opportunities to gain insights into unsafe factors and predict unsafe events. By integrating and analyzing diverse data sources such as human mobility, social media data, sensor data, and emergency response records, we can identify patterns and trends that may indicate potential safety risks. For example, \citet{zhangMultiviewJointGraph2021a} proposed a multi-view region embedding framework with human mobility, region attribution, and crime records of New York City to predict the number of crime events in each region. \citet{jiangITVInferringTraffic2021} designed a framework to utilize vehicle trajectories and road environment data to infer traffic violation-prone locations in Xiamen City.

\textbf{Traffic safety}, being a paramount concern for numerous countries, has emerged as a prominent focus in recent data fusion studies pertaining to urban security. \citet{wang2021gsnet} proposed a Geographical and Semantic spatio-temporal Network \textit{(GSNet)} model for predicting traffic accidents from taxi order, POI, and weather data. They conducted an experiment on New York City and Chicago based on real-world traffic accident datasets, the result demonstrated a satisfactory result for their model. Then \citet{wangTrafficAccidentRisk2023} designed a cross-scale feature fusion mechanism to integrate features from different scale data as well as a feature fusion component to integrate multi-view features. Based on this work, a multi-task learning framework that can simultaneously predict fine and coarse-grained traffic accident risks was proposed.

\textbf{Environment safety} and the mitigation of natural disasters have also been remarkable research focal points. These areas of study have garnered significant attention due to their profound implications for urban environments and the well-being of communities. Researchers have made notable strides in leveraging data fusion techniques to enhance environmental monitoring, early warning systems, and disaster response strategies. For instance, based on the existing study of trajectories, \citet{luoLetTrajectoriesSpeak2021} combined traffic trajectory information with road networks to efficiently identify traffic bottlenecks on the road. \citet{chen2018radar} and \citet{you2022panda} focused on studying the road obstacle situation or road risks after disasters in a city from its vehicle trajectory, satellite image, and meteorology data. Besides, \citet{songDeepMobLearningDeep2017} built an intelligent system named \textit{DeepMob} in Japan based on users trajectories, earthquake records, text reports, and transportation network data to analyze and predict human behavior and mobility following natural disasters.

\subsection{Social}
The advances in wireless communication and location acquisition technologies enable people to add a location dimension to traditional social networks \cite{narayananStudyAnalysisRecommendation2016}. With the advancement of data fusion advantage and the popularity of recommendation devices, services based on Location-based Social Networks (LBSN) have penetrated people's lives. Users on social media generate a variety of content with geo-information every day. Such posts combine geographical coordinates (always in GPS location) and user-generated content (text, image, or audio) which might be associated with the semantic meaning of those places. It is strongly promising to fusion the geographical information and other information through the users' posts to bridge the gap between users’ activities in digital and physical worlds and contribute to various downstream tasks such us POI or friend recommendation \cite{chenInformationCoverageLocation2015a,caoPointsofinterestRecommendationAlgorithm2020,parkHowUseLocationbased2018,fangTopkPOIRecommendation2023,liHMGCLHeterogeneousMultigraph2023} and community analysis and detection \cite{talpurStudyTouristSequential2018,liu2020spatiotemporal,xuMultidimensionalKnowledgeawareApproach2023}.

\textbf{Recommendation on LBSN} based on deep learning is a complex task in real applications as the irregular spatial structure of geo-location data is non-Euclidean and traditional neural network-based deep learning is incapable of such non-Euclidean data. So graph neural network is widely used to understand the spatial structure information with geo-location data. \citet{fangTopkPOIRecommendation2023} proposed a multi-graph fusion approach for POI recommendation by constructing a user-POI interaction graph on LBSN. For LBSN friend recommendation, \citet{liHMGCLHeterogeneousMultigraph2023} designed a learning framework by leveraging multi-graph to model raw LBSN data and defining various connections between nodes to represent spatio-temporal information. In their methodology, a contrastive learning model was proposed to integrate spatio-temporal features of human trajectories in cities for user node embedding learning.

\subsection{Environment}
Rapid urbanization may result in a potential threat to cities' environment. For example, it is reported that ninety percent of air pollution is caused by urban transportation. Environment protection is an essential topic for urban computing. We have witnessed much research on the environment from different aspects of urban computing, such as air quality prediction \cite{kok2017deep,iskandaryan2020air,kang2018air,wang2017developing,liang2023airformer}, noise controlling \cite{abbaspour2015hierarchal,jezdovia2018smart,jezdovic2021crowdsensing,dutta2017noisesense,liu2020internet} and weather forecasting \cite{zheng2015Forecasting,yonekura2018short,chen2022daily,song2019deep,ghoneim2017forecasting}. Various data sources cloud provide information from various aspects for us to sense the environment. Further, the fusion those information from multiple sources was proven to be effective in understanding and predicting the variation of the environment in a city.

\citet{zheng2015Forecasting} conducted research on cities' air quality based on multi-source big data. They proposed a multi-view hybrid model to predict future 48-hour air quality of a single station point by converging historical air quality data of one station, current meteorological data in the area, weather forecasting data as well as air quality data from other stations. The various meteorological factors and spatial relations between different stations are both considered by effectively fusing these data which significantly improved the capability and accuracy of fine-grain air quality prediction. Further, \citet{ma2023histgnn} utilized a graph neural network to model spatio-temporal correlations between different meteorological variables and different stations as graph neural networks have demonstrated excellent performances for modeling spatio-temporal information. This kind of research makes city-level fine-grained weather forecasting possible with multi-source spatio-temporal data.

\subsection{Energy}
From the perspective of energy, cities can be seen as machines that consume a significant amount of energy. With the acceleration of urbanization and the growing global energy concerns, technologies that enable the perception of energy consumption and energy efficiency have become increasingly important. In urban computing, the fusion of diverse data sources allows us to have a more comprehensive understanding of urban energy consumption and make more efficient urban planning or transportation decisions to help reduce energy consumption.

\subsubsection{Energy Efficient Urban Planning}

    Electric vehicles are regarded as a promising solution for green transportation and sustainable cities. With the gradual expansion of the market share of electrical vehicles, the demand for car charging facilities is increasing. One important issue in urban planning is the selection of charging station locations. Proper selection of these locations not only optimizes city management but also meets the daily commuting needs of users, reducing unnecessary energy consumption during the search for charging stations. The selection of charging station locations should take into account various factors, such as the economic situation of the region, the number of electric vehicles in circulation, residents' travel patterns, road conditions, and traffic conditions. This necessitates the integration and analysis of multiple data sources to estimate the regional charging demand and identify suitable locations for charging stations. \citet{tu2016optimizing} proposed a spatio-temporal demand coverage location model based on the taxi GPS data in Shenzhen, China as well as data from the charging stations. This model provided a charging station sitting approach by considering the waiting time which could hugely reduce energy consumption for charging. \citet{he2018optimal} developed a location method with the consideration of cars' driving range extracted from the GPS data. \citet{battaia2023milp} proposed a framework for charging situation sitting considering the real-life problem of charging infrastructure and sustainability.
    
\subsubsection{Energy Efficient Urban Transportation}
Transportation energy efficiency aims to recommend and navigate energy-efficient travel routes for every transportation participant. Nowadays, within the worldwide energy shortage, fuel prices have been continuously rising \cite{zhao2017effect}. However, traffic congestion in 439 cities in the United States resulted in a total loss of 3.3 billion gallons of wasted fuel \cite{schrank2019urban} in only one year. Traditional route planning studies primarily emphasized faster and more reliable routes, often overlooking the differences in energy consumption among different travel routes and modes for the same travel demand. For instance, in certain situations, taking a route through a congested city center may result in faster arrival times compared to taking a detour on a highway. Despite the first route being shorter in distance, the driver may experience delays due to frequent stops and slow speeds caused by city congestion.

In application, recommending energy-efficient routes for users is a highly complex task. Unlike traditional shortest path problems, it requires considering various factors such as vehicle efficiency, driving habits, road conditions, POIs, and more. The key to addressing this task lies in effectively integrating information from multiple sources.
\citet{wang2022Personalized} researched personalized fuel-efficient route recommendations based on history trajectories. The temporal information from the trajectories as well as the driving factors are input into a transformer model to compose fuel efficiency prediction on the target route. By combining with a genetic algorithm for the best recommendation, this research could effectively find the fuel-efficient routes on the real-world dataset. DeepFEC \cite{elmi2021deepfec} proposed a framework to predict road-level energy consumption through a fusion of contextual vehicle data such as type, weight and engine configuration, and spatio-temporal traffic data on roads. In this framework, a deep convolution residual model is designed to capture the spatial dynamic information from the data and a Bi-LSTM model is responsible for the temporal information. \citet{Oh2019VehicleED} is the first large dataset that combines trajectories of personal cars with users' information, making it possible to mine users' driving behaviors. Based on this multi-source dataset, \citet{laiPreferenceawareMetaoptimizationFramework2023} designed a meta-learning framework for predicting vehicle energy consumption in Ann Arbor, Michigan, USA that can be fine-tuned based on a user's driving preference.

\section{Challenges and Future Directions}
\label{chaellenges}
While urban multi-modal research has made significant advancements in recent years, several challenging issues persist, highlighting directions for potential future research.
As shown in Figure \ref{fig:challenge}, we summarize these challenges and suggest potentially feasible research directions as follows:

\begin{figure}[!h]
    \centering
    \vspace{-1em}
    \includegraphics[width=0.5\textwidth]{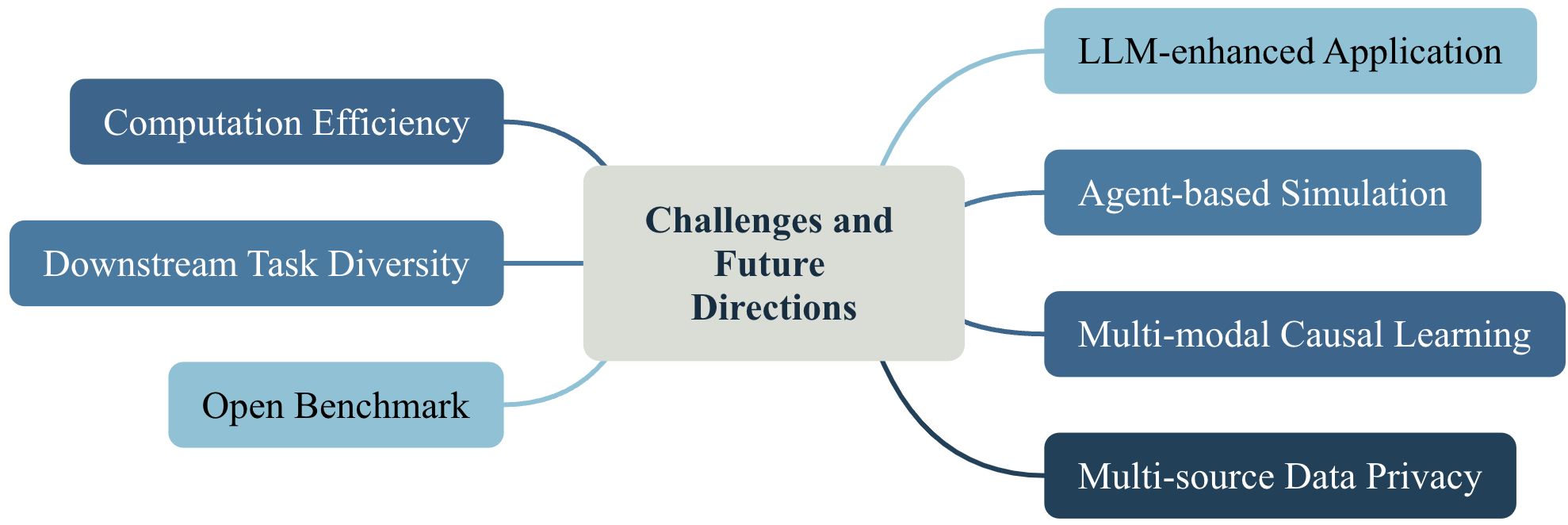}
    \caption{\add{Challenges and future directions of cross-domain data fusion in Urban Computing.}}
    \vspace{-1em}
    \label{fig:challenge}
\end{figure}

\begin{itemize}[leftmargin=*]
    \item \textbf{LLM-enhanced Application:} Since the advent of GPT-4 \cite{openai2023gpt4} and Sora \cite{sora2024}, the academic community has embarked on extensive research on the roles of LLM as predictors \cite{lin2023llm,hu2023unlocking}, enhancers \cite{yin2023heterogeneous,peng2023generating}, controllers \cite{foosherian2023enhancing,shen2023hugginggpt,xu2023urban}, and evaluators \cite{svikhnushina2023approximating,lin2023llm}. However, the field of cross-domain data fusion in urban computing is still in its initial stage in terms of exploring how to apply LLMs effectively. For example, there exists research \cite{roberts2023gpt4geo,bubeck2023sparks,manvi2023geollm} focusing on LLM as urban predictors but also as naive reasoners yet; whereas others \cite{hu2023rsgpt, kuckreja2023geochat} attempt to leverage LLM as model backbone but such works are limited within remote sensing domain instead of general urban computing. \add{A critical limitation currently facing the application of LLMs in Urban Computing is their inability to perceive spatial relationships and dependencies inherent in natural spaces \cite{zou2024learning}, a trait typically associated with language models. For instance, mainstream LLMs struggle to comprehend intricate spatial orientations and positions of urban entities. Moreover, the majority of publicly accessible LLMs are predominantly text-based and exhibit limited performance in other modalities. Nevertheless, we are encouraged by the growing number of successful initiatives in the graph learning domain that aim to enhance LLMs' comprehension of complex relations within graph data \cite{tang2023graphgpt,li2024urbangpt}. Furthermore, the latest multimodal LLMs, such as GPT-4o, demonstrate considerable promise in offering effective multimodal solutions.} Therefore, we eagerly anticipate that the research community will focus on various applications of LLMs in urban computing, exploring their impact similar to their roles in NLP and other domains.
    
    \item \textbf{Agent-based Simulation:} Agent-based simulation aims to model the behavior of individual components or agents to comprehend their interactions and how they collectively contribute to the functioning of the entire system \cite{xi2023rise,wang2023survey}. It has been extensively used in various fields such as biology \cite{an2021agent}, ecology \cite{gonzalez2023predicting}, sociology \cite{flache2022computational}, etc., to model systems where individual entities influence collective behavior. However, the early attempts at agent-based simulation \cite{farmer2009economy,geanakoplos2010leverage,zheng2022ai} are limited because they are not autonomous and need human-defined rules or goals, which may not simulate the complex dynamics of urban systems. Hence, LLM-driven agents such as Urban Generative Intelligence (UGI) \cite{xu2023urban}, can serve as up-to-date solutions for simulating urban dynamics based on cross-domain urban data. This paradigm not only propels forward the field of urban computing, but also paves the way for future cities that are more adaptive and responsive to the evolving needs of their inhabitants. 

    \item \textbf{Multi-modal Causal Learning:} Causal learning or inference aims to investigate causal relationships between variables, ensuring stable and robust learning and inference \cite{neuberg2003causality,pearl2016causal}. Integrating deep learning techniques with causal inference has shown great success in recent years, especially in the fields of spatio-temporal graph forecasting \cite{zhou2023maintaining,xia2024deciphering,li2022ood}, CV \cite{wang2020visual,zhang2020causal,lin2022causal}, NLP \cite{veitch2020adapting,zhang2023causal,tian2022debiasing}, and recommender systems \cite{zheng2021disentangling,gao2022causal}. However, regarding cross-domain data fusion in urban computing, the application of causal learning is still in its early stages. \add{One of the most severe challenges is to represent the complex cross-modal causality in urban scenarios. The deep learning community has yet to converge on a universally accepted approach for accurately yet efficiently representing cross-modal relations. However, as an increasing number of studies delve into the semantic essence of information across various modalities and advance the fields of graph learning, representation learning, and urban foundational models,} we eagerly anticipate further research on multi-modal causal learning of urban data, aiming to improve the interpretability of intricate and dynamic urban systems.
     
    \item \textbf{Multi-source Data Privacy:} Urban data can be highly sensitive, especially in multi-source scenarios in the field of economy and healthcare. When models are trained upon such shared data, there is a risk that they may memorize specific information from the training data, potentially compromising the privacy of individuals. \add{Particularly in the context of deploying data fusion models, the inclusion of data from diverse sources and modalities can indeed bolster the models' performance. However, this also heightens the risk of inadvertently exposing sensitive privacy data. Given these privacy concerns, many institutions and city governments are reluctant to share urban data, thereby impeding the deployment of deep learning models. This poses a significant barrier to the advancement of the community. Consequently, there is a pressing need for research that explores the integration of privacy-preserving techniques, such as differential privacy \cite{dwork2008differential,yang2023local} and federated learning \cite{zhang2021survey,li2021survey}. The goal is to protect data privacy while still reaping the benefits of cross-domain data fusion in urban computing.}

    \item \textbf{Open Benchmark:} The challenge of developing an open benchmark for cross-domain data fusion in urban computing lies in the complexity of integrating diverse data sources, such as sensor data, images, and even text, to understand urban environments comprehensively. This complexity arises due to the heterogeneity of data formats, modalities, and the need for effective fusion methods to extract meaningful insights. A potential solution involves collaborative efforts to standardize data formats, develop unified evaluation metrics, and establish shared benchmarks that facilitate the evaluation and comparison of cross-domain data fusion models.

    \item \textbf{Downstream Task Diversity:} Existing urban research predominantly concentrates on specific task domains such as transportation and urban planning, overlooking the inherent diversity of challenges in real-life urban environments. This limitation exists due to the compartmentalized nature of current research efforts, hindering a holistic understanding of cross-domain data fusion in urban computing. Therefore, we anticipate a more extensive scope of urban research encompassing diverse applications in the realms of economy, society, and environment, providing a thorough comprehension of the intricate conditions prevailing in urban settings.

    \item \textbf{Computation Efficiency:} Current urban research emphasizes the fulfillment of specific applications within cross-domain urban computing, but it overlooks the crucial aspect of computational efficiency. This oversight hampers the practical deployment of these computational models in real-life scenarios. Addressing this challenge requires our focus on optimizing computation efficiency, involving model compression (e.g., knowledge distillation \cite{gou2021knowledge}, low-rank decomposition \cite{yao2023zeroquant,li2023losparse}, and quantization \cite{dettmers2022llm,dettmers2023qlora}), efficient training (e.g., prompt tuning \cite{liu2023pre} and hardware-assisted attention acceleration \cite{dao2022flashattention,dao2023flashattention}), and efficient architecture (e.g., mixture of experts \cite{chen2022towards,shen2023scaling}), to enhance the feasibility and effectiveness of deploying cross-domain data fusion solutions in practical urban environments.
\end{itemize}

\section{Conclusion}
\label{conclusion}
In this survey, we present an extensive and up-to-date survey of cross-domain data fusion tailored for urban computing,  aiming to offer a fresh perspective on this evolving field by introducing a novel taxonomy that categorizes the reviewed fusion methods. In particular, we initially explore the data perspective to understand the significance of each modality and data source, classify the methodology into four types (i.e., \textit{feature-based}, \textit{alignment-based}, \textit{contrast-based}, and \textit{generation-based}), and further categorize cross-domain urban applications into seven types. Besides, We succinctly outline the possible challenges while shedding light on promising directions for future research. The potential for urban computing-related investigations within this captivating cross-domain fusion field is limitless. We hope this can ignite more curiosity and cultivate a long-lasting enthusiasm for cross-domain urban studies, therefore achieving real urban intelligence.


\appendix

\section{Github Link}
Please refer to \url{https://github.com/yoshall/Awesome-Multimodal-Urban-Computing} for comprehensive and up-to-date paper list.

\bibliographystyle{elsarticle-harv} 
\bibliography{multimodal_survey_ref}





\end{document}